\documentclass[10pt,journal,compsoc]{IEEEtran}
\usepackage{graphicx}
\usepackage{amsmath}
\usepackage{eqparbox}
\usepackage{graphicx}
\usepackage{subfigure}
\usepackage{wrapfig}
\usepackage{amsfonts}
\usepackage{setspace}
\usepackage{float}
\usepackage{color}
\usepackage{caption}
\usepackage{makecell}
\usepackage[linesnumbered, ruled, vlined]{algorithm2e}

\newcommand{\eg}{\textit{e.g.}}
\newcommand{\ie}{\textit{i.e.}}

\ifCLASSOPTIONcompsoc
  \usepackage[nocompress]{cite}
\else
  \usepackage{cite}
\fi

\ifCLASSINFOpdf
\else
\fi

\begin{document}
%
\title{Joint Graph Learning and Matching for Semantic Feature Correspondence}

%
%
%
%

\author{He Liu,
        Tao Wang,
        Yidong Li,
        Congyan Lang,
        Yi Jin
        and~Haibin~Ling
\IEEEcompsocitemizethanks{\IEEEcompsocthanksitem H. Liu, T. Wang, Y. Li , C. Lang and Y. Jin are with Beijing Key Laboratory of Traffic Data Analysis and Mining, Beijing Jiaotong University, Beijing 100044, China.\protect\\
E-mail: \{liuhe1996, twang, ydli, cylang, yjin\}@bjtu.edu.cn
\IEEEcompsocthanksitem H. Ling is with the Department of Computer Science, Stony Brook University, Stony Brook, NY 11794-2424. \protect\\
E-mail: hling@cs.stonybrook.edu}
}

%
%

\markboth{Under Review}%
{He \MakeLowercase{\textit{et al.}}: Joint Graph Learning and Matching for Semantic Feature Correspondence}
%



\IEEEtitleabstractindextext{%
\begin{abstract}
In recent years, powered by the learned discriminative representation via graph neural network (GNN) models, deep graph matching methods have made great progresses in the task of matching semantic features. However, these methods usually rely on heuristically generated graph patterns, which may introduce unreliable relationships to hurt the matching performance.
In this paper, we propose a joint \emph{graph learning and matching} network, named GLAM, to explore reliable graph structures for boosting graph matching. GLAM adopts a pure attention-based framework for both graph learning and graph matching.
Specifically, it employs two types of attention mechanisms, self-attention and cross-attention for the task. The self-attention discovers the relationships between features and to further update feature representations over the learnt structures; and the cross-attention computes cross-graph correlations between the two feature sets to be matched for feature reconstruction.
Moreover, the final matching solution is directly derived from the output of the cross-attention layer, without employing a specific matching decision module.
The proposed method is evaluated on three popular visual matching benchmarks (Pascal VOC, Willow Object and SPair-71k), and it outperforms previous state-of-the-art graph matching methods by significant margins on all benchmarks. Furthermore, the graph patterns learnt by our model are validated to be able to remarkably enhance previous deep graph matching methods by replacing their handcrafted graph structures with the learnt ones.
\end{abstract}

}

\maketitle

\IEEEdisplaynontitleabstractindextext

%
\IEEEpeerreviewmaketitle

\IEEEraisesectionheading{\section{Introduction}\label{sec:introduction}}

%
%
%
%
\IEEEPARstart{E}{stablishing} correspondences (or matches) between two sets of semantic features is an important problem in computer vision, and it has been widely involved in many applications such as panoramic stitching~\cite{BrownL07} and object tracking~\cite{gracker}. In the past decades, graph matching has become one of the most popular and powerful branches for feature correspondence,  which regards each feature set as a graph and models the feature correspondence task as a \textit{quadratic assignment problem} (QAP) where the unary similarities and pairwise affinities are incorporated~\cite{Conte2004,Foggia2014,survey_yan}.
Recently, many deep graph matching methods~\cite{NowakVBB18,PCA,DGM_consensus,CIE,SuperGlue} that employ \textit{graphic neural networks} (GNNs) to aggregate neighbor information and form structured representations of nodes and/or edges have demonstrated exciting matching performances.

\begin{figure}[!t]
\begin{center}
\includegraphics[width=\linewidth,height=6cm, trim=48 48 145 0,clip]{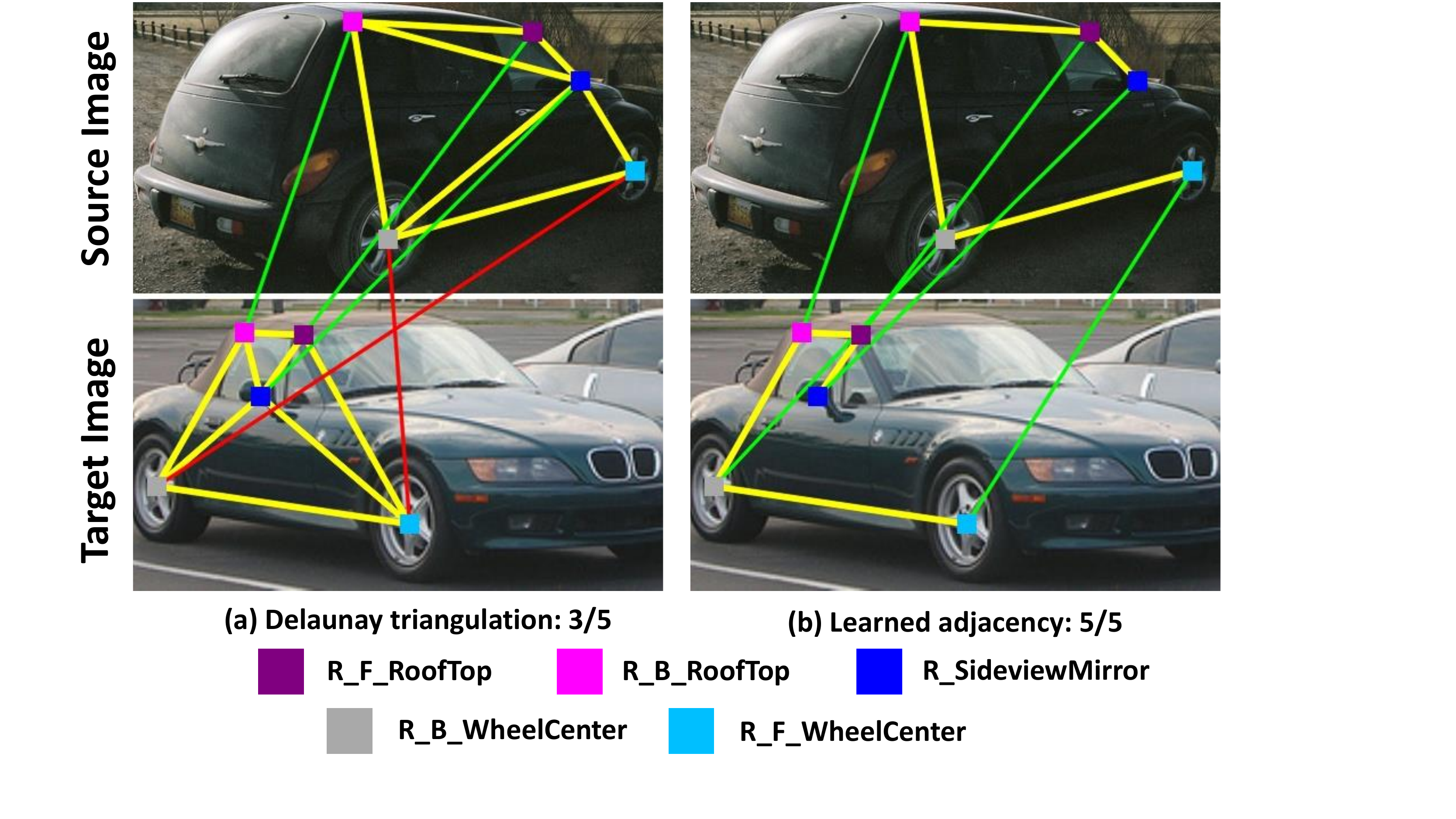}
\end{center}
\caption{The example of matching results derived from different graph structures that are built via (a) Delaunay triangulation, and (b) learned adjacency by our model. The keypoints and built edges are annotated as solid squares and yellow lines in each image, respectively. The correct matches and incorrect matches are represented by green lines and red lines cross two images, respectively.}
\label{fig:H_L_graphs}
\end{figure}

Despite of the progress made by the above mentioned graph matching approaches, they depend heavily on the input graph patterns, which are usually built upon the raw feature sets using some heuristic manners such as Delaunay triangulation, $k$-nearest neighbor, complete graphs, \textit{etc}. Such handcrafting strategies may fail to faithfully explore the semantic relationships between nodes and introduce unreliable structural biases into the built graphs.
For example, Fig.~\ref{fig:H_L_graphs} (a) illustrates the graph structures of two \emph{car} images from the Pascal VOC dataset~\cite{VOC}, which are built using the annotated keypoints via Delaunay triangulation. It is observed that the connections between keypoints \emph{R\_F\_RoofTop},\emph{R\_F\_WheelCenter} and \emph{R\_SideviewMirror} are inconsistent in the two graphs due to the significant viewpoint variation. Such differences in graph structures introduce incorrect relational biases to guide the graph matching, and finally lead to the incorrect matches between two keypoint sets.


Aiming to address the issues mentioned above, in this paper, instead of taking handcrafted graph structures as input, we work directly on the feature sets and learn simultaneously the discriminative feature representations and the semantic relationships to benefit the graph matching.
Inspired by the success of the transformer framework~\cite{transformer} that employs pure attention computation models for representation learning,
we design a joint \emph{graph learning and matching} (GLAM) network by integrating relationship learning, representation learning and matching decision into a unified attention computation framework.
Specifically, our attention module consists of two types of attention layers:
(1) the \emph{self-attention layer} (SAL) aims to learn reliable semantic relationships between keypoints for each feature set, and it forms structured feature representations according to the learnt graph structures; and
(2) the \emph{cross-attention layer} (CAL) computes cross-dependencies among the two feature sets to be matched for reconstructing the feature representations, and it also outputs a soft matching solution based on the computed cross-dependencies.


For evaluating the proposed GLAM network, we report its performance on three popular visual matching benchmarks including Willow Objects~\cite{LG2M}, Pascal VOC Keypoints~\cite{VOC} and SPair-71k~\cite{SPair}, in comparison with state-of-the-art deep graph matching algorithms. The extensive experimental results reveal that our network outperforms all the baseline algorithms by significant margins on all three datasets, which validates its effectiveness and adaptability on various scenarios. Specifically, our algorithm achieves matching accuracies of 99.6\%, 86.2\% and 82.5\%, outperforming the previously strongest competitors by 1.5\%, 6.1\% and 3.6\%, on Willow Objects, Pascal VOC Keyponits and SPair-71k, respectively.


We further validate the learnt graph structures in Sec.~\ref{sec:expe:learning} to illustrate their effectiveness.
We firstly visualize the learnt graph patterns of all categories in the Pascal VOC Keypoints dataset, which exhibit rich and discriminative semantic relationships between keypoints like the manually labeled ones. Moreover, we combine the learnt graph patterns with several state-of-the-art deep graph matching algorithms by replacing their heuristic strategies of graph construction while keeping their main bodies unchanged.
The comparison results show that all these algorithms benefit from the learnt graph patterns in different degrees, which demonstrates the effectiveness of the proposed attention module for relationship learning.
Fig.~\ref{fig:H_L_graphs} (b) shows a representative sample where the learnt graph structures are consistent across two images. Employing the learnt graph patterns to boost graph matching, we achieve perfect matching results between the two keypoint sets.
Our code is publicly available on the Github website~\footnote{https://github.com/LiuHeBJTU/GLAM}.

In summary, with the proposed learnable full attention network for feature corresponding, this paper makes contribution in three-fold:
\begin{itemize}
	\item we analyze shortcomings of graph construction strategies in previous algorithms and propose to learn reliable graph structures to boost graph matching, which is, to the best of our knowledge, among the first attempts to learn proper graph patterns in the deep graph matching framework;
	\item unlike previous deep graph matching algorithms that aggregate node/edge embeddings using various GNN models, we design a fully attention-based network that integrates the learning of graph structures, node representations and matching decisions into a unified framework; and
	\item we validate the excellent performance of the proposed GLAM network on three public benchmarks, and provide insightful analysis and discussion about the effectiveness of learnt graph patterns.
\end{itemize}

%
%


In the rest of this paper, Sec.~\ref{sec:related} summarizes related work; then Sec.~\ref{sec:problem} briefly reviews feature corresponding and graph matching formulation; after that, Sec.~\ref{sec:GLAM} describes the proposed framework for deep graph matching based on attention mechanism. We finally present experimental validations for graph matching in Sec.~\ref{sec:expe:matching} and  graph learning in Sec.~\ref{sec:expe:learning}, respectively.

\section{Related works} \label{sec:related}

\subsection{Feature correspondence}

To solve the feature correspondence problem, several early methods are devoted to searching the optimal corresponding solutions under optimization scheme. For example, Lian and Zha~\cite{LapMatching} model the feature corresponding problem into minimizing a linear assignment function with quadratic regularization items, and relax it to a concave quadratic program problem that is effectively solved by a large scale concave optimizer. Subsequently, Wei~\textit{et al.}~\cite{LapMatching2} extend it to the point matching problem in presence of outliers. For efficiency, the authors deduce the objective function to a function with few nonlinear terms by relaxing the one-to-one matching constraints, and adopt the branch-and-bound algorithm for optimization. Furthermore, CODE~\cite{code} incorporates a coherence-based decision boundary under the assumption that pixels in a local region share the same motion, and embeds a non-linear regression technique into a correspondence likelihood model to solve the feature correspondence problem.

The similar assumption has also been utilized in other studies~\cite{Co-Seg, fast-hough, HVIV, shapematch, NM-Net} that model the feature corresponding problem as a correspondence selection task. Specifically, based on the observation that nearby features on the same object typically share similar homographies, Co-Seg~\cite{Co-Seg}, HPM~\cite{fast-hough} and HVIV~\cite{HVIV} try to identify the correct correspondences by computing the densities of homographies via Hough voting from correspondence candidates. 
In addition, considering that corresponding points on two similar shapes generally have similar shape contexts, Belongie~\textit{et al.}~\cite{shapematch} solve the keypoint corresponding by employing a descriptor, named~\emph{shape context}, to each point for improving the matching accuracy.
Although these methods achieve significant improvement in the feature matching between rigid objects with different viewpoints, they usually fail to obtain satisfied solutions for semantic matching problems in the case of non-rigid objects where keypoints with close semantic relationships do not always share the same motion. Aiming to address the issues mentioned above, instead of directly utilizing the spatial local information for feature correspondence selection, NM-Net~\cite{NM-Net} employs a compatibility-specific neighbor mining layer as a group model to explore reliable neighbors for each correspondence, and embeds it into a hierarchical deep learning network to classify the correspondences into positive or negative ones.


In the above-mentioned methods, the matching solutions are mainly determined under the unary keypoint similarities with local consistent constraints, while the structural agreements between correspondences are not fully explored in their procedure.
By fully exploring structural cues, graph-based methods~\cite{ZhouT13,Cho0DP14,WangLLFH19,LeeLS20} have become one of the most popular and effective branches for semantic feature correspondence, in which features in the object are organized as a graph and the problem of feature correspondence is hence reformulated as the graph matching problem.

\subsection{Graph matching}

Due to the combinatorial nature of the graph matching problem, many efforts~\cite{ABPF,tr_2005ICCV_1,Fact_GM,PFGM,GNCCP,IPFP,Spec_Tech,Graduated_Assignment,BGM,RWGM,PBGM} are devoted to developing approximate algorithms to find acceptable solutions by relax some constraints. Readers can refer the comprehensive surveys~\cite{Conte2004,Foggia2014,survey_yan} for detailed introduction of traditional learning-free graph matching methods.

Recently, with the growing interest in utilizing deep neural network for structured data, learning graph matching with \emph{graph neural network} (GNN) has gained significant success.
A classic way is to relax the graph matching problem to the \emph{linear assignment problem} (LAP) and focus on learning representative node/edge embeddings or robust affinity metrics via message passing mechanism.
For example, Nowak~\textit{et al.}~\cite{NowakVBB18} introduce a Siamese GNN encoder to generate normalized node embeddings for each graph to be matched, and then predict the matching solution by minimizing the cosine distance between the generated embeddings.
Wang~\textit{et al.}~\cite{PCA,PCA-PAMI} employ the \textit{graph convolutional network} (GCN) framework~\cite{KipfW17} to produce node embeddings by aggregating inner-graph and cross-graph structure information, and adopt the Sinkhorn network~\cite{Sinkhorn_network} to solve the relaxed linear assignment problem.
Fey~\textit{et al.}~\cite{DGM_consensus} propose a two-stage graph matching framework that starts from an initial ranking of soft correspondences computed via graph neural networks, and iteratively re-ranks the solution by synchronous message passing networks to reach neighborhood consensus between matched node pairs.

Different from the above mentioned methods that focus mainly on the learning of node and/or edge embeddings and relax the quadratic matching constraints, several recently proposed methods work directly on pairwise affinities and embed strong differentiable solvers for quadratic optimization.
For instance, Zanfir~\textit{et al.}~\cite{DGM} formulate graph matching as a quadratic assignment problem under learnable unary and pairwise affinities, and adopt a spectral matching algorithm~\cite{Spec_Tech} as the combinatorial solver for optimization.
Rol{\'{\i}}nek~\textit{et al.}~\cite{BBGM} relax the graph matching problem based on Lagrangian decomposition, which is solved by
embedding blackbox implementations of a heavily optimized solver~\cite{SwobodaRAKS17} based on dual block coordinate ascent.
Reformulating graph matching as Koopmans-Beckmann's QAP~\cite{LoiolaANHQ07} to minimize the adjacency discrepancy of graphs to be matched, Gao \textit{et al.}~\cite{qcDGM} adopt the Frank-Wolf algorithm~\cite{Frank&Wolfe56} as the solver to obtain approximate solutions.
Besides, Wang~\textit{et al.}~\cite{LGM} integrates learning of affinities and solving for combinatorial optimization into a unified learnable framework, where the graph matching problem is transformed to a binary vertex classification problem of a constructed assignment graph. Later, a similar classification-based framework is proposed in~\cite{NGM}, which models the unary similarities and pairwise agreements into the adjacent matrix of a so-called association graph, and reformulates the graph matching problem as a node classification task under soft one-to-one constraints.

The GNN-based methods mentioned above perform message-passing along edges to form discriminative node representations, and their performances depend largely on the graph structures. However, all of them take the graph structures generated by some heuristic strategies as input, hence their performances are still limited in challenging scenarios where the generated graph patterns may not accurately reflect the inherent relationships between nodes (\textit{e.g.}, Fig~\ref{fig:H_L_graphs}(a)).

\subsection{Attention mechanism}
Attention mechanisms have become an integral part of compelling sequence modeling and transduction models in various \emph{natural language processing} (NLP) tasks. A notable work is the transformer model~\cite{transformer} that relies entirely on self-attention to compute representations of its input and output, which has achieved great success in NLP. Inspired by that, many efforts have been devoted to designing attention-based models in the computer vision community~\cite{DosovitskiyB0WZ21, KhanNHZKS21}.
For example, Xu~\textit{et al.}~\cite{tf_vision_2} propose an attention-based generative adversarial network for text-to-image generation, which consists of multi-stage refinement modules that are designed for generating high-quality sub-regions according to the text-image attention computed by transformer encoder.
For image segmentation, Ye~\textit{et al.}~\cite{tf_vision_3} introduce a cross-modal self-attention network to effectively learn long-range dependencies from constructed multimodal features that represent both visual and linguistic information, enabling the model to adaptively focus on important image regions and informative keywords.
To improve the quality of scene segmentation, Fu~\textit{et al.}~\cite{tf_vision_1} introduce a dual attention network consisting of position attention module and channel attention module, both of which are designed on top of the transformer coder.
For solving image recognition problem, Alexey~\textit{et al.}~\cite{image_tfer} split image into a sequence of flattened patches whose linear projection features are directly taken as the input of transformer encoder to learn the embedding of class tokens. More recently, automatically searching transformer architecture~\cite{ChenPFL21iccv} shows further improvements in visual recognition and downstream tasks.

Recently, several studies have been devoted to introducing attention mechanism into graph neural networks for graph representation learning. Example models include GAT~\cite{GAT} that leverages masked self-attentional layers in a convolution-style neural network, and
U2GNN~\cite{U2GNN} that induces a powerful aggregation function leveraging a self-attention mechanism followed by a recurrent transition.
As for the graph matching task, Yu~\textit{et al.}~\cite{CIE} propose a node and edge embedding strategy that enables the information in each channel to be merged independently, thus improving the robustness of the learned affinities.
SuperGlue~\cite{SuperGlue} borrows the attention encoder from the transformer model and integrates it into graph neural networks to generate discriminative node representations. Specifically, it jointly takes spatial relationships and visual information into considerations for node embedding using intra-graph and inter-graph attentions. As reported in these studies, embedding attention mechanism into graph neural networks improves matching accuracy, but the message passing procedure for updating node/edge embeddings still relies on the graph structures constructed in advance.

Our work falls into the group of deep graph matching algorithms and is designated on the top of attention mechanism. Compared with previous arts, our learning framework focuses on the learning of  not only the structured representations but also their relational inductive biases, which have not been well-explored before. The excellent performance of the proposed method is illustrated in Sec.~\ref{sec:expe:matching}, and the effectiveness of learnt graph patterns is further validated in Sec.~\ref{sec:expe:learning}.


\section{Problem Formulation} \label{sec:problem}

\begin{figure*}[t]
\begin{center}
\includegraphics[width=\linewidth,height=8cm, trim=0 0 0 0,clip]{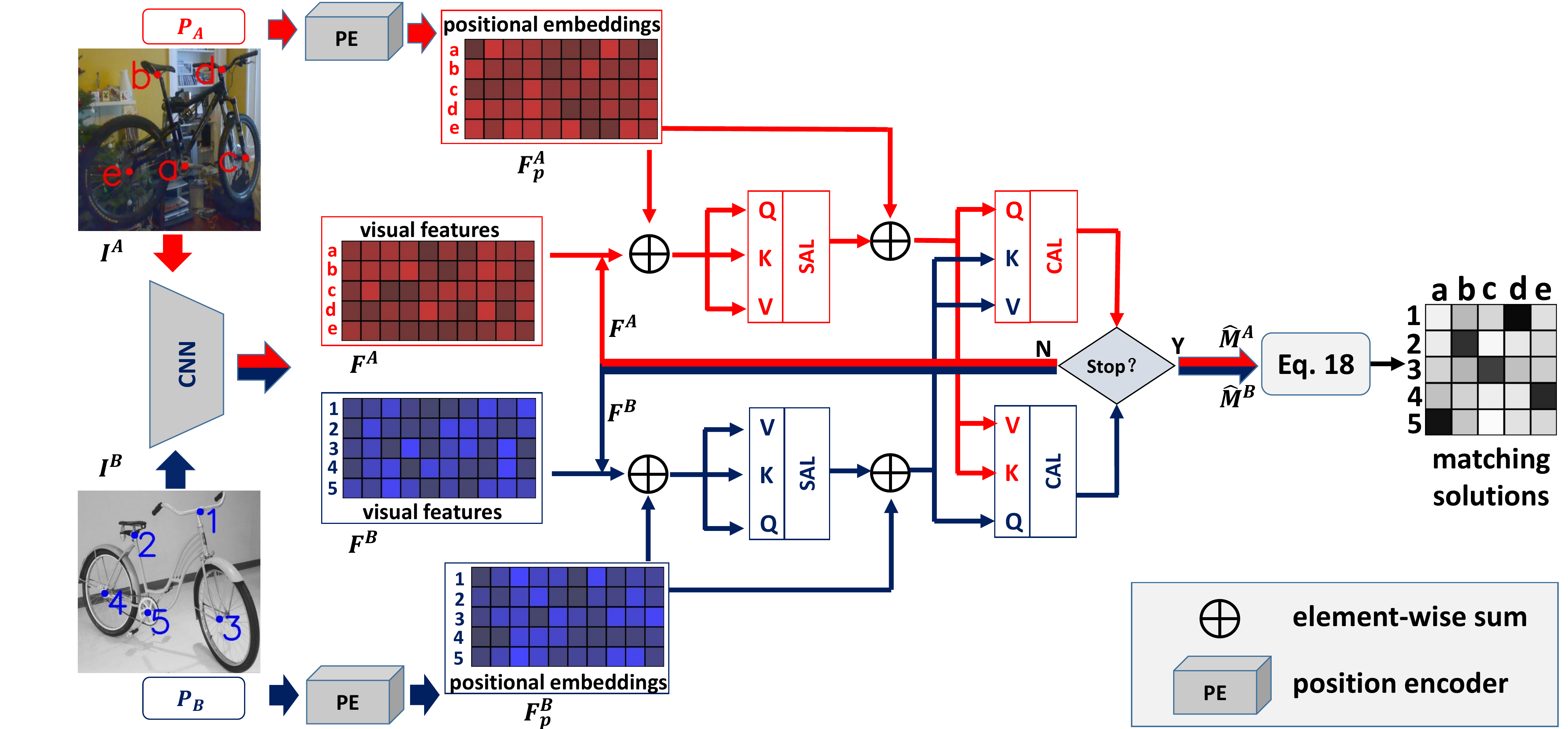}
\end{center}
\caption{The pipeline of the proposed GLAM framework. The element-wise sum of the visual features extracted by CNNs and positional embeddings generated from geometric coordinates is taken as the input of the following learnable attention module. The attention module consists of two types of attention layers: \textit{self-attention layer} (SAL) that learns the graph structure to update the keypoint attributes, and \textit{cross-attention layer} (CAL) that further enhances the representations by incorporating information of the counterpart graph and finally outputs the correspondences between two keypoint sets. }
\label{pipeline}
\end{figure*}
\subsection{Feature Correspondence via Graph Matching}\label{sec:formulation}
We now describe the problem of semantic feature correspondence and review its graph matching formulation. Given two images $I^A$ and $I^B$, two sets of semantic keypoints $\mathbb{V}^A=\{v^A_i|i=1,\cdots,n_A\}$ and $\mathbb{V}^B=\{v^B_i|i=1,\cdots,n_B\}$ are annotated on them, respectively.
For simplicity, we assume the two sets to be matched have the same size $n$, \ie, $n_A=n_B=n$,  while the following formulation/algorithms can be easily extended to handle varied sizes, \eg, by adding dummy keypoints.
The goal of feature correspondence is to find the optimal global correspondences $X\in\{0,1\}^{n\times n}$  from the possible correspondence space $\mathcal{X}=\mathbb{V}^A \times{\mathbb{V}^B}$, under one-to-one matching constraints, \ie, $X\textbf{1}_n=\textbf{1}_n$ and $X^T\textbf{1}_n=\textbf{1}_n$, where $\textbf{1}_n$ denotes a vector of $n$ ones.

As a popular way for feature correspondence, graph matching methods generally build graphs on feature sets to impose structural agreements during finding optimal matching solutions. Specifically, through graph construction strategies such as Delaunay triangulation and $k$-nearest neighbors, the graphs built upon the keypoint sets can be denoted as $\mathbb{G}^A=\{\mathbb{V}^A,\mathbb{E}^A,\mathcal{V}^A,\mathcal{E}^A\}$ and $\mathbb{G}^B=\{\mathbb{V}^B,\mathbb{E}^B,\mathcal{V}^B,\mathcal{E}^B\}$ respectively, where $\mathbb{E}^A$ and $\mathbb{E}^B$ are the edge sets that are usually represented by adjacent matrices, $\mathcal{V}^A, \mathcal{V}^B\in \mathbb{R}^{n\times{d_V}}$ and $\mathcal{E}^A, \mathcal{E}^B \in \mathbb{R}^{n\times{d_E}}$ are the attributes of nodes and edges that may be generated via hand-crafted features or learnable deep features. Given the two graphs, graph matching methods formulate the feature correspondence problem as maximizing the global consistence
\begin{equation}
\mathcal{H}=\sum_{i,a}c_{ia}X_{ia}+\sum_{i,j,a,b}d_{ia,jb}X_{ia}X_{jb},
\end{equation}
where $c_{ia}$ denotes the unary node similarity between node $v_{i}^{A}$ in $\mathbb{G}^{A}$ and node $v_{a}^{B}$ in $\mathbb{G}^{B}$ and  $d_{ia,jb}$ measures the pairwise affinity between edge $(v_{i}^{A},v_{j}^{A})$ in $\mathbb{G}^{A}$ and edge $(v_{a}^{B},v_{b}^{B})$ in $\mathbb{G}^{B}$.

\subsection{Learning of Graph Matching}
In the task of learning feature correspondence, we aim to learn a function $\mu$ 
that maps two feature sets to feature correspondences
\begin{equation}
    X = \mu(F^A, F^B),
\end{equation}
where $F^A, F^B \in \mathbb{R}^{n\times d_V}$ are the input sets of $d_V$-dimensional features  and $X \in \{0,1\}^{n\times n}$ the output assignment matrix.

Under the graph matching pipeline, this function can be in general decomposed into three functions: \textbf{(1)} a graph generation function $\omega$ 
that maps a feature set to graph structures
\begin{equation}
    M = \omega(F),
\end{equation}
where $F \in \mathbb{R}^{n\times d_V}$ is the input feature set and $M \in \mathbb{R}^{n\times n}$ the output weighted adjacency matrix;  \textbf{(2)} a feature updating function $\phi$ that forms structured features according to the graph structures
\begin{equation}
(\tilde{F}^A, \tilde{F}^B) = \phi(F^A, F^B, M^A, M^B),
\end{equation}
where $\tilde{F}^A$ and $\tilde{F}^B$ are the updated feature sets; and \textbf{(3)} a matching decision function $\tau$ that maps two graphs with updated features to the final matching solution
\begin{equation}
X = \tau (\tilde{F}^A, \tilde{F}^B, M^A, M^B).
\end{equation}

In previous deep graph matching algorithms (\eg, ~\cite{NowakVBB18,PCA,DGM_consensus,CIE,SuperGlue,LGM,NGM,BBGM,qcDGM}), the graph generation functions $\omega$ is designed in a heuristic way, for instance Delaunay triangulation graphs, $k$-nearest neighbor graphs or complete graphs; the feature updating functions $\phi$ are simulated by various multi-layer GNN models; and the matching decision function $\tau$ are usually implemented as the differentiable versions of some combinatorial solvers, \textit{e.g.}, the Sinkhorn network~\cite{Sinkhorn_network}.

In order to explore reliable graph structures to boost graph matching, we proposed to learn a proper graph generation function $\omega$, rather than using a heuristic one. Furthermore, these three functions are not designed separately, but are intertwined into a unified attention-based framework, which is described in detail in Sec.~\ref{sec:attention}.

\section{Full Attention-based Graph Matching Network} \label{sec:GLAM}

Given two images with the semantic keypoints to be matched, our GLAM network first extracts the visual attributes and positional embeddings, and takes the element-wise sum of them as the input of the following learnable attention module.
The attention module adopts both self-attention and cross-attention mechanism for learning graph structures, updating keypoint features and finally computing the matching solutions.
Note that, our network produces only soft matching solutions for the necessary of training, and we adopt the Hungarian method~\cite{hungarian,munkres} as a post-procedure to provide desired discrete solutions in the inferring step.

Fig.~\ref{pipeline} illustrates the pipeline of the proposed GLAM framework, and detailed descriptions of each key component are provided in the following subsections.

\subsection{Feature extraction}
Given an image $I^A$, we follow most deep graph matching methods to produce the raw visual features using a \emph{Convolution Neural Network} (CNN). Specifically, we use VGG16~\cite{VGG16} to prepare the feature maps of the whole image, and employ interpolation strategy to extract the visual features for keypoints.

Considering that locations of keypoints  provide geometric cues, we augment the visual features with location information by designing a learnable position encoder $\rho$ that transforms the keypoint locations into high-dimension positional embeddings.
Denoting the visual features of $n$ keypoints in image $I^A$ as $F^{A}\in \mathbb{R}^{n\times d}$, and the location coordinates as $P^A\in \mathbb{R}^{n\times 2}$, the feature augment procedure can be expressed as
\begin{equation}
F^{A} \leftarrow F^{A} + \rho(P^A),
\end{equation}
where the position encoder $\rho$ is parameterized as a \emph{multi-layer perceptron} (MLP) that maps $2$-dimension coordinates into $d$-dimension embeddings. The same CNN and position encoder $\rho$ are also applied on image $I^B$ to produce the keypoint attributes $F^{B}$.

Note that, to make full use of the location information, in each iteration we add the positional embeddings with the updated features output from self-attention module and cross-attention module.

\subsection{The Attention Module} \label{sec:attention}
As the core component of the proposed framework, the attention module consists of \emph{self-attention layer} (SAL) and \emph{cross-attention layer} (CAL), which are devoted to updating features based on the intra- and inter-dependencies between keypoints and finally outputing the matching result.

The key idea of the attention layer is akin to the transformer encoder~\cite{transformer} that maps the queries, keys and values to output vectors and uses the outputs for feature updating.
In SAL, our GLAM network learns the intra-dependencies, \ie, \emph{the weighted graph}, between keypoints in the same image, and then updates the representations according to the learnt relationship between keypoints. By contrast, CAL is designed for incorporating the information from the other keypoint set to enhance the discrimination of features, and it finally outputs a soft matching result between two keypoint sets.

The SAL and CAL are stacked iteratively in the attention module, of which the pipeline is summarized in Algorithm~\ref{algo}.

\subsubsection{Self-attention layer}
Given the input keypoints with attributes $F^A$ extracted from image $I^A$, we design a multi-head SAL to learn the graph structures and then update keypoint features based on learnt structures.

Specifically, we first initialize the queries $Q^A$, keys $K^A$ and values $V^A$ as $Q^A=K^A=V^A = F^A$, and then project them into different latent spaces as
\begin{equation}                                                                                      
\bar{Q}^A_{i} \leftarrow Q^A \bar{W}^Q_{i},\quad
\bar{K}^A_{i} \leftarrow K^A \bar{W}^K_{i},\quad
\bar{V}^A_{i} \leftarrow V^A \bar{W}^V_{i},
\end{equation}
where $\bar{W}^Q_{i}, \bar{W}^K_{i},\bar{W}^V_{i} \in \mathbb{N}^{d\times d_S}$ are the projection parameters of the $i$-th attention head in SAL.

Subsequently, the attentions between the queries and keys are computed by
\begin{equation}
\label{eq:self_attention_M}
\bar{M}^A_{i} \leftarrow {\rm softmax}\Big(\frac{\bar{Q}^A_{i}(\bar{K}^A_{i})^\textrm{T}}{\sqrt{d_S}} \Big),
\end{equation}
where the operator ${\rm softmax}(\cdot)$ performs softmax normalization along each row of the input matrix.
Based on the attention matrix $\bar{M}_{i}$, we can reconstruct the keypoint attributes by
\begin{equation}
    \bar{F}^A_{i} \leftarrow \bar{M}^A_{i}\bar{V}^A_{i}.
\end{equation}
The concatenation of updated attributes of all attention heads is transformed into a new feature space
\begin{equation}
\bar{F}^A \leftarrow [\bar{F}^A_{1}, \bar{F}^A_{2},\cdots, \bar{F}^A_{N_S}]W^S,
\end{equation}
where $N_S$ is the number of attention heads, $W^S\in\mathbb{N}^{(d_S\times{N_S})\times d}$ is a learnable parameter matrix and $[\cdot,\cdots,\cdot]$ concatenates its input along the channel wise.

Finally, the updated features are passed through a residual learning~\cite{resnet} architecture
\begin{equation}
F^A \leftarrow {\rm ReLU}(F^A + \bar{F}^A),
\end{equation}
where ${\rm ReLU}(z)=\max(0,z)$ maps its input in $[0,+\infty)$.

As each training sample contains a pair of input feature sets $(F^A, F^B)$, we apply the same self-attention computations with same parameters, $\bar{W}^Q_{i}, \bar{W}^K_{i},\bar{W}^V_{i}$ and $W^S$, on $F^B$ to learn its graph structures and update the keypoint attributes.

\begin{algorithm}[!t]
\setstretch{1.2}
\caption{The attention module.}
\label{algo}
\KwIn{

$F^A, F^B$: visual features of keypoints in $I^A$ and $I^B$;
$\rho{(P^A)},\rho{(P^B)}$: positional embeddings in $I^A$ and $I^B$\;
$N_L$: the number of attention layers\;
\mbox{$N_S, N_C$: the numbers of self- and cross-attention heads;}
$\bar{W}^Q_{i},\bar{W}^K_{i},\bar{W}^V_{i}, W^S $: learnable parameters in SAL\;
$\widehat{W}^Q_{i},\widehat{W}^K_{i},\widehat{W}^V_{i}, W^C $: learnable parameters in CAL.
}
\KwOut{the matching solution $X$.}

initialization: $F^A\leftarrow F^A + \rho{(P^A)}$;~$F^B\leftarrow F^B + \rho{(P^B)}$\;
\For {$t=1$; $t\leq N_L$; $t++$ }
{
    \tcp{\textit{self-attention layer}}
    $Q^A \leftarrow F^A, K^A \leftarrow F^A, V^A \leftarrow F^A$\;
    $Q^B \leftarrow F^B, K^B \leftarrow F^B, V^B \leftarrow F^B$\;
    \For {$i=1$; $i\leq N_S$; $i++$}
    {
    \mbox{$\bar{Q}^A_{i} \leftarrow Q^A \bar{W}^Q_{i}, \bar{K}^A_{i} \leftarrow K^A \bar{W}^K_{i}, \bar{V}^A_{i} \leftarrow V^A \bar{W}^V_{i} $;}\\
    $\bar{F}^A_{i} \leftarrow {\rm softmax}(\frac{\bar{Q}^A_{i}(\bar{K}^A_{i})^\textrm{T}}{\sqrt{d_S}}) \bar{V}^A_{i}$\;
    \mbox{$\bar{Q}^B_{i} \leftarrow Q^B \bar{W}^Q_{i}, \bar{K}^B_{i} \leftarrow K^B \bar{W}^K_{i}, \bar{V}^B_{i} \leftarrow V^B \bar{W}^V_{i} $;}\\
    $\bar{F}^B_{i} \leftarrow {\rm softmax}(\frac{\bar{Q}^B_{i}(\bar{K}^B_{i})^\textrm{T}}{\sqrt{d_S}}) \bar{V}^B_{i}$\;
    }
    $\bar{F}^A \leftarrow [\bar{F}^A_{1}, \bar{F}^A_{2},\cdots \bar{F}^A_{N_S}]W^S$\;
    $\bar{F}^B \leftarrow [\bar{F}^B_{1}, \bar{F}^B_{2},\cdots \bar{F}^B_{N_S}]W^S$\;
    $F^A\leftarrow {\rm ReLU}(F^A + \bar{F^A}) + \rho{(P^A)}$\;
    $F^B\leftarrow {\rm ReLU}(F^B + \bar{F^B}) + \rho{(P^B)}$\;

    \tcp{\textit{cross-attention layer}}
    $Q^A \leftarrow F^A, K^A \leftarrow F^B, V^A \leftarrow F^B$\;
    $Q^B \leftarrow F^B, K^B \leftarrow F^A, V^B \leftarrow F^A$\;
    \For {$i=1$; $i\leq N_C$; $i++$}
    {
    \mbox{$\widehat{Q}^A_{i} \leftarrow Q^A \widehat{W}^Q_{i}, \widehat{K}^A_{i} \leftarrow K^A \widehat{W}^K_{i}, \widehat{V}^A_{i} \leftarrow V^A \widehat{W}^V_{i} $;}\\
    $\widehat{M}^A_{i} \leftarrow {\rm sinkhorn}({\rm sgd}(\frac{\widehat{Q}^A_{i}(\widehat{K}^A_{i})^\textrm{T}}{\sqrt{d_C}})) $\;
    \mbox{$\widehat{Q}^B_{i} \leftarrow Q^B \widehat{W}^Q_{i}, \widehat{K}^B_{i} \leftarrow K^B \widehat{W}^K_{i}, \widehat{V}^B_{i} \leftarrow V^B \widehat{W}^V_{i} $};\\
    $\widehat{M}^B_{i} \leftarrow {\rm sinkhorn}({\rm sgd}(\frac{\widehat{Q}^B_{i}(\widehat{K}^B_{i})^\textrm{T}}{\sqrt{d_C}})) $\;
    $\widehat{F}^A_{i} \leftarrow \widehat{M}^A_{i} \widehat{V}^A_{i}, \quad \widehat{F}^B_{i} \leftarrow \widehat{M}^B_{i} \widehat{V}^B_{i} $\;
    }
    $\widehat{F}^A \leftarrow [\widehat{F}^A_{1}, \widehat{F}^A_{2},\cdots \widehat{F}^A_{N_C}]W^C$\;
    $\widehat{F}^B \leftarrow [\widehat{F}^B_{1}, \widehat{F}^B_{2},\cdots \widehat{F}^B_{N_C}]W^C$\;
    $F^A\leftarrow {\rm ReLU}(F^A + \widehat{F}^A) + \rho{(P^A)}$\;
    $F^B\leftarrow {\rm ReLU}(F^B + \widehat{F}^B)+ \rho{(P^B)}$\;
}

$X \leftarrow \frac{1}{2 N_C}\sum^{N_C}_{i=1} (\widehat{M}^A_{i} + (\widehat{M}^B_{i})^\textrm{T})$\;
\Return{$X$}\;
\end{algorithm}

\subsubsection{Cross-attention Layer}

 Different from SAL, CAL aims to enhance the keypoints attributes in image $I^A$ using the information from the counterpart image $I^B$. Thus, we need to initial the queries $Q^A$, keys $K^A$ and values $V^A$ as
 \begin{equation}
 Q^A\leftarrow F^A, \quad
 K^A\leftarrow F^B, \quad
 V^A\leftarrow F^B.
 \end{equation}

Similar to SAL,  CAL employs a multi-head architecture to project the input attributes into different latent spaces as
\begin{equation}
\widehat{Q}^A_{i} \leftarrow Q^A \widehat{W}^Q_{i},\quad
\widehat{K}^A_{i} \leftarrow K^A \widehat{W}^K_{i},\quad
\widehat{V}^A_{i} \leftarrow V^A \widehat{W}^V_{i},
\end{equation}
where $\widehat{W}^Q_{i}, \widehat{W}^K_{i},\widehat{W}^V_{i} \in \mathbb{N}^{d\times d_C}$ are the projection parameters of the $i$-th attention head in CAL.

Afterward, the cross-attentions between the queries and keys are computed as
\begin{equation}
\widehat{M}^A_{i} = {\rm sinkhorn}\left({\rm sgd}\Big(\frac{\widehat{Q}^A_{i} (\widehat{K}^A_{i})^\textrm{T}}{\sqrt{d_C}} \Big) \right),
\end{equation}
where ${\rm sgd}(z)=\frac{1}{1+e^{-z}}$ is a sigmoid function that maps its input into $(0,1)$ for meeting the nonnegative requirement of the input of Sinkhorn network~\cite{Sinkhorn_network}. Unlike most attention mechanisms that use softmax function for row-wise normalization, here we adopt Sinkhorn network to impose both row- and column-wise normalization. Subsequently, the reconstruction of the query attributes is conducted by
\begin{equation}
\widehat{F}^A_{i} = \widehat{M}^A_{i} \widehat{V}_{i}.
\end{equation}


Similar to SAL, we transform the concatenation of the outputs of attention heads through a learnable parameter matrix $W^C \in \mathbb{N}^{(d_C\times{N_C})\times d}$ as
\begin{equation}
\widehat{F}^A = \big[\widehat{F}^A_{1}, \widehat{F}^A_{2},\cdots \widehat{F}^A_{N_C}\big] W^C,
\end{equation}
where $N_C$ is the number of attention heads in CAM. Finally we update the keypoint attributes using a residual learning scheme
\begin{equation}
F^A \leftarrow {\rm ReLU}(F^A + \widehat{F}^A).
\end{equation}

Note that the mirror-like computation is also applied in CAL to update the feature set $F^B$ and compute the cross-attention matrix $\widehat{M}^B_{i}$, and we average these cross-attention matrices at the last layer as the output matching solution
\begin{equation}
X \leftarrow \frac{1}{2 N_C}\sum^{N_C}_{i=1} \big(\widehat{M}^A_{i} + (\widehat{M}^B_{i})^\textrm{T} \big).
\end{equation}

%
%
%

\subsection{Loss function}
Given the predicted soft matching solutions $X$ and the ground-truth matching solutions $X^{gt}$, we consider labeling positive matches as a binary classification task, and use the differences between them as the supervision signal for end-to-end training. Thus the loss for the predicted solutions is a weighted cross entropy function
\begin{equation}\label{loss}
  \mathcal{L}= -\sum_{i=1}^{n^2} \Big[ w\ \textbf{x}_i^{gt} \log(\textbf{x}_i) \ + \ (1-w) (1-\textbf{x}_i^{gt}) \log(1-\textbf{x}_i) \Big],
\end{equation}
where $\textbf{x}$ and $\textbf{x}^{gt}$ are the vectorization of $X$ and $X^{gt}$, respectively, and $w$ is a hyper-parameter that balances the loss to avoid the dominance of the negative candidate matches during training.

\section{Experiments on Graph Matching}\label{sec:expe:matching}
To validate the effectiveness of our GLAM model, we evaluate it on three public visual semantic matching benchmarks, named Willow Object dataset~\cite{LG2M}, Pascal VOC Keypoints~\cite{VOC} and SPair-71k~\cite{SPair}, in comparison with several state-of-the-art graph matching methods including GMN~\cite{DGM}, SuperGlue~\cite{SuperGlue}, PCA~\cite{PCA}, IPCA~\cite{PCA-PAMI}, LCS~\cite{LGM}, qc-DGM~\cite{qcDGM}, DGMC~\cite{DGM_consensus}, CIE~\cite{CIE}, BBGM~\cite{BBGM}, NGM~\cite{NGM} and NGM-v2~\cite{NGM}. Among these baseline methods, IPCA~\cite{PCA-PAMI} extends the PCA~\cite{PCA} by adding an iterative update strategy for cross-graph embeddings. NGM-v2~\cite{NGM} is the upgraded version of NGM~\cite{NGM} by augmenting the raw features with SplineCNN~\cite{splineCNN}. Furthermore, SuperGlue~\cite{SuperGlue} and CIE~\cite{CIE} are the methods designed by integrating the attention mechanism into graph neural networks for feature aggregation.

\subsection{Settings}
In experiments, we generate the visual feature  following PCA~\cite{PCA} that produces the raw keypoint features by interpolating the image feature maps extracted through the standard VGG16 backbone~\cite{VGG16}. The dimensions of visual features and positional embeddings are both 1024 in practice.

For the attention module, we set the number of attention layers $N_L= 3$, the numbers of attention heads $N_S=N_C=8$, and fix the dimensions of the features as $d_S = d_C = d = 1024$. For CAL, considering that too much iterations of normalization in Sinkhorn network~\cite{Sinkhorn_network} will weaken the gradient used for training our model, we set the number of row- and column-normalization iterations for cross-relationships to 5. In the weighted cross entropy loss, we set the weight $w=5$ to balance the significance of positive labels and negative labels in training.

\subsection{Comparison on Benchmarks}

\subsubsection{Willow Object dataset}

\begin{table}[t]
\centering
\caption{Matching accuracy (\%) on the Willow Object dataset.  \textbf{Bold numbers} represent the best results.}
\label{table:willow}
\footnotesize
\begin{tabular}
{@{\hspace{1.5mm}}l@{\hspace{1.mm}} | @{\hspace{1.5mm}}c@{\hspace{1.mm}} @{\hspace{1.5mm}}c@{\hspace{1.mm}} @{\hspace{1.5mm}}c@{\hspace{1.mm}} @{\hspace{1.mm}}c@{\hspace{1.mm}} @{\hspace{1.mm}}c@{\hspace{1.mm}}
| @{\hspace{1.mm}}c@{\hspace{1.mm}}}
  \hline
    Algorithm & Car & Duck & Face & Motorbike & Winebottle & Ave.\\
  \hline
  GMN~\cite{DGM}                & 74.3  & 82.8  & 99.3          & 71.4  &76.7   & 80.9 \\
  PCA~\cite{PCA}                & 84.0  & 93.5  & \textbf{100}  & 76.7  &96.9   & 90.2\\
  LCS~\cite{LGM}                & 91.2  & 86.2  & \textbf{100}  & 99.4  &97.9   & 94.9 \\
  qc-DGM~\cite{qcDGM}           & 98.0  & 92.8  & \textbf{100}  & 98.8  &99.0   &97.7\\
  DGMC~\cite{DGM_consensus}     & 98.3  & 90.2  & \textbf{100}  &98.5  & 98.1  & 97.0  \\
  CIE~\cite{CIE}& 82.2  & 81.2  & \textbf{100}  & 90.0  & 97.6  & 90.2  \\
  BBGM~\cite{BBGM}              & 96.9  & 89.0  & \textbf{100}  & 99.2  & 98.8  & 96.8  \\
  NGM~\cite{NGM}                & 84.2  &77.6   &99.4  & 76.8  & 88.3  & 85.3\\
  NGMv2~\cite{NGM}                & 97.4  &93.4   & \textbf{100}  & 98.6  & 98.3  & 97.5\\
  SuperGlue~\cite{SuperGlue}    &98.2   &93.3   &\textbf{100}   &99.1   &99.8 &98.1\\
  \hline
  GLAM(ours)                          & \textbf{99.0} & \textbf{99.0} & \textbf{100} & \textbf{100} & \textbf{100} & \textbf{99.6}\\
  GLAM-\={S}                 &98.4           &93.8           &\textbf{100}  & \textbf{100} & \textbf{100} & 98.4 \\
  GLAM-\={C}                  &98.6           &95.7           &\textbf{100}  & 99.8         & \textbf{100} & 98.8
   \\
\hline
\end{tabular}
\end{table}

The Willow Object dataset provided in~\cite{LG2M} contains five categories, each of which is represented by 40 or more images with different instances. In this dataset, each target object is annotated with 10 distinctive landmarks, and each pair of keypoint sets from the same category can form a matching sample. For splitting the dataset, we following~\cite{LGM,PCA,PCA-PAMI} choose 20 images from each category for training and keep the rest for testing. In practice, any pair of images from the same category can be taken as a sample, thus the training set contains totally 2,000 training samples. For testing, we randomly select 1,000 pairs of images from the testing set of each category to evaluate the compared approaches.

The experimental results on Willow Object dataset~\cite{LG2M} are summarized in Table~\ref{table:willow}. Due to lack of scale, pose and illumination changes, this dataset is considered relatively easy. The heuristic graph construction strategies of the compared baseline algorithms are able to explore consistent graph patterns across each category, and thus most of them achieve satisfying matching performance on this dataset, especially in the \emph{Face} category. While our model works directly on keypoint sets to learn proper graph patterns, and it obtain the best matching accuracies in all categories and rises the average accuracy to 99.6\%.

\subsubsection{Pascal VOC Keypoint}

\begin{table*}[ht]
\centering
\caption{Comparison of matching accuracy (\%) on the Pascal VOC dataset. Bold \textbf{numbers} represent the best results. }
\label{table:Pascal}
\begin{tabular}
{@{\hspace{.1mm}}c@{\hspace{.2mm}} | @{\hspace{.75mm}}c@{\hspace{.5mm}} @{\hspace{.5mm}}c@{\hspace{.5mm}} @{\hspace{.5mm}}c@{\hspace{.5mm}} @{\hspace{.5mm}}c@{\hspace{.5mm}} @{\hspace{.5mm}}c@{\hspace{.5mm}}
@{\hspace{.5mm}}c@{\hspace{.5mm}} @{\hspace{.5mm}}c@{\hspace{.5mm}} @{\hspace{.5mm}}c@{\hspace{.5mm}} @{\hspace{.5mm}}c@{\hspace{.5mm}} @{\hspace{.5mm}}c@{\hspace{.5mm}} @{\hspace{.5mm}}c@{\hspace{.5mm}} @{\hspace{.5mm}}c@{\hspace{.5mm}} @{\hspace{.5mm}}c@{\hspace{.5mm}} @{\hspace{.5mm}}c@{\hspace{.5mm}} @{\hspace{.5mm}}c@{\hspace{.5mm}} @{\hspace{.5mm}}c@{\hspace{.5mm}} @{\hspace{.5mm}}c@{\hspace{.5mm}} @{\hspace{.5mm}}c@{\hspace{.5mm}} @{\hspace{.5mm}}c@{\hspace{.5mm}} @{\hspace{.5mm}}c@{\hspace{.5mm}}
| @{\hspace{.65mm}}c@{\hspace{.5mm}}}
  \hline
    Algo. & aero & bike & bird & boat & bot. &bus & car &cat& cha. &cow & tab. &dog & hor.& mbi. & per. &pla. &she. &sofa& tra. & tv& AVG\\
  \hline
  GMN~\cite{DGM} & 31.9 &47.2& 51.9& 40.8 &68.7 &72.2 &53.6 &52.8 &34.6 &48.6 &72.3 &47.7 &54.8 &51.0 &38.6 &75.1 &49.5 &45.0 &83.0 &86.3 &55.3  \\
  SuperGlue~\cite{SuperGlue} &30.4 & 42.0 &44.7 &58.8 &84.4 &77.1 &58.2 &42.4 &35.1 &43.0 &94.7 &41.9 &43.3 &40.5 &44.1 &97.7 &47.8& 49.8 &93.7 &93.3 &58.1\\
  PCA~\cite{PCA}& 51.2 &61.3& 61.6 &58.4 &78.8 &73.9 &68.5 &71.1 &40.1 &63.3 &45.1 &64.4 &66.4 &62.2 &45.1 &79.1 &68.4 &60.0 &80.3 &91.9 &64.6 \\
  IPCA~\cite{PCA-PAMI} &51.0 &64.9 &68.4 &60.5 &80.2 &74.7 &71.0 &73.5 &42.2 &68.5 &48.9 &69.3 &67.6 &64.8 &48.6 &84.2 &69.8 &62.0 &79.3 &89.3 &66.9\\
  LCS~\cite{LGM}   & 46.9 &58.0& 63.6& 69.9& 87.8& 79.8& 71.8 &60.3 &44.8& 64.3 &79.4& 57.5& 64.4 &57.6 &52.4 &96.1& 62.9 &65.8 &94.4 &92.0& 68.5  \\
  qc-DGM~\cite{qcDGM}& 49.6 &64.6 &67.1& 62.4 &82.1& 79.9& 74.8& 73.5& 43.0& 68.4 &66.5 &67.2 &71.4 &70.1& 48.6 &92.4 &69.2 &70.9 &90.9& 92.0 &70.3\\
  DGMC~\cite{DGM_consensus}  & 50.4 &67.6 &70.7 &70.5 &87.2& 85.2 &82.5 &74.3 &46.2 &69.4 &69.9 &73.9 &73.8& 65.4 &51.6 &98.0 &73.2& 69.6 &94.3 &89.6 &73.2  \\
  CIE~\cite{CIE}& 51.2 &69.2 &70.1 &55.0 &82.8& 72.8& 69.0& 74.2& 39.6 &68.8 &71.8 &70.0 &71.8 &66.8 &44.8 &85.2 &69.9& 65.4 &85.2 &92.4 &68.9  \\
  BBGM~\cite{BBGM}&61.5 &75.0 & 78.1 & \textbf{80.0} & 87.4 & 93.0 & 89.1 & 80.2 &58.1 &77.6 &76.5 &79.3 &78.6 &78.8 &66.7 &97.4 &76.4 &77.5 &97.7 & \textbf{94.4} &80.1  \\
  NGM~\cite{NGM} &50.1 &63.5 &57.9 &53.4 &79.8 &77.1 &73.6& 68.2& 41.1& 66.4 &40.8 &60.3 &61.9& 63.5& 45.6 &77.1 &69.3 &65.5 &79.2 &88.2 &64.1 \\
  NGMv2~\cite{NGM}&61.8 &71.2 &77.6 &78.8 &87.3 &93.6 & 87.7 &79.8 &55.4 & 77.8 & 89.5 &78.8 & 80.1 &79.2 & 62.6 & 97.7 & 77.7 & 75.7 & 96.7 & 93.2 & 80.1\\
  \hline
  GLAM(ours)  &\textbf{72.3} & \textbf{76.8} & \textbf{84.3} & 77.4 & \textbf{94.9} & \textbf{95.7} & \textbf{93.8} &\textbf{85.9} & \textbf{72.6} &\textbf{87.9} & \textbf{100} & \textbf{86.2} & \textbf{85.2} & \textbf{85.3} &\textbf{71.4} &\textbf{98.9} & \textbf{83.8} &\textbf{80.5} & \textbf{98.8} & 92.8 &\textbf{86.2}  \\
  GLAM-\={S} &34.9 & 47.5 & 59.9 & 68.8 & 83.8 & 86.6 & 82.9 &62.9 & 44.6 & 58.2 & 94.0 & 60.0 & 64.8 & 58.8 & 46.6 &98.0 & 64.7 & 65.0 & 96.3 & 92.9 & 68.6 \\
  GLAM-\={C} &59.6 & 73.0 & 75.5 & 70.4 & 86.2 & 89.5 & 86.9 &80.0 & 63.4 & 76.2 & 71.3 & 80.2 & 77.4 & 77.2 & 55.0 &94.5 & 76.1 & 78.3 & 94.4 & 93.5 & 77.9 \\
\hline
\end{tabular}
\end{table*}
Pascal VOC~\cite{VOC} with Berkeley annotations of keypoints~\cite{berkeley} contains 20 categories of instances labeled with semantic keypoint locations. Compared with Willow Object dataset~\cite{LG2M}, this dataset is considered more challenging because the instances may vary its scale, pose, illumination and the number of inlier keypoints. Following previous arts~\cite{LGM,PCA,PCA-PAMI} we split the dataset into two groups, \ie, 7,020 images for training and 1,682 for testing. In training, we take each pair of images from the same category as the training sample. For testing, we randomly select 1,000 samples from each category of the testing set to evaluate the performance of compared methods.

\begin{figure*}[htbp]
\centering
\subfigure[image pair]
{
\includegraphics[width=2.75cm]{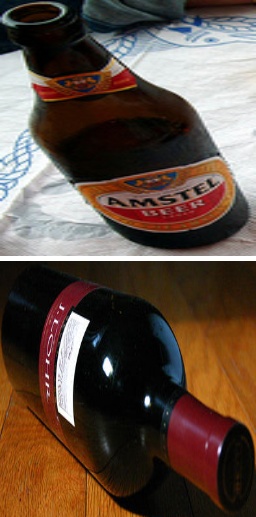}
}
\subfigure[GMN: 1/8]{
\includegraphics[width=2.75cm]{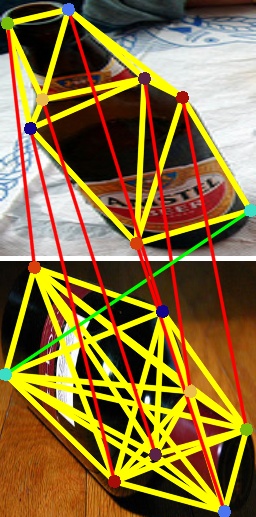}
}
\subfigure[PCA: 0/8]{
\includegraphics[width=2.75cm]{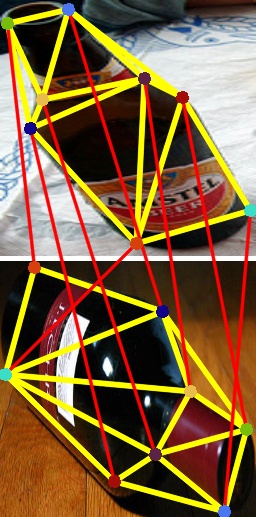}
}
\subfigure[IPCA: 0/8]{
\includegraphics[width=2.75cm]{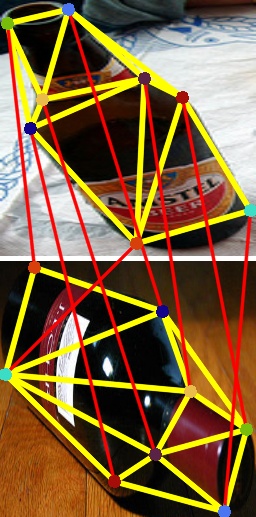}
}
\subfigure[CIE: 0/8]{
\includegraphics[width=2.75cm]{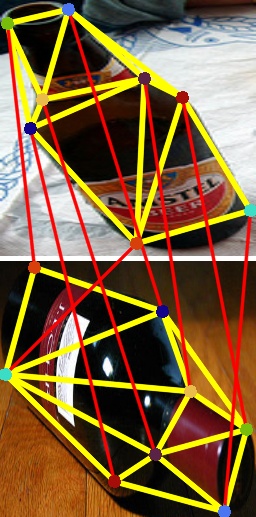}
}
\subfigure[DGMC: 0/8]{
\includegraphics[width=2.75cm]{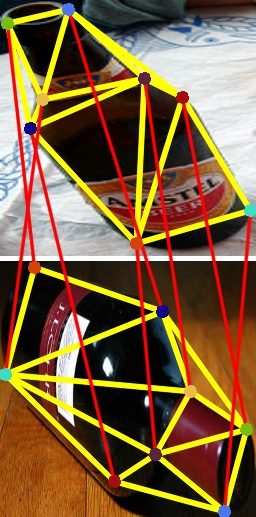}
}
\subfigure[LCS: 3/8]{
\includegraphics[width=2.75cm]{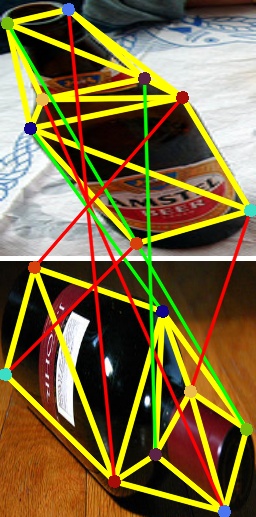}
}
\subfigure[NGM: 0/8]{
\includegraphics[width=2.75cm]{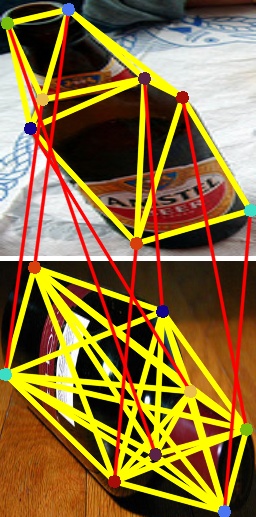}
}
\subfigure[NGMV2: 2/8]{
\includegraphics[width=2.75cm]{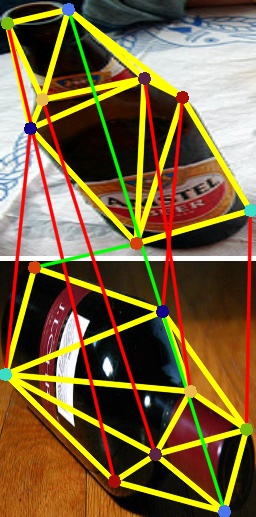}
}
\subfigure[BBGM: 2/8]{
\includegraphics[width=2.75cm]{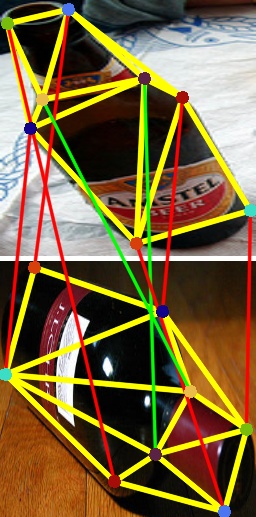}
}
\subfigure[SuperGlue: 0/8]{
\includegraphics[width=2.75cm]{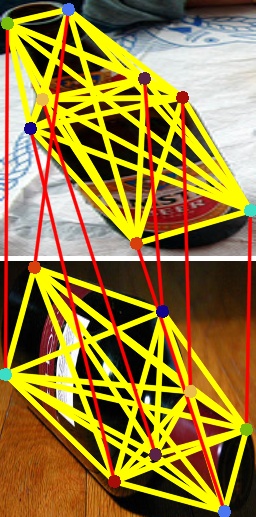}
}
\subfigure[GLAM(ours): 8/8]{
\includegraphics[width=2.75cm]{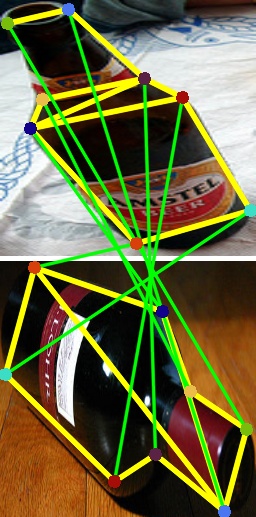}
}

\subfigure{
\includegraphics[width=15cm, height=0.6cm]{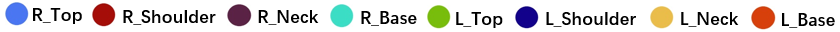}
}
\caption{The representative examples of matching solutions solved by all compared methods on category \emph{bottle}.}
\label{fig:bottle}
\end{figure*}

\begin{figure*}[htbp]
\centering
\subfigure[image pair]
{
\includegraphics[width=2.75cm]{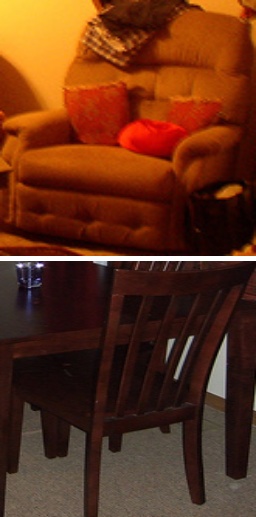}
}
\subfigure[GMN: 0/7]{
\includegraphics[width=2.75cm]{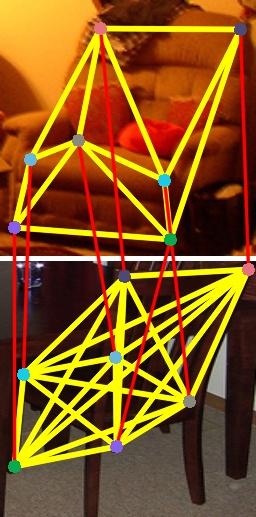}
}
\subfigure[PCA: 0/7]{
\includegraphics[width=2.75cm]{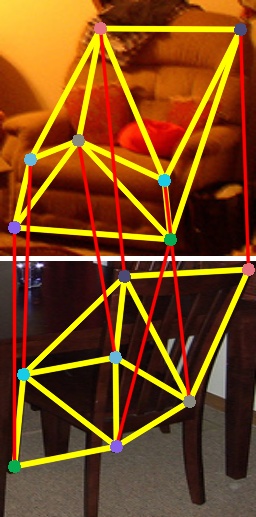}
}
\subfigure[IPCA: 3/7]{
\includegraphics[width=2.75cm]{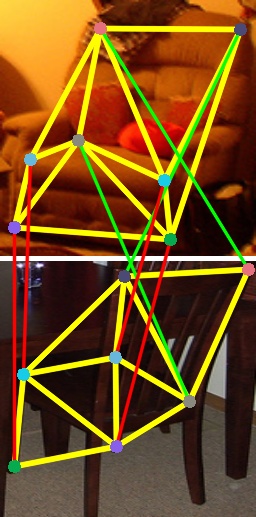}
}
\subfigure[CIE: 0/7]{
\includegraphics[width=2.75cm]{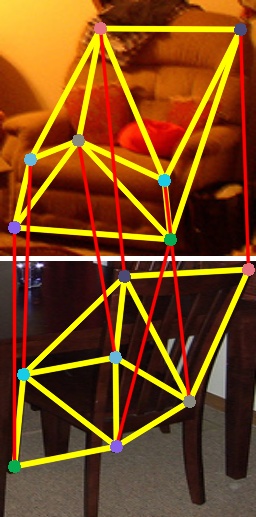}
}
\subfigure[DGMC: 3/7]{
\includegraphics[width=2.75cm]{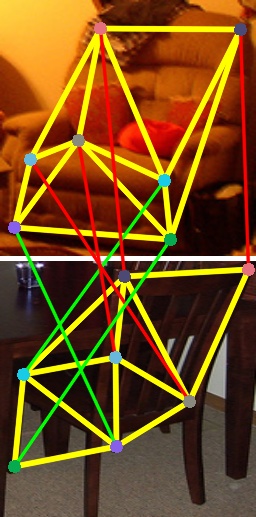}
}
\subfigure[LCS: 0/7]{
\includegraphics[width=2.75cm]{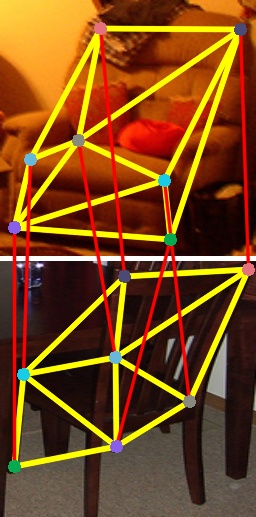}
}
\subfigure[NGM: 0/7]{
\includegraphics[width=2.75cm]{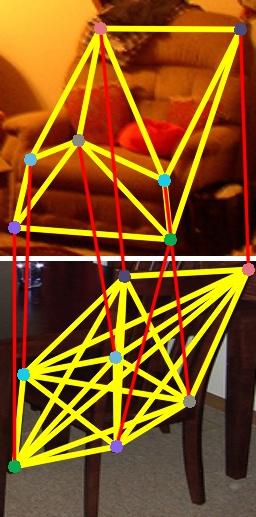}
}
\subfigure[NGMV2: 1/7]{
\includegraphics[width=2.75cm]{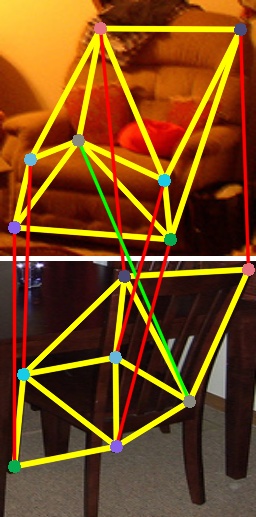}
}
\subfigure[BBGM: 1/7]{
\includegraphics[width=2.75cm]{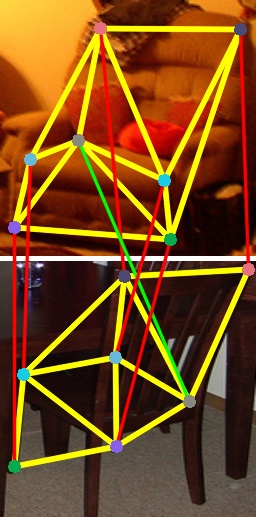}
}
\subfigure[SuperGlue: 0/7]{
\includegraphics[width=2.75cm]{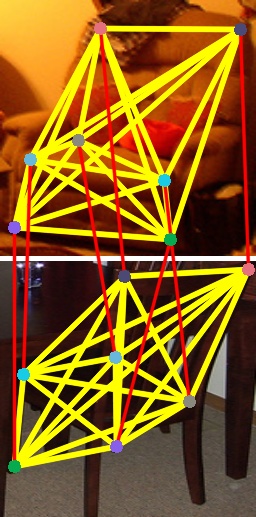}
}
\subfigure[GLAM(ours): 7/7]{
\includegraphics[width=2.75cm]{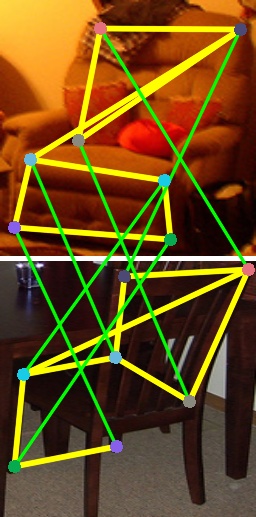}
}

\subfigure{
\includegraphics[width=15cm, height=0.7cm]{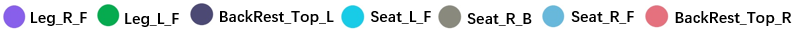}
}
\caption{The representative examples of matching solutions solved by all compared methods on category \emph{chair}.}
\label{fig:chair}
\end{figure*}

\begin{figure*}[htbp]
\centering
\subfigure[image pair]
{
\includegraphics[width=2.75cm]{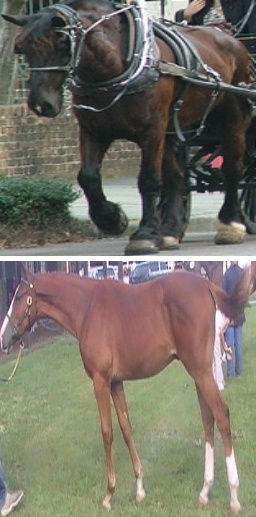}
}
\subfigure[GMN: 7/13]{
\includegraphics[width=2.75cm]{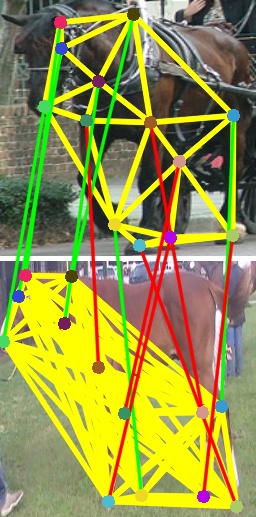}
}
\subfigure[PCA: 8/13]{
\includegraphics[width=2.75cm]{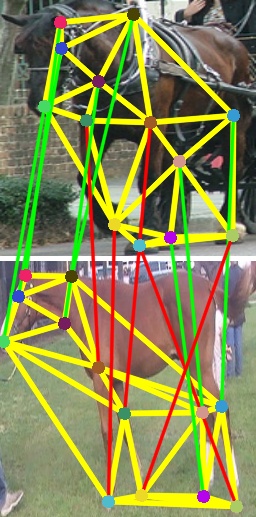}
}
\subfigure[IPCA: 8/13]{
\includegraphics[width=2.75cm]{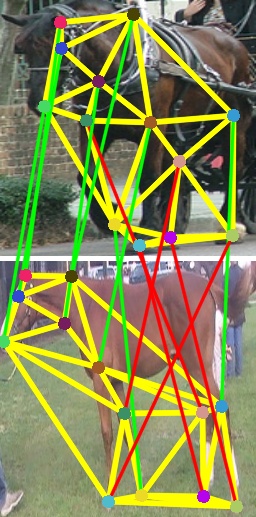}
}
\subfigure[CIE: 9/13]{
\includegraphics[width=2.75cm]{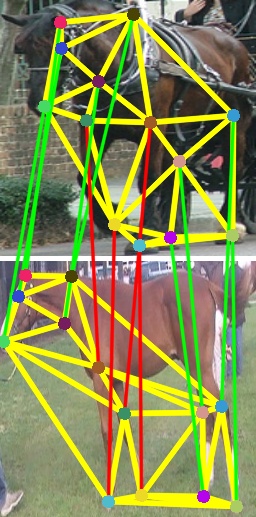}
}
\subfigure[DGMC: 8/13]{
\includegraphics[width=2.75cm]{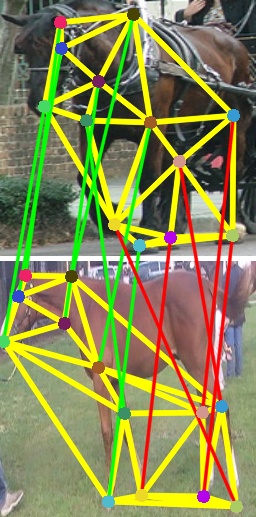}
}
\subfigure[LCS: 9/13]{
\includegraphics[width=2.75cm]{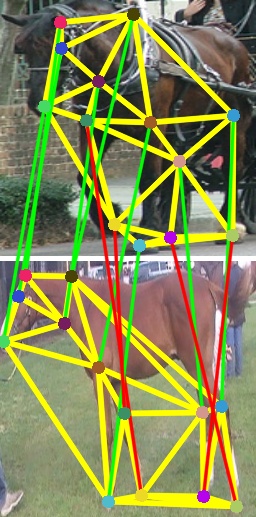}
}
\subfigure[NGM: 9/13]{
\includegraphics[width=2.75cm]{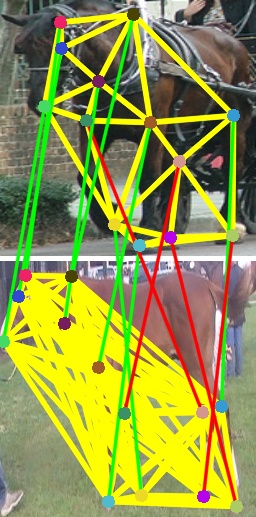}
}
\subfigure[NGMV2: 7/13]{
\includegraphics[width=2.75cm]{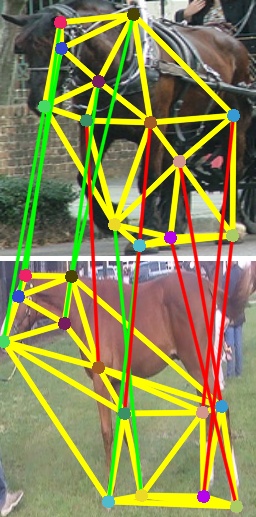}
}
\subfigure[BBGM: 7/13]{
\includegraphics[width=2.75cm]{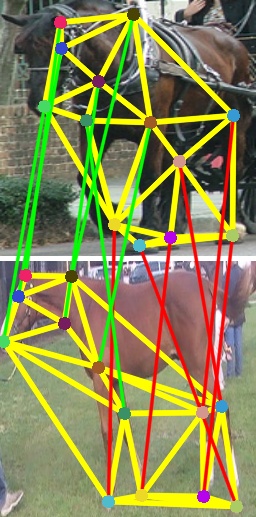}
}
\subfigure[SuperGlue: 9/13]{
\includegraphics[width=2.75cm]{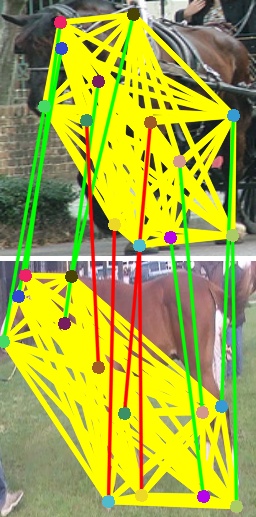}
}
\subfigure[GLAM(ours): 13/13]{
\includegraphics[width=2.75cm]{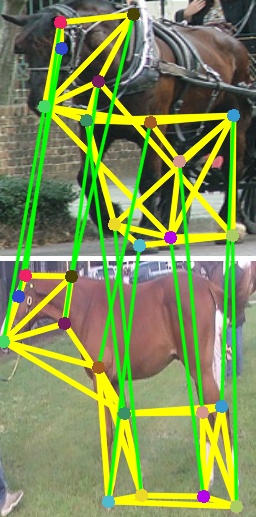}
}
\subfigure{
\includegraphics[width=14cm, height=1.1cm]{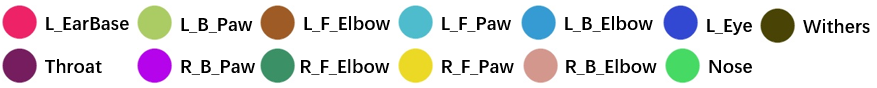}
}
\caption{The representative examples of matching solutions solved by all compared methods on category \emph{horse}.}
\label{fig:horse}
\end{figure*}

Table~\ref{table:Pascal} reports the matching accuracy of the compared methods on Pascal VOC~\cite{VOC}. Due to the challenges derived from scaling, viewpoint change and outliers, the compared baselines are unable to explore stable relationships between keypionts using heuristic strategies, and most of them fails to gain satisfied matching results on this dataset.
Through learning relationships between keypoints by the attention module, our GLAM model is able to discover more reliable structural patterns among the instances to be matched, and utilizes them for the following representation learning and the final matching decision.
As a result, our GLAM model achieves the best matching results on all categories expect \emph{boat} and \emph{tv}, and rises the average matching accuracy to 86.2\%, significantly surpassing the strongest competitors  (80.1\% for BBGM~\cite{BBGM} and NGM-v2~\cite{NGM}) by 6.1\%,



Several representative examples of graph matching selected from \emph{bottle}, \emph{chair} and \emph{horse} are shown in Figs.~\ref{fig:bottle}, \ref{fig:chair} and~\ref{fig:horse} respectively, where our method discovers more consistent and semantically meaningful graph structures and achieves obviously better matching results than the compared baseline algorithms.
Specifically, in Fig.~\ref{fig:bottle} the two bottles have severe changes in pose and viewpoint, and there are many inconsistent edges between the two graphs constructed heuristically, which contributes greatly to the failure of the baseline methods in finding correct correspondences. For our GLAM model, the edges in both of the learnt graphs cover the contour of bottles in general, which drives our model to  focus more on the affinities reflecting the semantic relationships on shapes of bottle.
Similarly, Fig~\ref{fig:chair} shows a representative example of \emph{chair} images where great changes exist in viewpoint, illumination and even the material of the chair. The graphs generated by the compared baseline algorithms fail to discover inherent relationships between keypoints, and they introduce many noise relationships, \eg, the adjacency between \emph{Leg\_R\_F} (annotated with purple nodes) and other keypoints related with \emph{Seat}. Thus, all of these graph matching methods explore less correct matches between the two chair instances.
On the contrary, the learnt graph of our model generally covers the contour of the chair instances and excludes most unreliable relationships. As a result, our model successes to find all correct correspondences between the two keypoint sets.
In the \emph{horse} images in Fig.~\ref{fig:horse}, we observe that all heuristic graphs establish many unreliable relationships between keypoints with less semantic correlations, \eg, the \emph{Ear} is adjacent with \emph{TailBase}, and the \emph{Nose} is associated with \emph{Paw}.  While, our learnt graph structures in both images exclude such unreliable relationships, and most relationships are built between the keypoints that have close semantic meanings. For example, the keypoints related to \emph{Paw} are more connected with \emph{Elbow} that has close correlations in biology, and the keypoints in the head are more likely to be adjacent with each other. Under such graph structures, our model achieves the best performance on the \emph{horse} category.

\subsubsection{SPair-71k}

\begin{table*}[th]
\small
\centering
\caption{Comparison of matching accuracy (\%) on the SPair-71k dataset. Bold \textbf{numbers} represent the best results.}
\label{table:SPair71K}
\begin{tabular}
{@{\hspace{1.mm}}l@{\hspace{.1mm}} | @{\hspace{.5mm}}c@{\hspace{.5mm}} @{\hspace{.5mm}}c@{\hspace{.5mm}} @{\hspace{.5mm}}c@{\hspace{.5mm}} @{\hspace{.5mm}}c@{\hspace{.5mm}} @{\hspace{.5mm}}c@{\hspace{.5mm}}
@{\hspace{.5mm}}c@{\hspace{.5mm}} @{\hspace{.5mm}}c@{\hspace{.5mm}} @{\hspace{.5mm}}c@{\hspace{.5mm}} @{\hspace{.5mm}}c@{\hspace{.5mm}} @{\hspace{.5mm}}c@{\hspace{.5mm}} @{\hspace{.5mm}}c@{\hspace{.5mm}} @{\hspace{.5mm}}c@{\hspace{.5mm}} @{\hspace{.5mm}}c@{\hspace{.5mm}} @{\hspace{.5mm}}c@{\hspace{.5mm}} @{\hspace{.5mm}}c@{\hspace{.5mm}} @{\hspace{.5mm}}c@{\hspace{.5mm}} @{\hspace{.5mm}}c@{\hspace{.5mm}} @{\hspace{.5mm}}c@{\hspace{.95mm}} | @{\hspace{.5mm}}c@{\hspace{1.5mm}}}
  \hline
    Algo. & aero & bike & bird & boat & bott. &bus & car &cat& chair &cow &dog & hor.& mbi. & per. &plant &she. & train& tv& AVG\\
  \hline
  GMN~\cite{DGM}& 55.4 & 49.8 & 80.1 & 58.4 & 64.0 & 71.6 & 64.0 & 68.0 & 45.2 & 68.6 & 53.9 & 50.2 & 65.3 & 46.2 & 84.0 & 57.4 & 78.3 & 80.5 & 63.4 \\
  PCA~\cite{PCA} &	64.9 & 47.4 & 79.1 & 58.1 & 64.3 & 77.0 & 65.2 & 67.3 & 48.7 & 68.9 & 54.6 & 53.8 & 65.2 & 58.1 & 91.8 & 59.6 & 73.7 & 81.7 & 65.5 \\
  IPCA~\cite{PCA-PAMI} & 63.4 & 48.4 & 80.2 & 62.5 & 66.4 & 79.2 & 68.8 & 68.2 & 54.1 & 67.5 & 62.6 & 55.4 & 61.6 & 56.8 & 88.6 & 57.9 & 77.3 & 85.4 & 66.9\\
  NGM~\cite{NGM} &60.6 & 39.5 & 65.8 & 57.8 & 57.3 & 77.6 & 69.6 & 56.1 & 51.0 & 66.4 & 57.6 & 48.8 & 63.6 & 42.7 & 76.0 & 49.9 & 71.7 & 81.4 & 60.7\\
  NGMV2~\cite{NGM}	 & 67.7 & 58.3 & 87.2 & 77.2 & 70.7 & 97.4 & 87.3 & 74.3 & 65.1 & 79.6 & 72.2 & 67.1 & 75.4 & 71.0 & \textbf{99.5} & 75.4 & 93.5 & 97.4 & 78.7\\
  QC-DGM~\cite{qcDGM} & 59.6 & 46.3 & 76.5 & 54.9 & 63.5 & 86.7 & 72.6 & 65.1 & 55.8 & 67.3 & 59.3 & 56.0 & 62.5 & 52.7 & 84.0 & 48.0 & 72.8 & 95.5 & 65.5\\
  CIE~\cite{CIE} & 64.4 & 53.4 & 82.3 & 62.5 & 69.5 & 76.8 & 67.0 & 64.5 & 57.6 & 70.2 & 57.5 & 61.8 & 67.2 & 58.7 & 88.9 & 56.1 & 73.4 & 95.7 & 68.2\\
  LCS~\cite{LGM} &57.0 & 57.8 & 82.6 & 73.5 & \textbf{72.6} & 92.0 & 75.2 & 70.2 & 61.5 & 76.3 & 60.8 & 74.2 & 74.4 & 72.2 & 98.9 & 77.9 & 86.9 & 97.8 & 75.7\\
  DGMC~\cite{DGM_consensus}  & 54.8 &44.8 &80.3 &70.9 &65.5& 90.1 &78.5 &66.7 &66.4 &73.2 &66.2 &66.5 &65.7& 59.1 &98.7 &68.5 &84.9& 98.0 &72.2   \\
  SuperGlue~\cite{SuperGlue} & 58.1 &54.2 &83.7 &80.7 &71.1 &88.6 &78.2 &68.1 &69.3 &74.9 &64.7 &69.6 &68.3 &72.0 &95.7 &71.8 &86.1& 96.3 &75.1  \\
  BBGM~\cite{BBGM}&66.9 &57.7 &85.8& 78.5 &66.9 &95.4 &86.1 &74.6 &68.3 &78.9& 73.0& 67.5& \textbf{79.3}& 73.0& 99.1 &74.8 & \textbf{95.0} &98.6 &78.9 \\
  \hline
  GLAM(ours) & \textbf{70.3} & \textbf{61.3} & \textbf{90.3} & \textbf{88.0} & 70.9 & \textbf{98.0} & \textbf{90.0} & \textbf{74.6} & \textbf{78.5} & \textbf{85.0} & \textbf{74.5} & \textbf{76.9} & 75.8 & \textbf{79.6} & 99.2 & \textbf{79.1} & 92.2 & \textbf{99.9} & \textbf{82.5}\\
  GLAM-\={S} & 52.8 & 48.6 & 80.6 & 78.7 & 69.1 & 89.0 & 80.4 & 66.5 & 63.9 & 70.1 & 50.8 & 67.3 & 70.5 & 65.7 & 98.9 & 68.2 & 82.7 & 98.2 & 72.3 \\
  GLAM-\={C} & 57.0 &54.6 &88.7 &82.2& 69.7& 89.7 &83.5 &70.6 &73.0 &75.9 &66.2 &68.9 &72.6 &68.5 & 99.4 &69.8 &88.7 &96.1 & 76.4 \\
\hline
\end{tabular}
\end{table*}

With the collection of total 70,958 pairs of images from Pascal 3D+~\cite{3DPascal} and Pascal VOC 2012~\cite{VOC2012}, the SPair-71k dataset~\cite{SPair} is very well organized with rich annotations for learning. Compared with Pascal VOC Keypoint~\cite{VOC}, the pair annotations in this dataset have more challenging and diverse variations in viewpoint, scale, truncation and occlusion, thus reflecting the more generalized visual correspondence problem in real-world. Furthermore, it removes two ambiguous and poorly annotated categories, \ie, \emph{dining table} and \emph{sofa}, from the Pascal VOC dataset.

In Table~\ref{table:SPair71K} we report the matching results of the proposed GLAM method on the SPair-71k dataset in comparison with  state-of-the-art graph matching methods.  As reported, our GLAM model achieves the best matching results on most of the 18 categories expect \emph{bottle}, \emph{motorbike}, \emph{plant} and \emph{train}, and rises the average matching accuracy to 82.5\%, surpassing the strongest competitor (78.9\% for BBGM) by 3.6\%. It is demonstrated  that our model have better generalization ability in more difficult visual matching problems in real-world.


\subsection{Ablation Studies}

\subsubsection{The Number of Attention Layers }
The number of attention layers determines the iterations of feature updating, and has direct influence on the performance of our model. To explore the impact of the attention layer iterations on the matching accuracy and time consumption, we vary the number of attention layers from 1 to 6, and test the proposed model on Pascal VOC~\cite{VOC}.

\begin{figure}[h]
\centering
  \subfigure[]{
    \label{subfig:iter_acc}
    \includegraphics[width=7cm,height=4.cm]{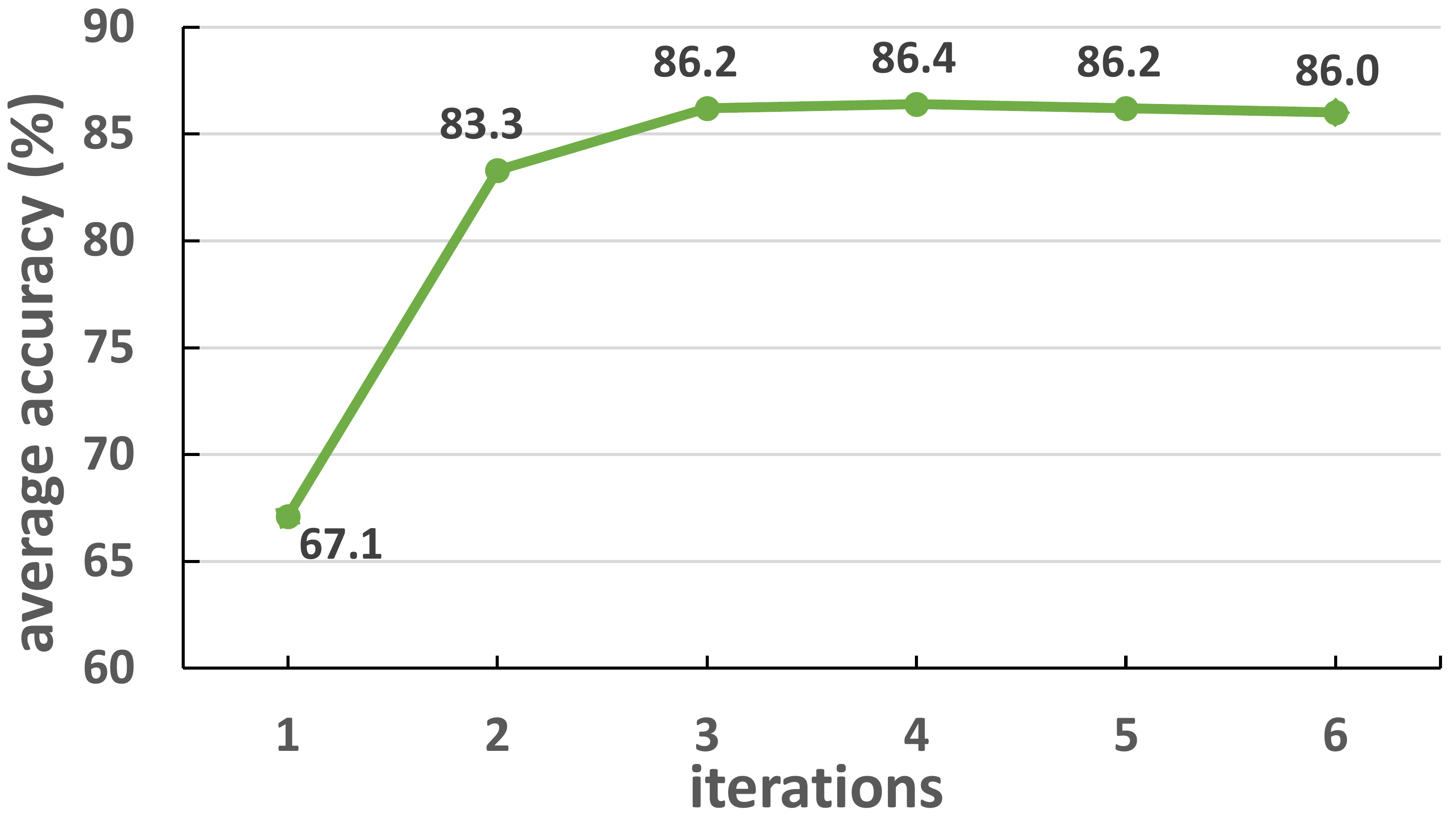}
    }
  \subfigure[]{
    \label{subfig:iter_time}
    \includegraphics[width=7cm,height=4.cm]{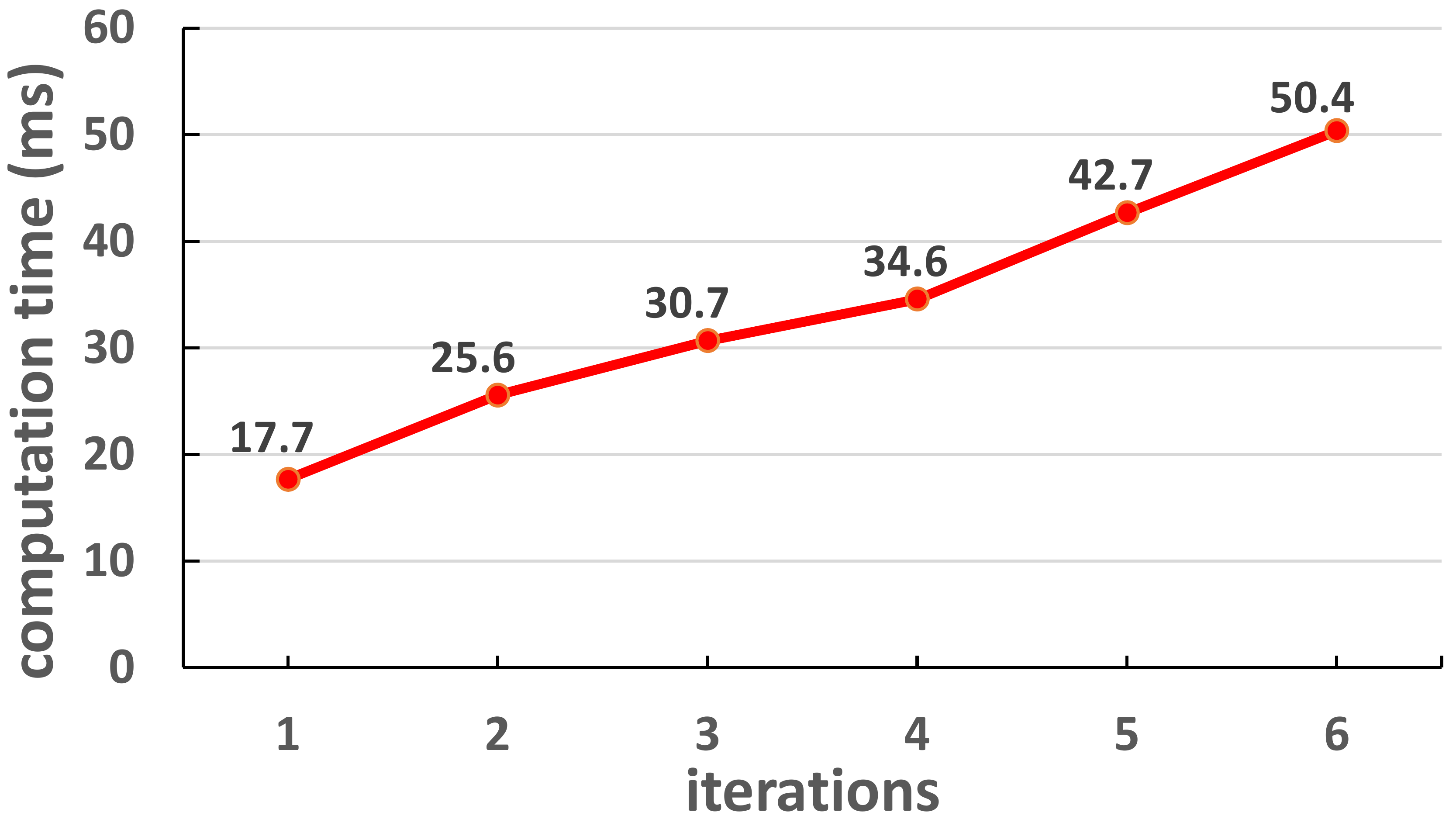}
    }
  \caption{(a) The average matching accuracy under varying attention iterations; (b) The computational time under varying attention iterations.}
  \label{fig:examples}
  \vspace{-0.3cm}
\end{figure}

As expected, the computation time grows almost linearly with the number of attention layers increasing, which has been illustrated in Fig.~\ref{subfig:iter_time}.
As shown in Fig.~\ref{subfig:iter_acc}, the average matching accuracy is only 67.1\% when only a single attention layer is used in our model.
The matching accuracy rises rapidly to 83.3\% once we stack two attention layers in our model, and it reaches saturation when the number of attention layers is large than 3.
As a result, we set the number of attention layers to 3 throughout our experiments to balance the matching accuracy and the time consumption.


\subsubsection{Attention Module}
To explore the contributions of different attention modules in our model, we design two downgraded version of our framework, named GLAM-\={S} and GLAM-\={C}, where \={S} and \={C} denote removing the SAL module and the CAL module from GLAM, respectively.

As reported in Table~\ref{table:willow}, both GLAM-\={S} and GLAM-\={C} achieve comparable results with GLAM on the Willow Object dataset, and they outperform all compared baseline algorithms. It indicates that using only one of the proposed attention layer is able to produce discriminative representations for keypoints in these relatively simple scenarios.
However, on the more difficult datasets, Pascal VOC (Table~\ref{table:Pascal}) and SPair-71k (Table~\ref{table:SPair71K}), the performance of our framework deteriorates remarkably once one of the proposed attention modules is removed.
Specifically, the average matching accuracy declines by 17.6\% and 8.3\% on Pascal VOC when dropping the SAL module and the CAL module, respectively.
As for SPair-71k, removing of the SAL module and the CAL module results in the decrease of matching accuracy by 10.2\% and 6.1\%, respectively.

As illustrated, both self-attention and cross-attention modules contribute significantly in our framework, and the combination of them help us to better learn discriminative representations for keypoint matching.

\section{Experiments on Graph Learning}\label{sec:expe:learning}

The main motivation of this paper lies in learning graph patterns, instead of utilizing handcrafted graph structures as in previous arts, to boost graph matching. In order to further illustrate the learnt graph patterns and explore their benefits to graph matching, in this section we first visualize learnt graph patterns of each category on the Pascal VOC dataset, and then report the comparison of the baseline algorithms by replacing the handcrafted graph structures in these algorithms with our learnt graph patterns.

\subsection{Learnt graph patterns} \label{sec:learnt_pattern}

\begin{table*}[t]
\centering
\caption{The effectiveness of learnt adjacency (LA) and manual semantic adjacency (MA) for graph matching on the Pascal VOC dataset. }
\label{table:MA_LA}
\begin{tabular}
{@{\hspace{.1mm}}c@{\hspace{.2mm}} | @{\hspace{.75mm}}c@{\hspace{.5mm}} @{\hspace{.5mm}}c@{\hspace{.5mm}} @{\hspace{.5mm}}c@{\hspace{.5mm}} @{\hspace{.5mm}}c@{\hspace{.5mm}} @{\hspace{.5mm}}c@{\hspace{.5mm}}
@{\hspace{.5mm}}c@{\hspace{.5mm}} @{\hspace{.5mm}}c@{\hspace{.5mm}} @{\hspace{.5mm}}c@{\hspace{.5mm}} @{\hspace{.5mm}}c@{\hspace{.5mm}} @{\hspace{.5mm}}c@{\hspace{.5mm}} @{\hspace{.5mm}}c@{\hspace{.5mm}} @{\hspace{.5mm}}c@{\hspace{.5mm}} @{\hspace{.5mm}}c@{\hspace{.5mm}} @{\hspace{.5mm}}c@{\hspace{.5mm}} @{\hspace{.5mm}}c@{\hspace{.5mm}} @{\hspace{.5mm}}c@{\hspace{.5mm}} @{\hspace{.5mm}}c@{\hspace{.5mm}} @{\hspace{.5mm}}c@{\hspace{.5mm}} @{\hspace{.5mm}}c@{\hspace{.5mm}} @{\hspace{.5mm}}c@{\hspace{.5mm}}
| @{\hspace{.65mm}}c@{\hspace{.5mm}}}
  \hline
    Algo. & aero & bike & bird & boat & bot. &bus & car &cat& cha. &cow & tab. &dog & hor.& mbi. & per. &pla. &she. &sofa& tra. & tv& AVG\\
  \hline
  GMN~\cite{DGM} & 31.9 &47.2& 51.9& 40.8 &68.7 &72.2 &53.6 &52.8 &34.6 &48.6 &72.3 &47.7 &54.8 &51.0 &38.6 &75.1 &49.5 &45.0 &83.0 &86.3 &55.3  \\
  GMN~\cite{DGM}+MA & 58.3 & 76.4 & 70.3 & 50.6 & 83.6 & 75.1 & 79.0 & 71.6 & 51.0 & 73.6 & 74.5 & 68.1 & 73.4 & 74.4 & 58.3 & 82.1 & 75.2 & 57.3 & 93.0 & 93.7 & 72.0  \\
  GMN~\cite{DGM}+LA & 58.5 & 77.4 & 71.6 & 52.5 & 84.7 & 80.3 & 79.2 & 77.4 & 59.0 & 82.4 & 73.6 & 73.6 & 79.4 & 70.2 & 49.5 & 84.0 & 80.7 & 54.7 & 94.6 & 93.8 & 73.9 \\
  \hline
  PCA~\cite{PCA}& 51.2 &61.3& 61.6 &58.4 &78.8 &73.9 &68.5 &71.1 &40.1 &63.3 &45.1 &64.4 &66.4 &62.2 &45.1 &79.1 &68.4 &60.0 &80.3 &91.9 &64.6 \\
  PCA~\cite{PCA}+MA& 62.8 & 70.3 & 75.5 & 58.0 & 81.1 & 80.6 & 84.3 & 79.0 & 55.2 & 77.2 & 59.6 & 77.6 & 80.3 & 72.1 & 69.2 & 82.6 & 79.2 & 66.6 & 89.1 & 92.2 & 74.6 \\
  PCA~\cite{PCA}+LA& 61.2 & 70.2 & 73.8 & 64.0 & 82.4 & 82.3 & 82.5 & 78.0 & 58.8 & 79.2 & 72.9 & 76.7 & 80.1 & 73.8 & 49.9 & 88.7 & 81.5 & 71.4 & 88.9 & 92.3 & 75.4 \\
  \hline
  IPCA~\cite{PCA-PAMI} &51.0 &64.9 &68.4 &60.5 &80.2 &74.7 &71.0 &73.5 &42.2 &68.5 &48.9 &69.3 &67.6 &64.8 &48.6 &84.2 &69.8 &62.0 &79.3 &89.3 &66.9 \\
  IPCA~\cite{PCA-PAMI}+MA &63.1 & 73.9 & 76.3 & 65.0 & 82.6 & 78.7 & 84.8 & 78.1 & 50.2 & 77.0 & 67.7 & 76.6 & 76.6 & 73.9 & 70.0 & 84.2 & 77.0 & 73.1 & 81.1 & 91.4 & 75.1\\
  IPCA~\cite{PCA-PAMI}+LA &60.2 & 70.0 & 71.6 & 67.3 & 83.1 & 78.9 & 82.8 & 75.1 & 51.3 & 75.2 & 71.3 & 75.1 & 73.8 & 73.4 & 50.6 & 89.5 & 75.7 & 66.0 & 81.7 & 91.4 & 73.2\\
  \hline
  qc-DGM~\cite{qcDGM} & 49.6 &64.6 &67.1& 62.4 &82.1& 79.9& 74.8& 73.5& 43.0& 68.4 &66.5 &67.2 &71.4 &70.1& 48.6 &92.4 &69.2 &70.9 &90.9& 92.0 &70.3\\
  qc-DGM~\cite{qcDGM}+MA & 56.7 &74.0& 70.0 &62.5& 84.1 &81.9 &79.4 &79.3 &50.6 &80.7& 77.8 & 75.0 & 81.7 & 73.5 & 61.6 & 90.4 & 78.6 & 74.9 & 90.0& 93.1& 75.8\\
  qc-DGM~\cite{qcDGM}+LA &61.9 & 78.6 &69.1 &62.0 &84.5 &82.5  &79.5 &80.2 &54.5  &83.1& 86.7 &77.4 & 83.3 & 75.8& 53.3& 92.9 &81.1 &67.3& 89.1& 93.8 &76.8 \\
  \hline
  CIE~\cite{CIE} & 51.2 &69.2 &70.1 &55.0 &82.8& 72.8& 69.0& 74.2& 39.6 &68.8 &71.8 &70.0 &71.8 &66.8 &44.8 &85.2 &69.9& 65.4 &85.2 &92.4 & 68.9 \\
  CIE~\cite{CIE}+MA& 58.0 & 63.4 & 69.6 & 52.4 & 82.2 & 75.6 & 75.1 & 74.2 & 44.3 & 70.4 & 76.2 & 71.3 & 68.6 & 65.1 & 54.7 & 75.8 & 71.0 & 58.5 & 90.1 & 92.7 & 69.5  \\
  CIE~\cite{CIE}+LA& 54.4 & 66.7 & 72.1 & 53.6 & 80.5 & 74.1 & 73.8 & 73.5 & 44.8 & 70.8 & 79.0 & 72.4 & 70.0 & 62.6 & 49.0 & 74.3 & 70.3 & 61.8 & 84.1 & 92.7 & 69.0  \\
  \hline
  BBGM~\cite{BBGM} &61.5 &75.0 & 78.1 & 80.0 & 87.4 & 93.0 & 89.1 & 80.2 &58.1 &77.6 &76.5 &79.3 &78.6 &78.8 &66.7 &97.4 &76.4 &77.5 &97.7 &94.4 &80.1  \\
  BBGM~\cite{BBGM}+MA&73.5 & 75.2 & 83.2 & 76.3 & 87.6 & 91.6 & 90.1 & 85.5 & 61.4 & 85.0 & 72.8 & 84.6 & 79.7 & 80.7 & 74.6 & 99.2 & 82.4 & 82.9 & 96.0 & 94.9 & 82.8  \\
  BBGM~\cite{BBGM}+LA&70.1 & 75.8 & 84.2 & 82.4 & 86.0 & 90.0 & 90.7 & 84.9 & 65.6 & 82.9 & 81.6 & 82.9 & 80.3 & 78.9 & 62.7 & 98.2 & 81.2 & 78.6 & 97.2 & 95.1 & 82.5 \\
  \hline
  NGM~\cite{NGM} &50.1 &63.5 &57.9 &53.4 &79.8 &77.1 &73.6& 68.2& 41.1& 66.4 &40.8 &60.3 &61.9& 63.5& 45.6 &77.1 &69.3 &65.5 &79.2 &88.2 &64.1 \\
  NGM~\cite{NGM}+MA &65.4 & 75.0 & 72.8 & 54.4 & 82.3 & 80.9 & 85.9 & 78.1 & 55.7 & 76.6 & 53.5 & 72.5 & 76.2 & 75.8 & 60.5 & 86.8 & 73.8 & 79.5 & 87.0 & 91.9 & 74.2 \\
  NGM~\cite{NGM}+LA &63.1 & 75.6 & 70.8 & 62.6 & 83.0 & 85.1 & 85.3 & 77.5 & 59.9 & 82.0 & 49.4 & 75.5 & 79.2 & 76.1 & 52.8 & 89.4 & 80.1 & 75.5 & 81.5 & 92.0 & 74.8 \\
  \hline
  NGMv2~\cite{NGM} &61.8 &71.2 &77.6 &78.8 &87.3 &93.6 & 87.7 &79.8 &55.4 & 77.8 & 89.5 &78.8 & 80.1 &79.2 & 62.6 & 97.7 & 77.7 & 75.7 & 96.7 & 93.2 & 80.1\\
  NGMv2~\cite{NGM}+MA& 65.9 & 74.6 & 78.4 & 72.2 & 87.5 & 89.3 & 88.3 & 80.6 & 61.6 & 83.6 & 87.2 & 78.8 & 78.8 & 75.5 & 65.4 & 98.2 & 78.5 & 72.9 & 99.3 & 92.8 & 80.5\\
  NGMv2~\cite{NGM}+LA& 62.9 & 76.9 & 77.1 & 80.6 & 86.8 & 88.2 & 87.5 & 83.5 & 68.2 & 84.2 & 79.2 & 78.9 & 82.4 & 79.5 & 60.9 & 96.3 & 80.9 & 72.0 & 98.3 & 93.1 & 80.9 \\
\hline
\end{tabular}
\end{table*}

To explore the learnt graph patterns, we take the average self-attention dependencies output from the last updating iteration as the final learnt adjacency matrices, and for each category we average the matrices in this category as its graph pattern.

For obtaining the learnt adjacency matrix, we randomly choose 2,000 pairs of images from each category in the training set and pass them to our trained network. The output matrices $\bar{M}$ (\eg,~\ref{eq:self_attention_M}) of the self-attention module in the last attention layer are counted for the learnt adjacency matrix of the graph. Considering there are $N_S$ attention heads in the self-attention module, we take the average of all attention heads as the weighted adjacency matrix:
\begin{equation}
A = \frac{1}{N_S}\sum_{i=1}^{N_S} \bar{M}_{(i)}.
\end{equation}
Note that, on the Pascal VOC dataset the images in the same category may contain different numbers of labeled keypoints, so we expand the computed weighted adjacency matrix to the maximum size by adding dummy rows and columns for absent keypoints.
Subsequently, the computed weight adjacency matrices of all chosen images in the same category are averaged as the learnt graph pattern of this category.

\begin{figure}[htbp]
\centering
\subfigure[person]{
\includegraphics[width=4.0cm]{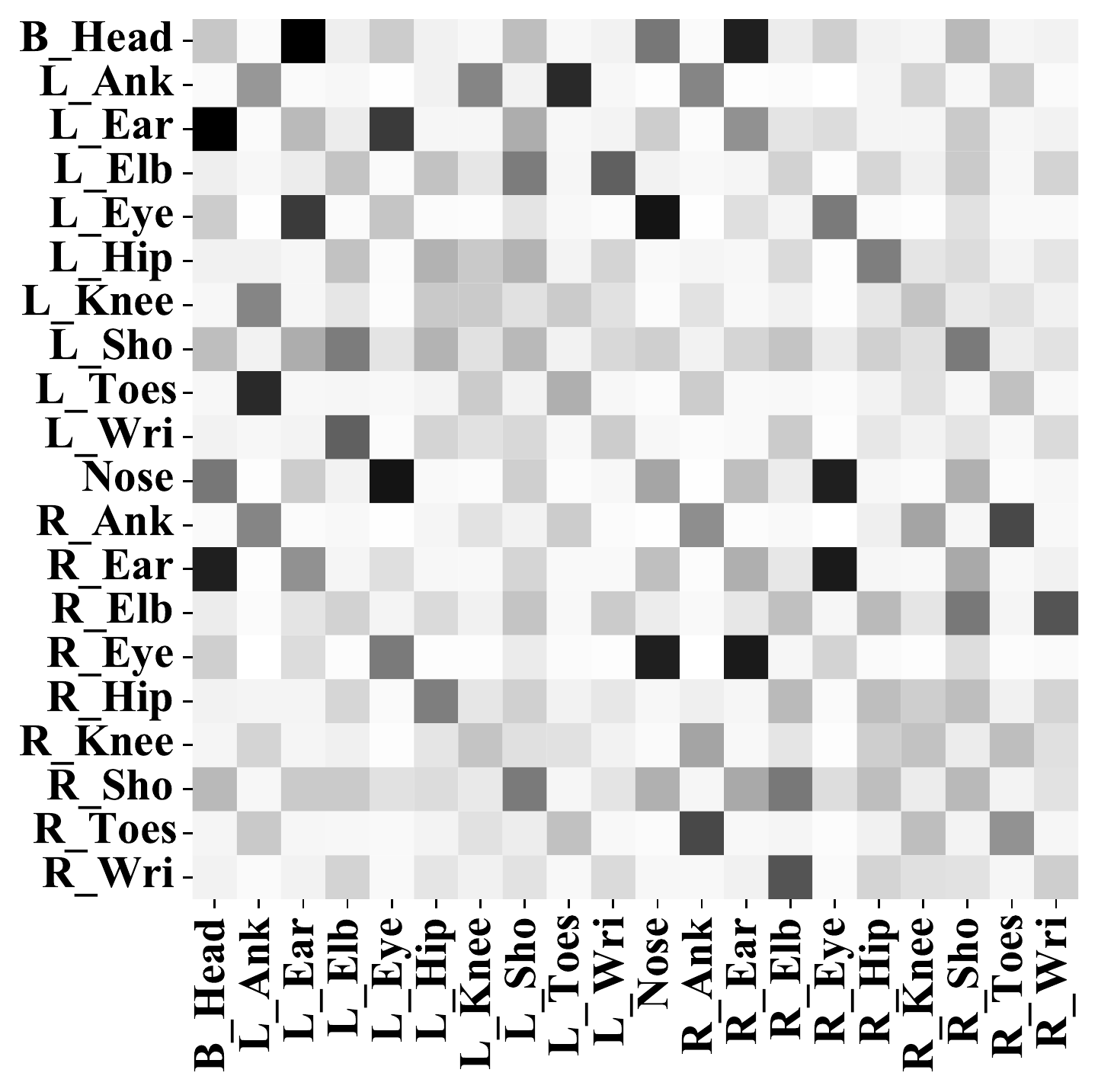}
}
\quad
\subfigure[aeroplane]{
\includegraphics[width=4.0cm]{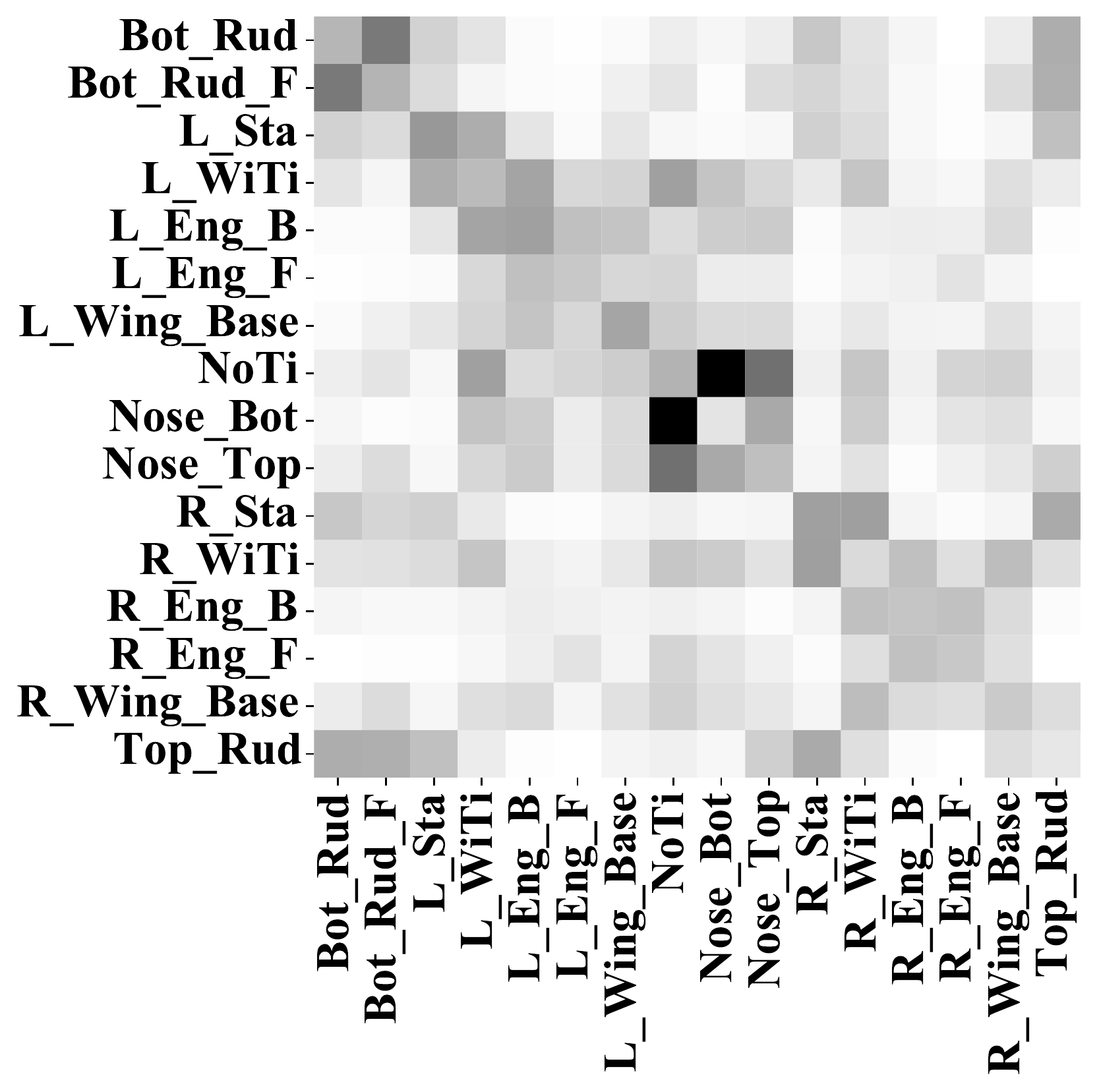}
}
\quad
\subfigure[bird]{
\includegraphics[width=4cm]{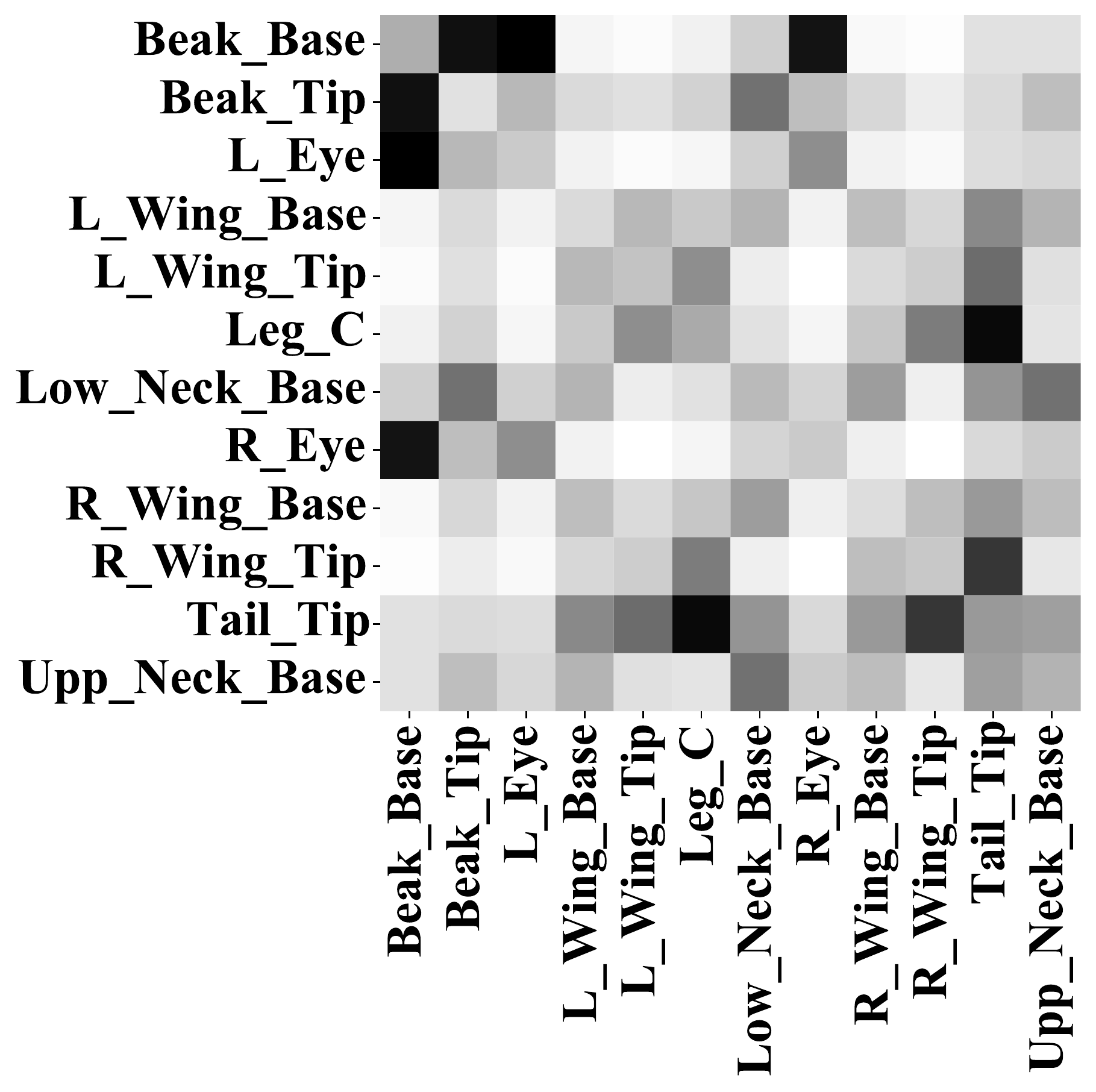}
}
\quad
\subfigure[car]{
\includegraphics[width=4cm]{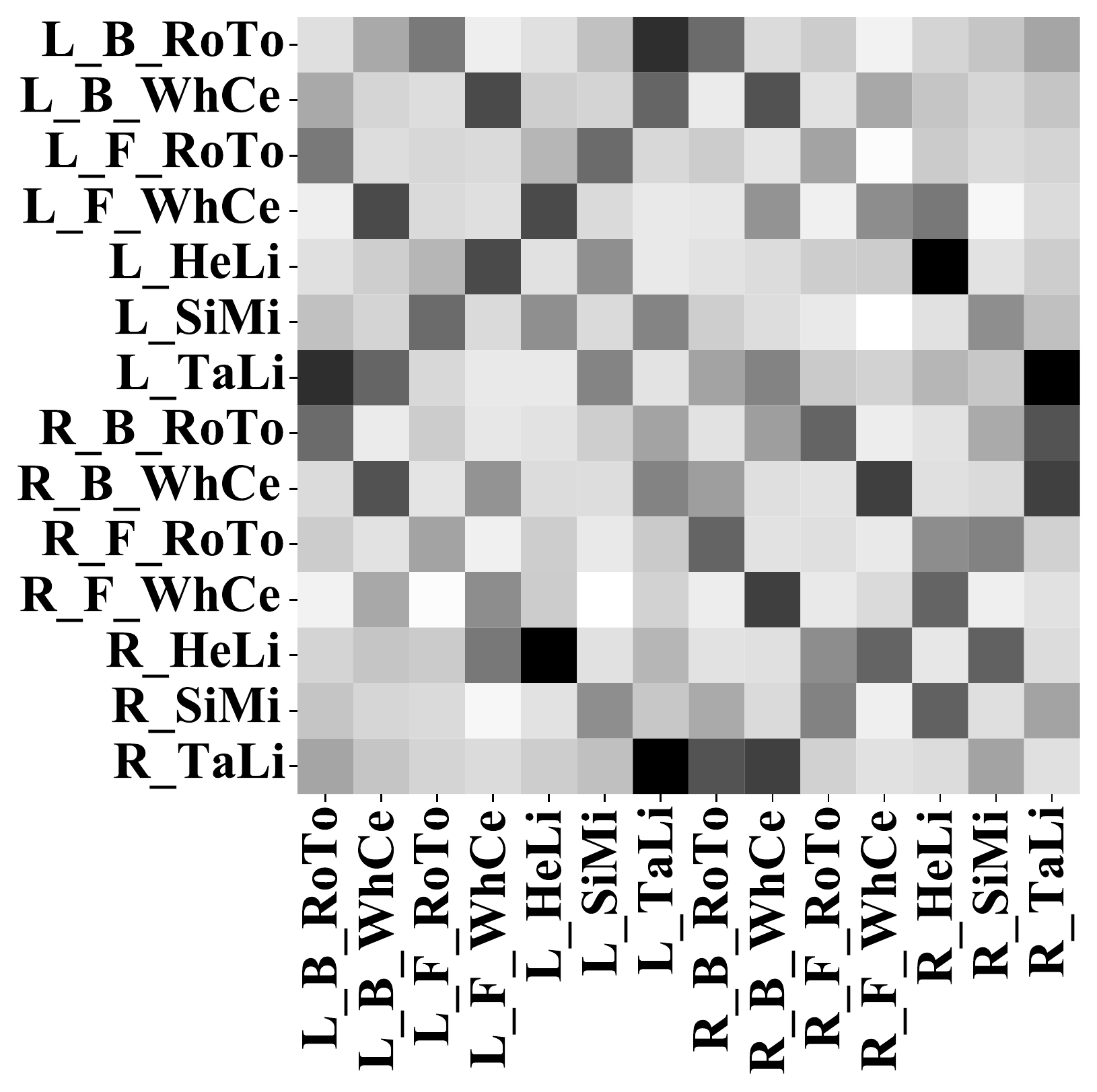}
}
\quad
\subfigure[cow]{
\includegraphics[width=4cm]{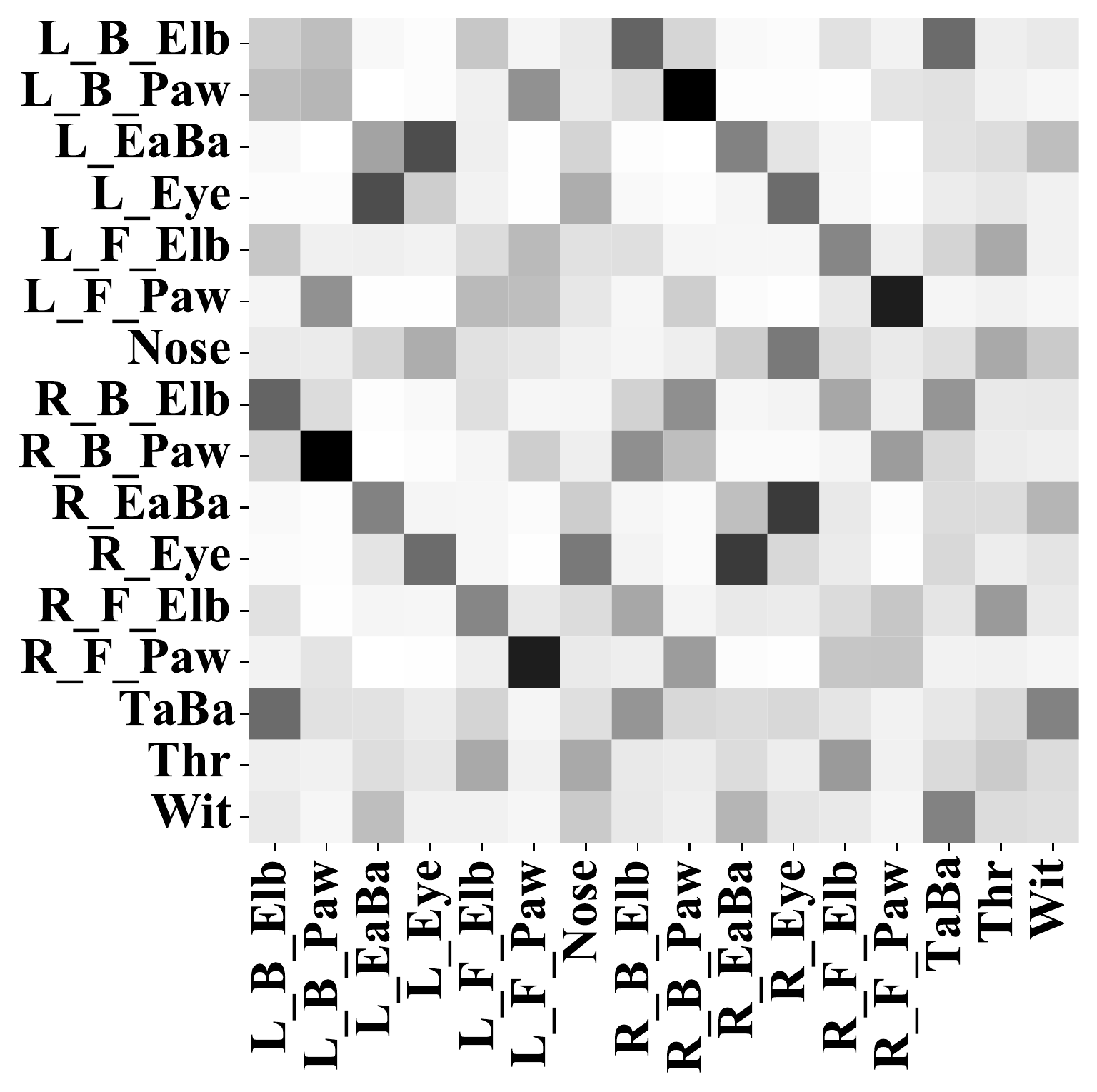}
}
\quad
\subfigure[dog]{
\includegraphics[width=4cm]{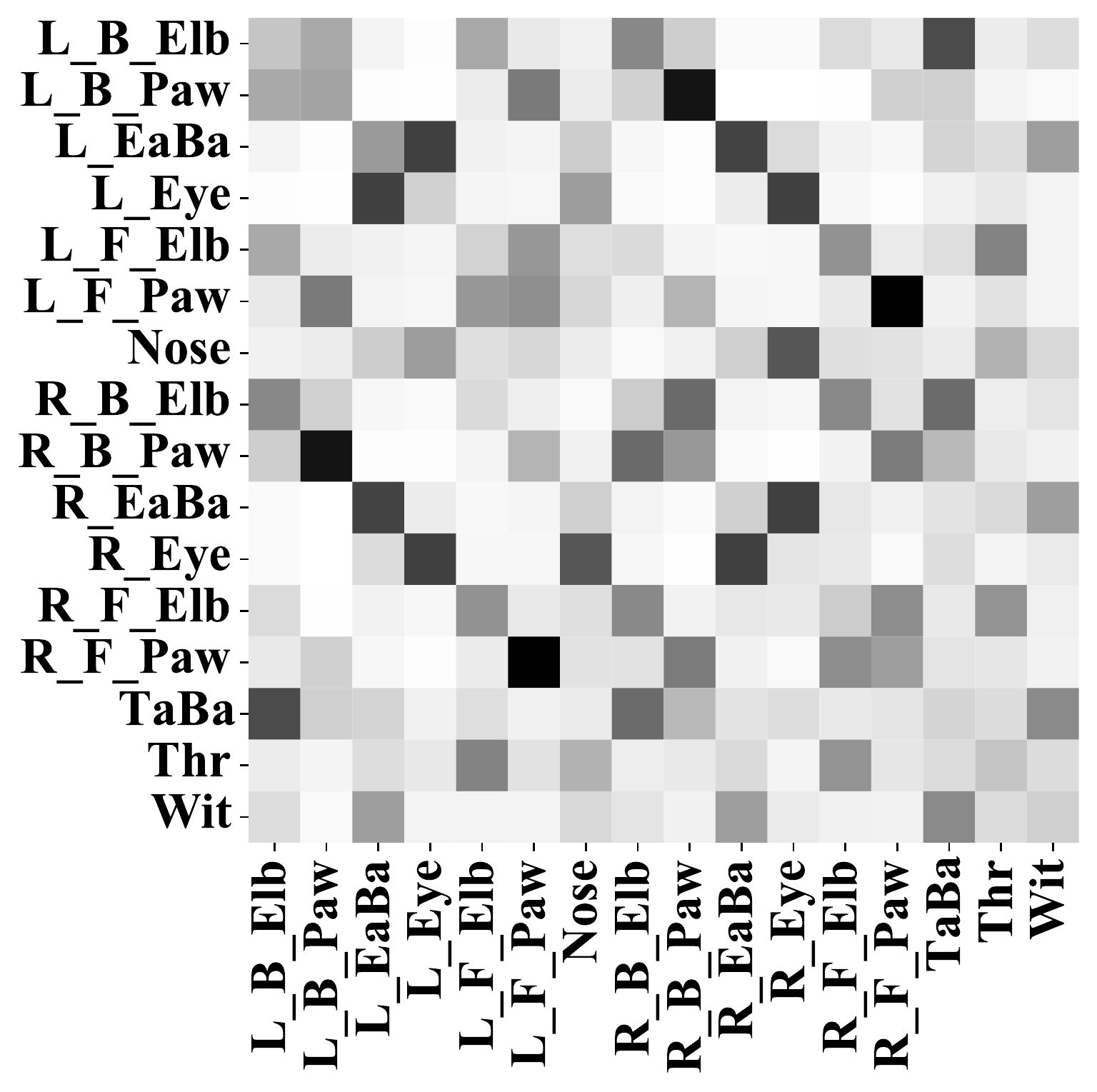}
}
\quad
\subfigure[cat]{
\includegraphics[width=4cm]{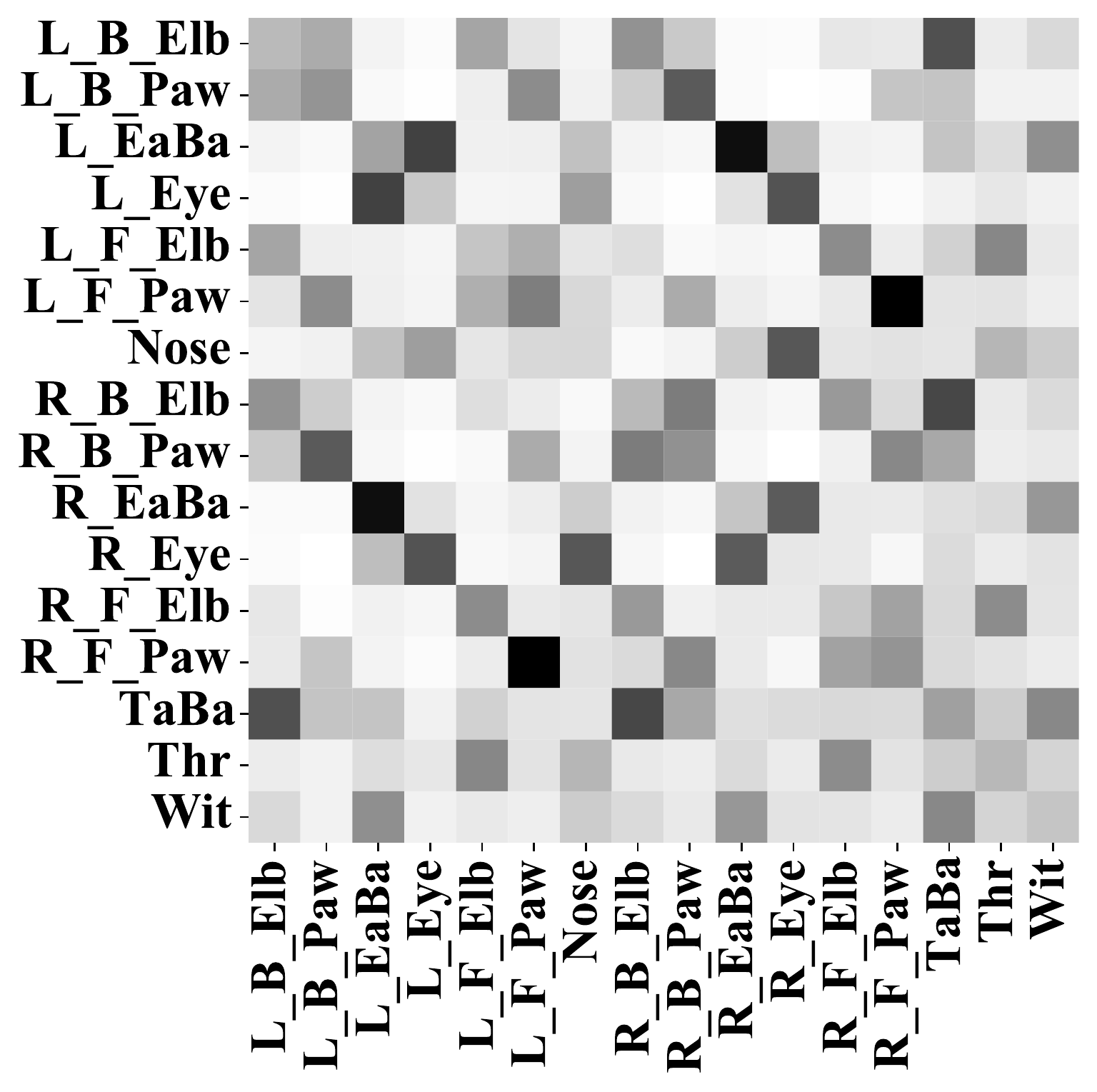}
}
\quad
\subfigure[horse]{
\includegraphics[width=4cm]{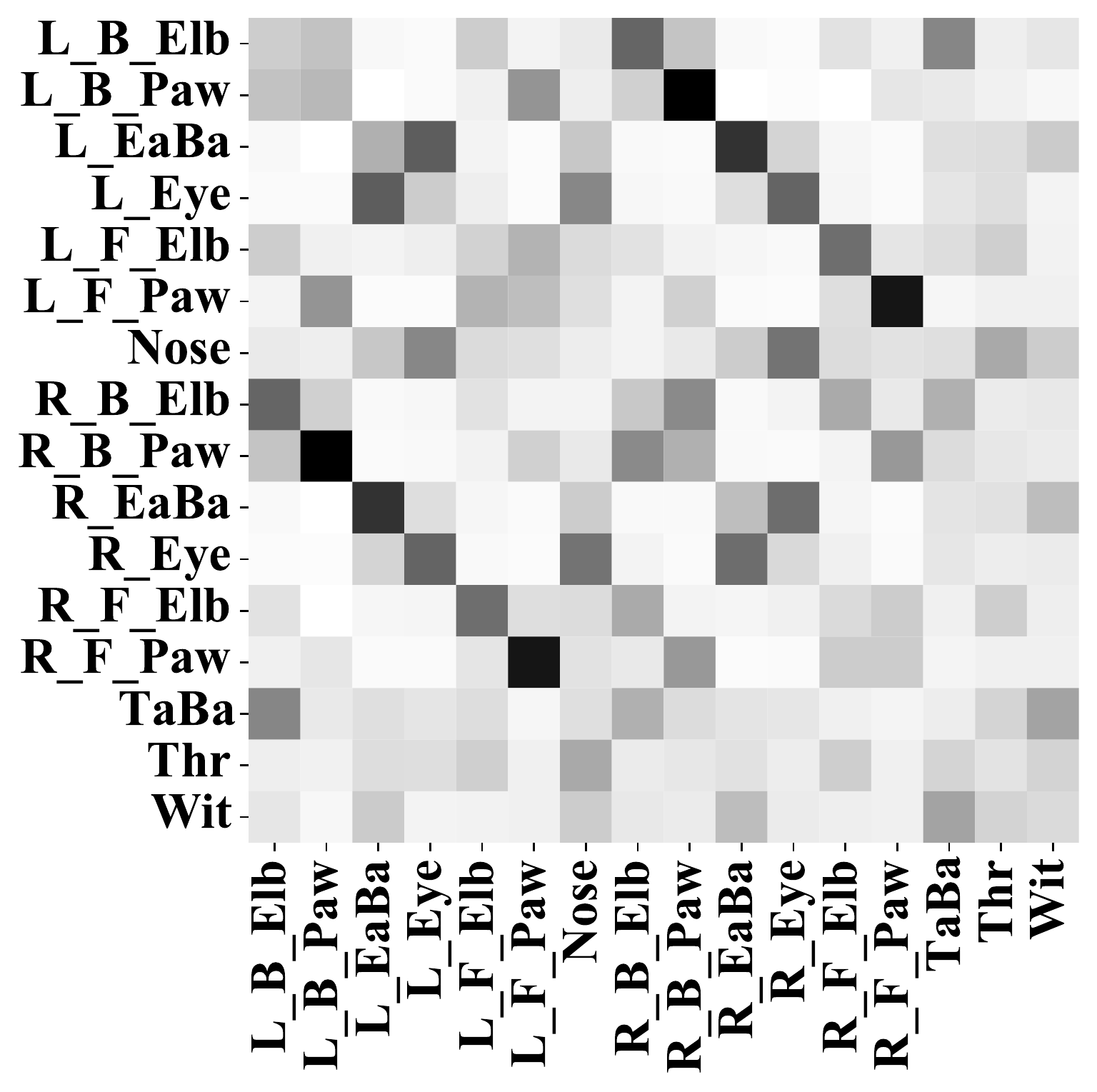}
}
\quad
\subfigure[sheep]{
\includegraphics[width=4cm]{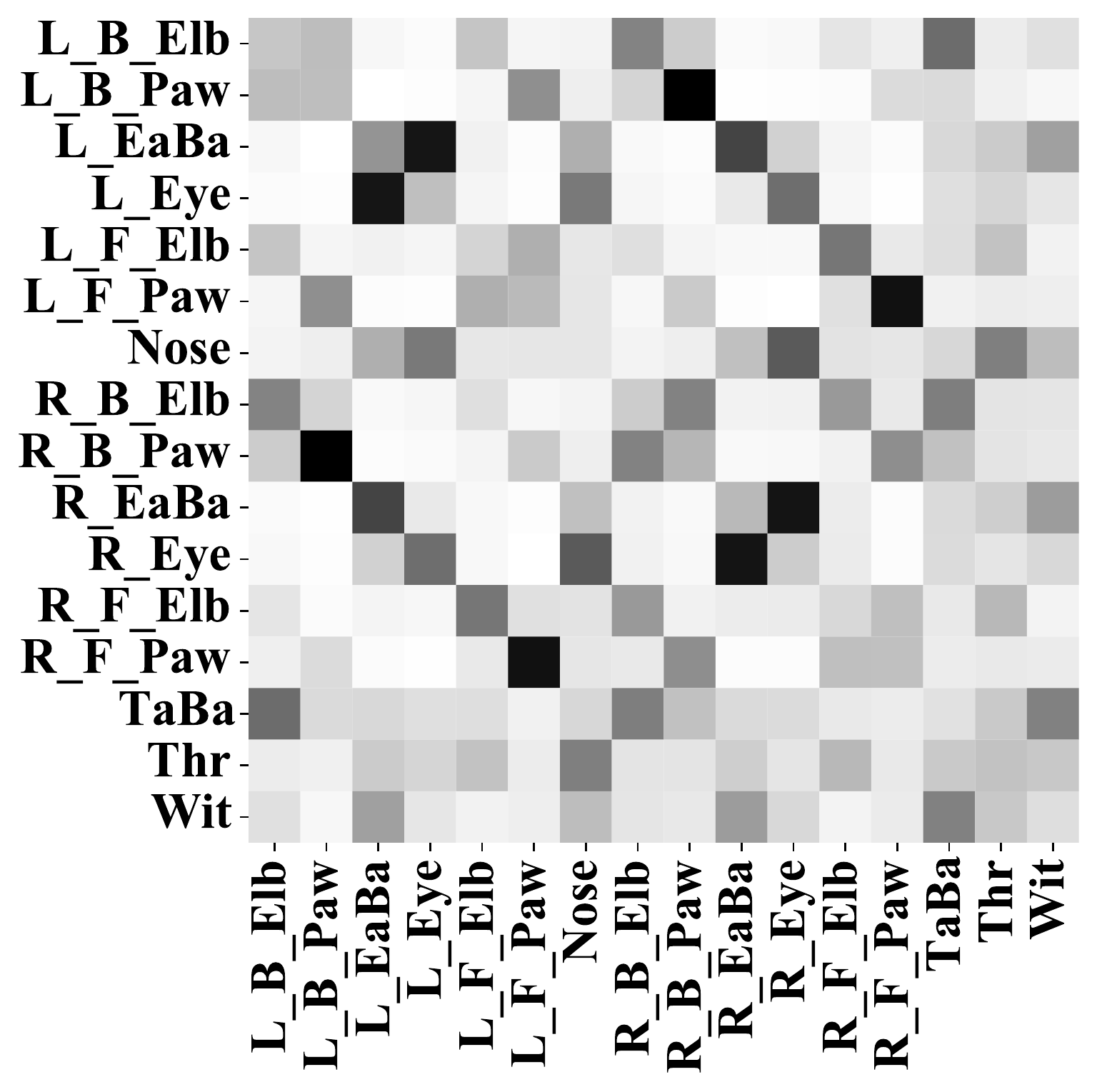}
}
\quad
\subfigure[sofa]{
\includegraphics[width=4cm]{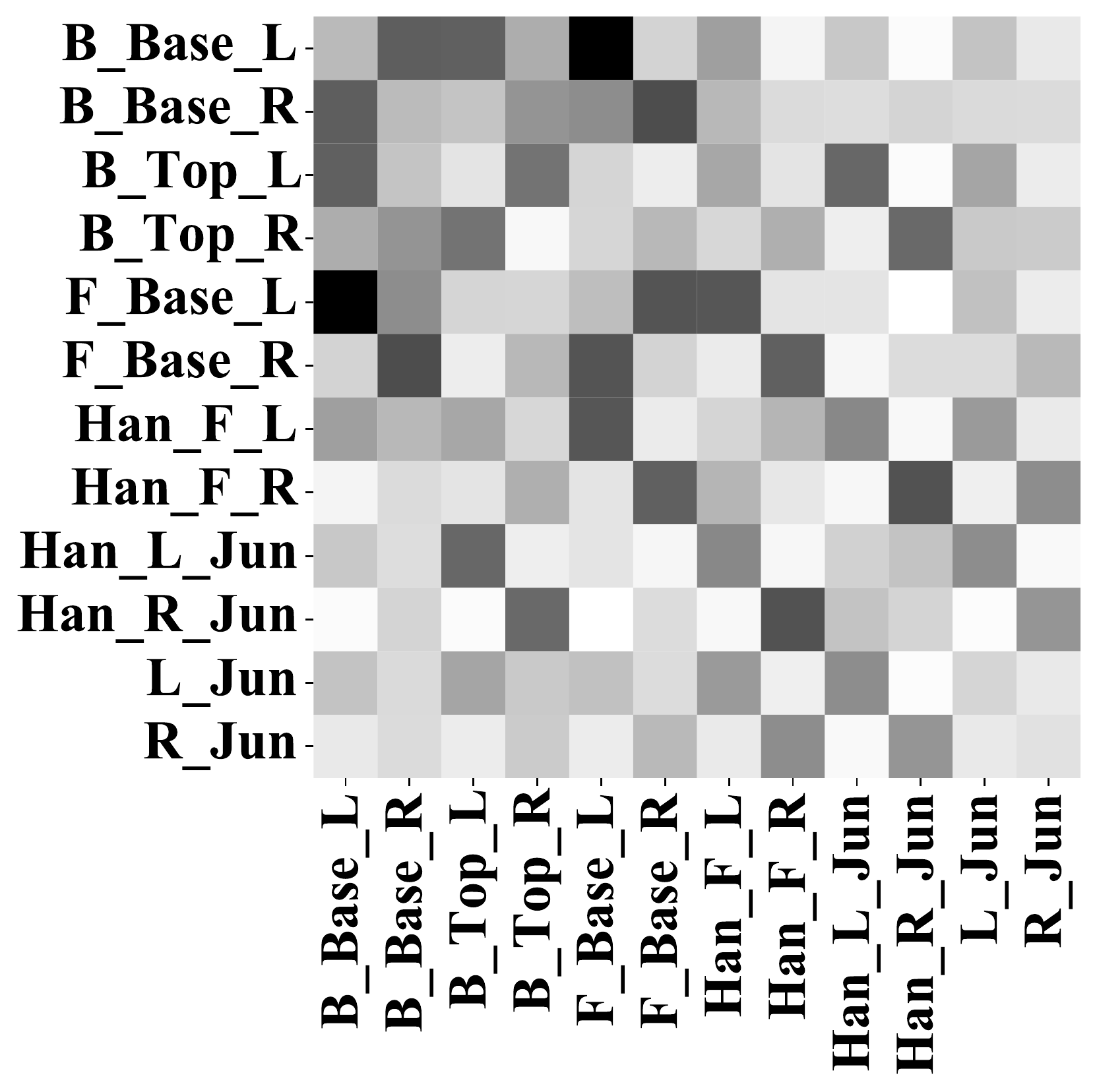}
}
\caption{The learnt adjacency matrix of the categories with more annotations. The darker the element, the larger the value, the lighter the element, the smaller the value.}
\label{fig:hmDiff}
\end{figure}

\begin{figure}[htbp]
\centering
\subfigure[bicycle]{
\includegraphics[width=4cm]{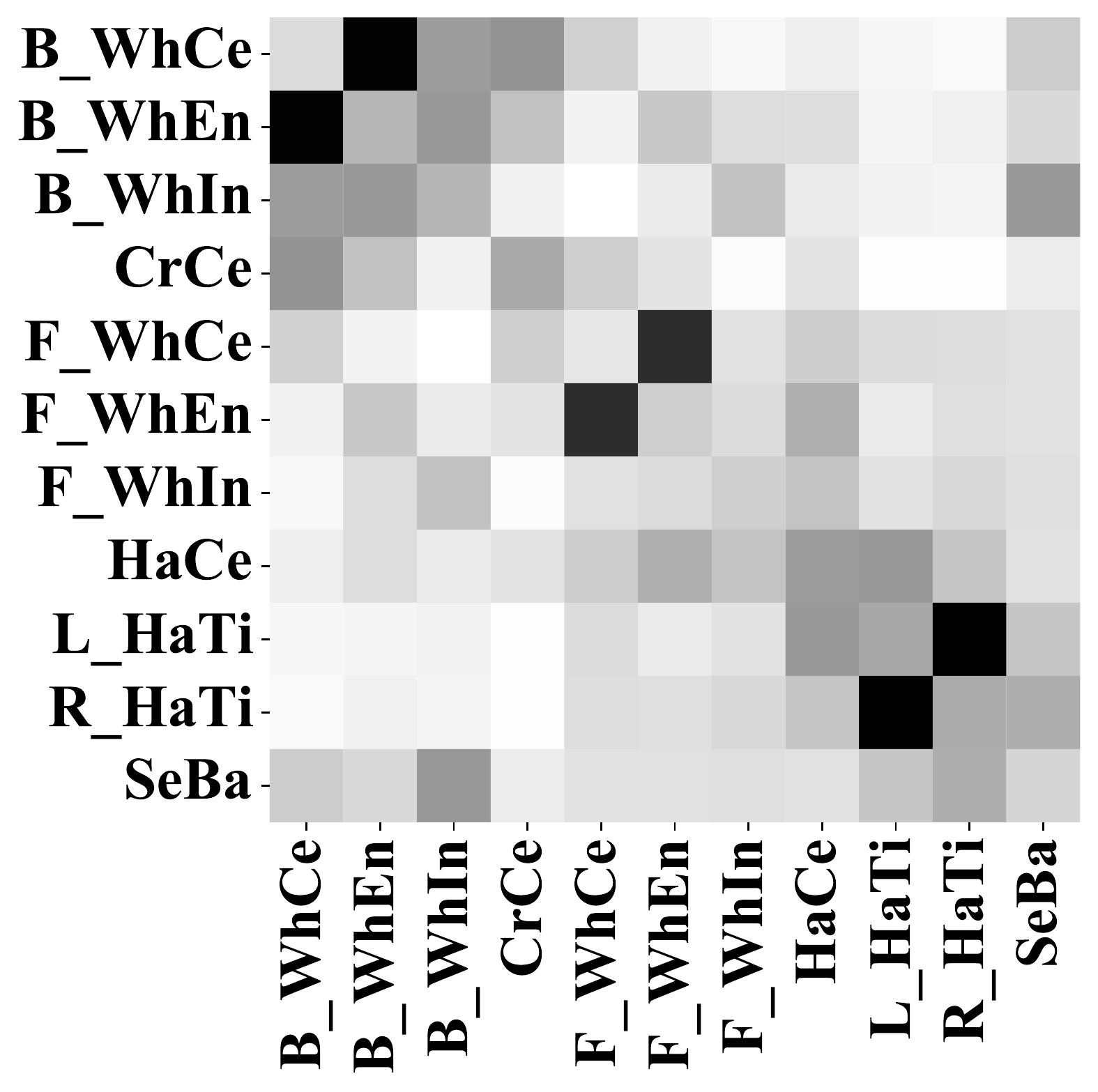}
}
\quad
\subfigure[boat]{
\includegraphics[width=4cm]{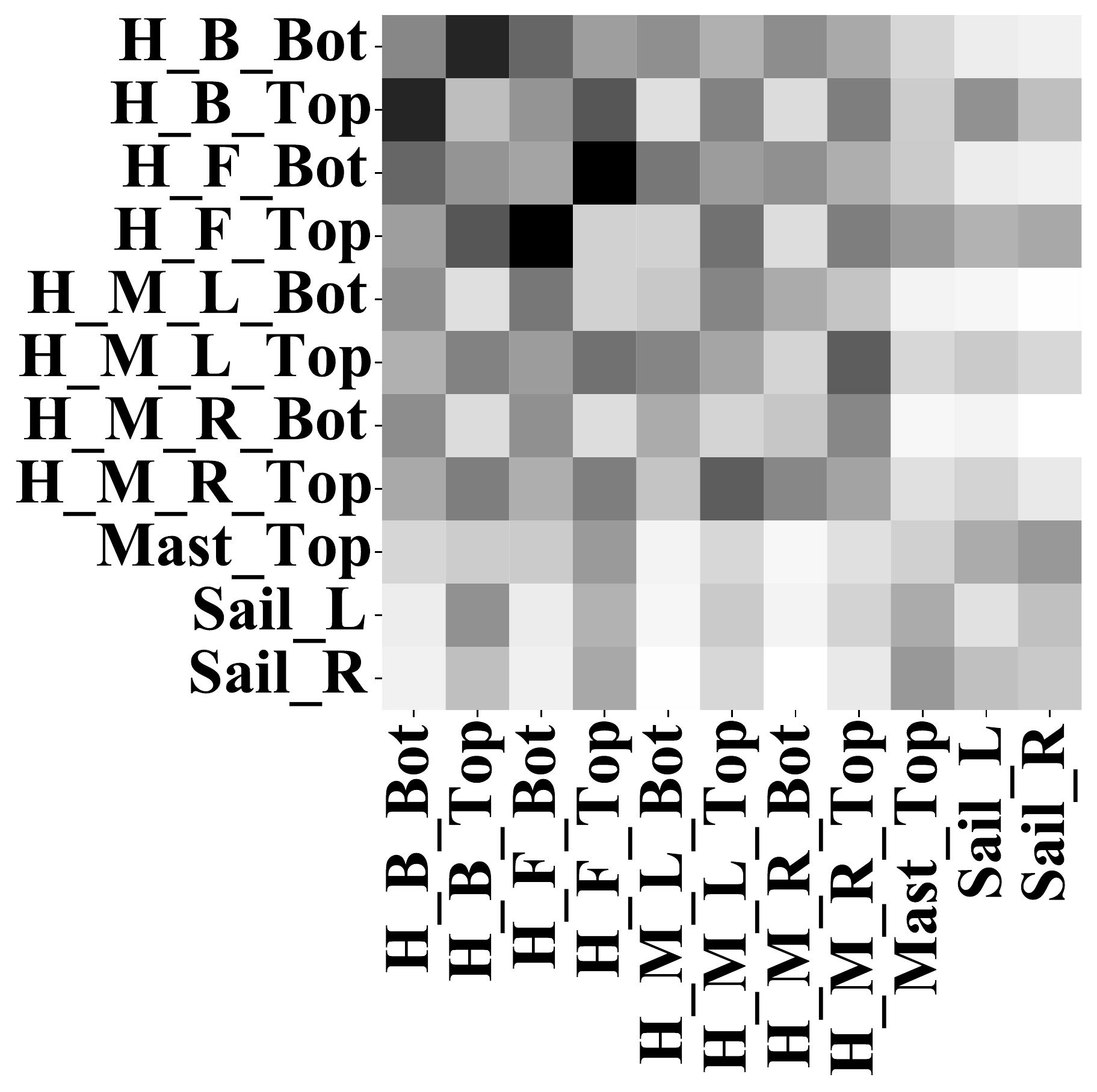}
}
\quad
\subfigure[bottle]{
\includegraphics[width=4cm]{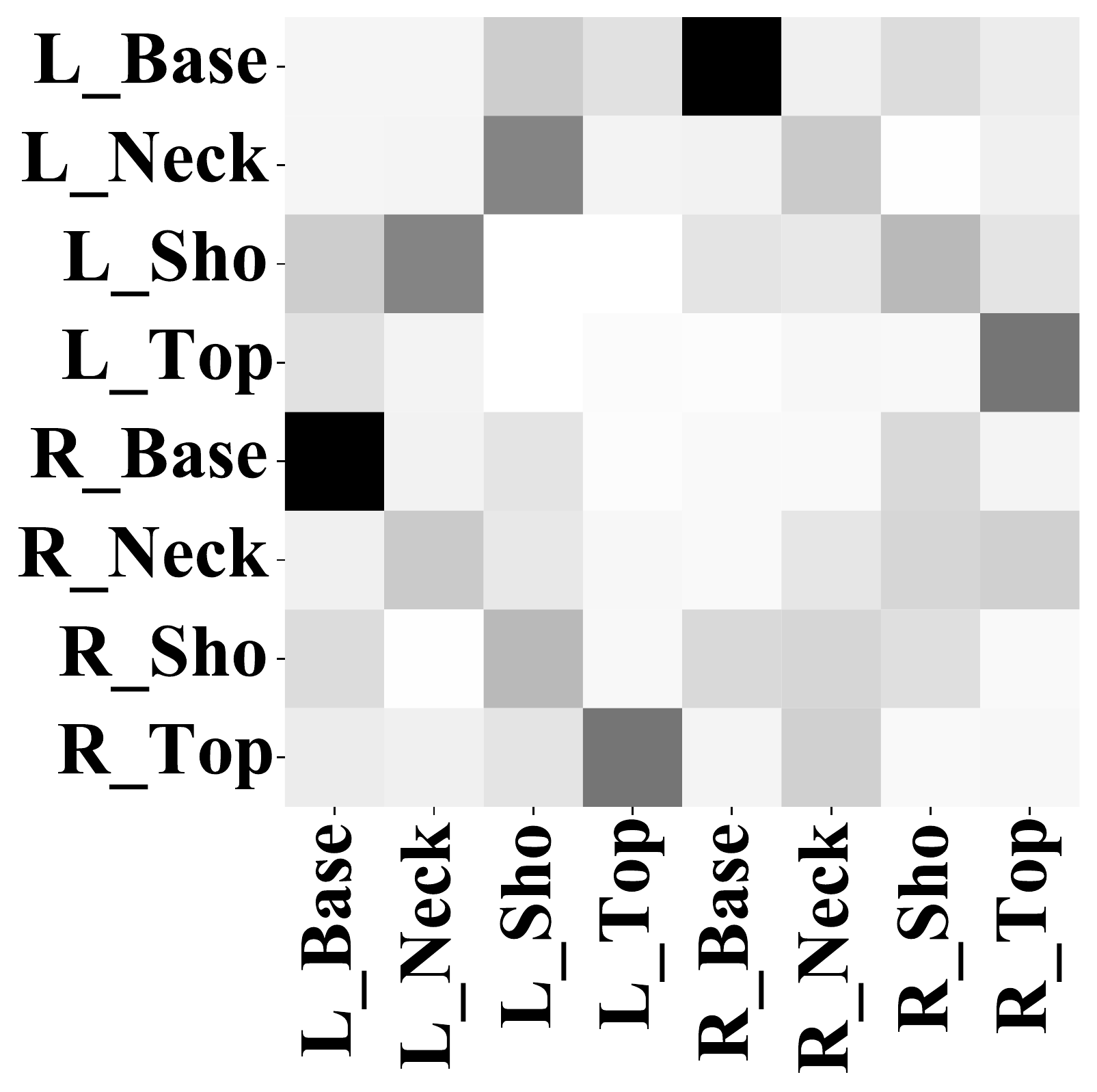}
}
\quad
\subfigure[bus]{
\includegraphics[width=4cm]{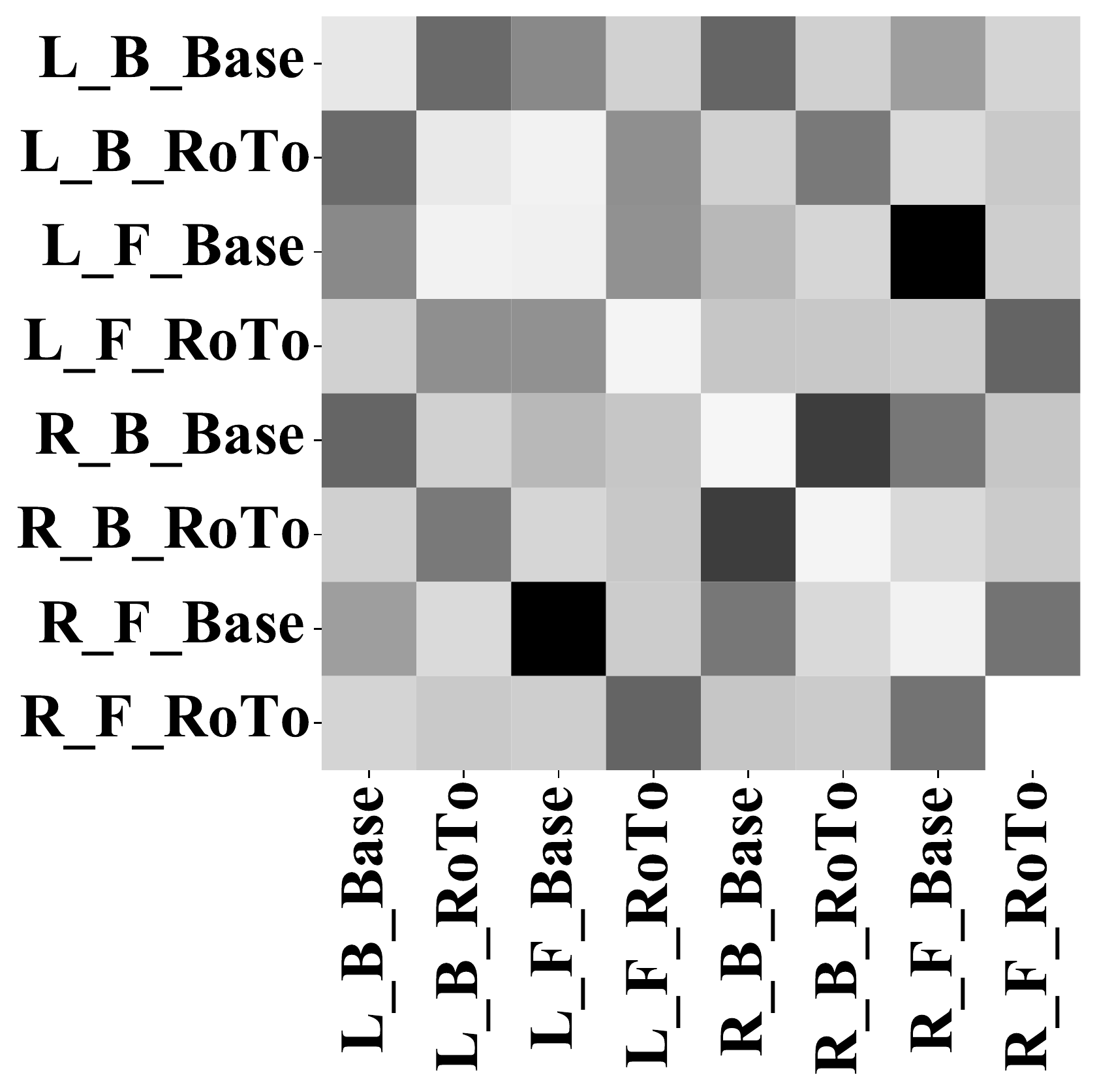}
}
\quad
\subfigure[chair]{
\includegraphics[width=4cm]{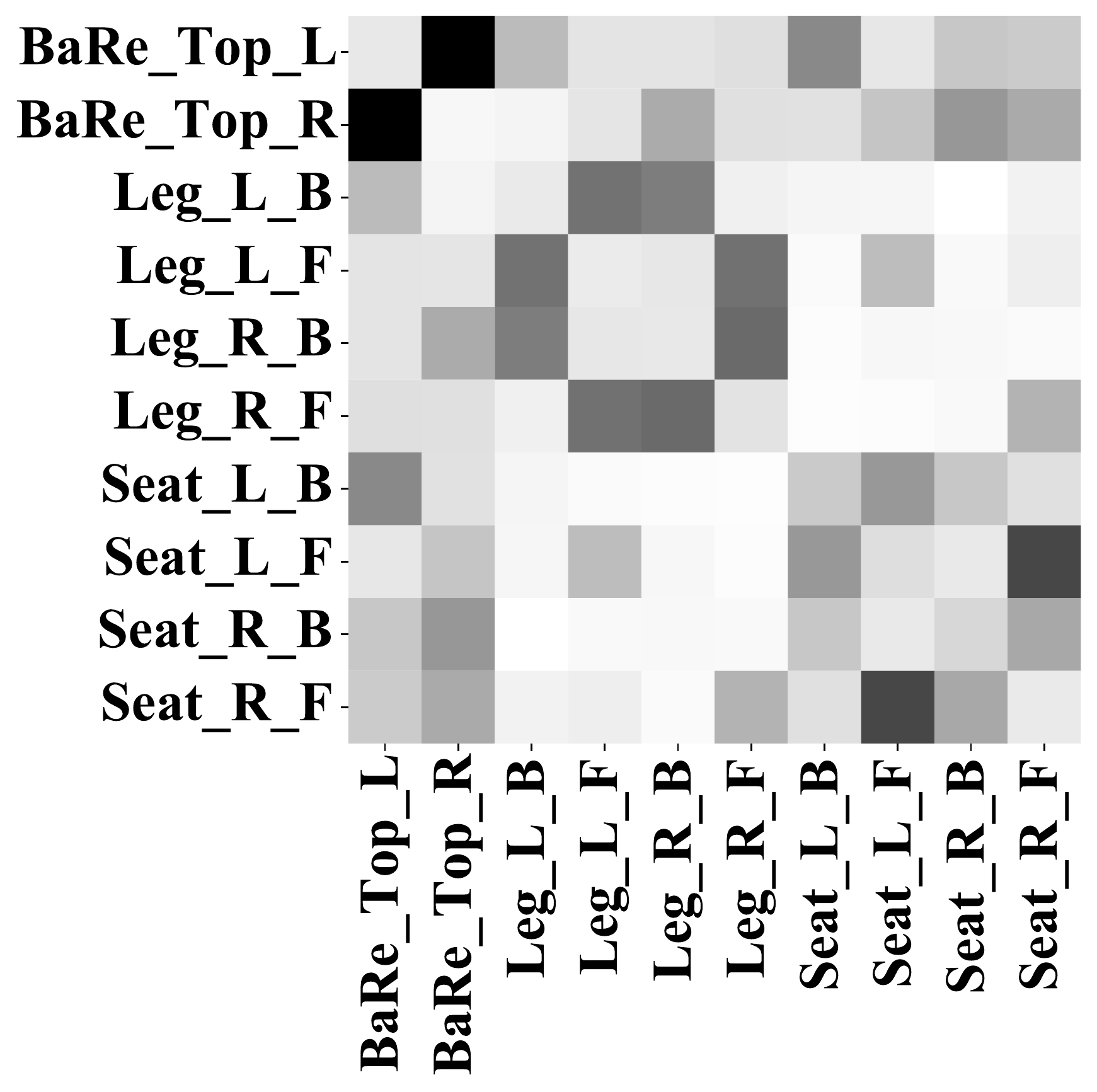}
}
\quad
\subfigure[diningtable]{
\includegraphics[width=4cm]{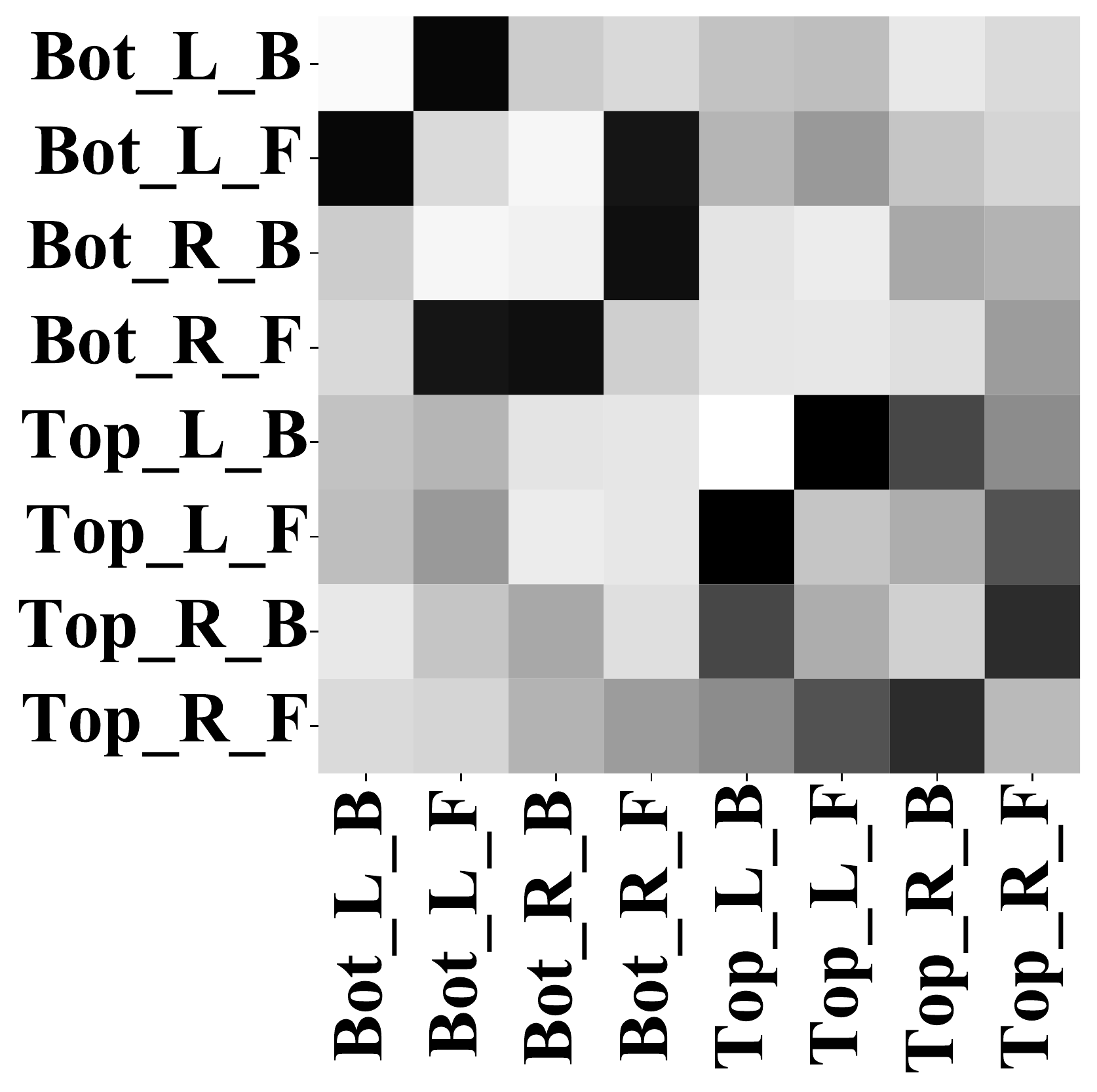}
}
\quad
\subfigure[motorbike]{
\includegraphics[width=4cm]{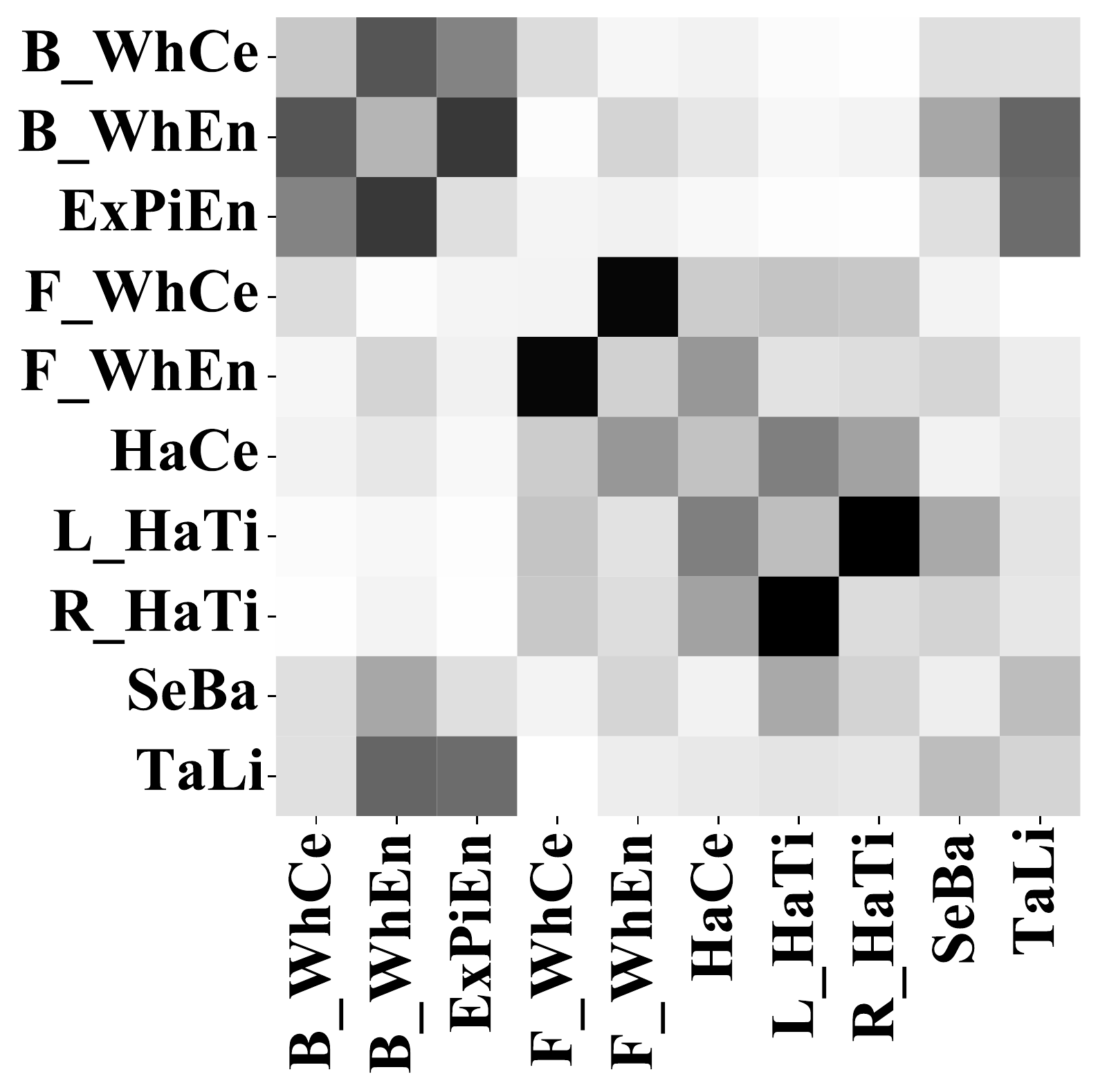}
}
\quad
\subfigure[pottedplant]{
\includegraphics[width=4cm]{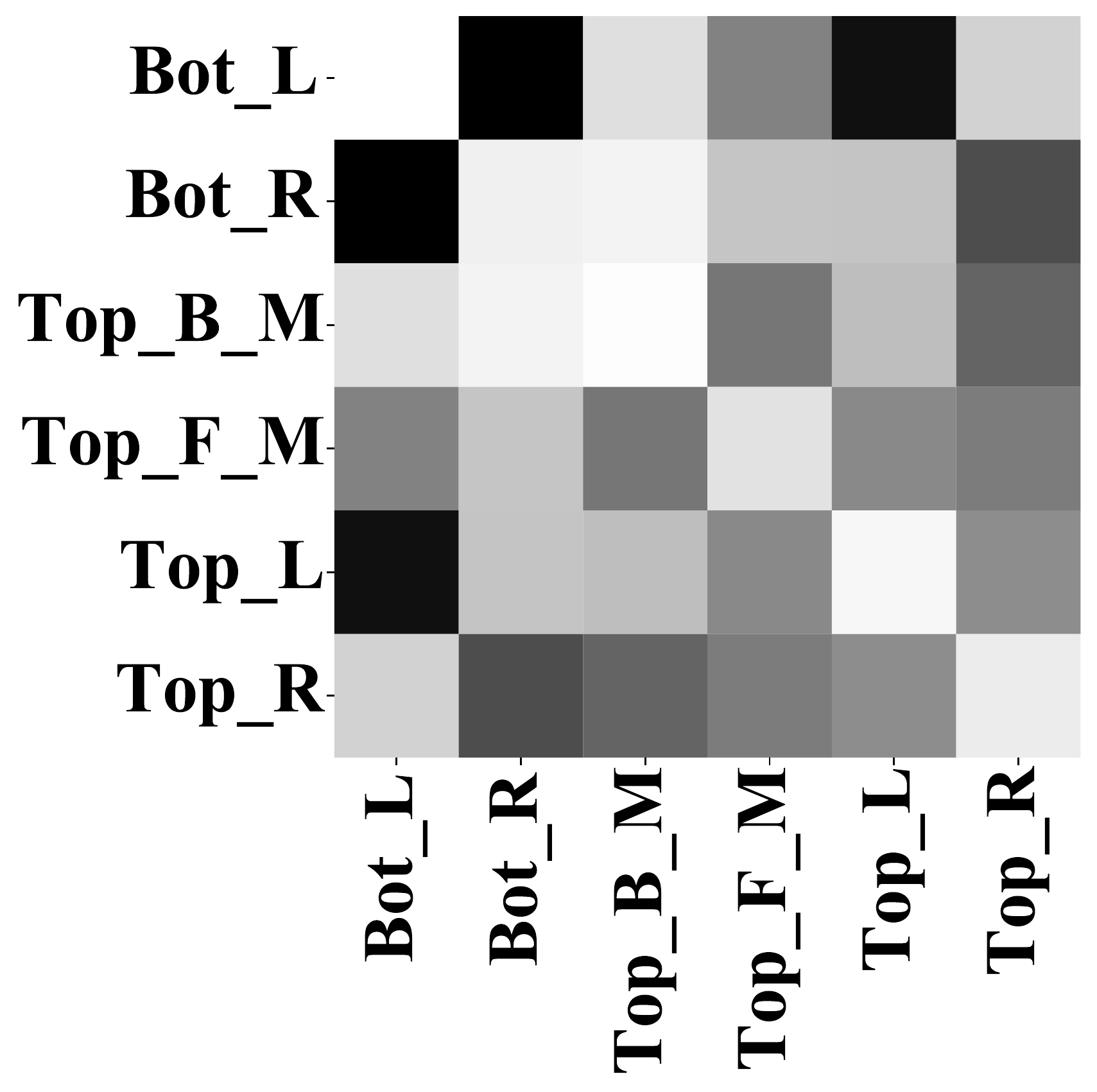}
}
\quad
\subfigure[train]{
\includegraphics[width=4cm]{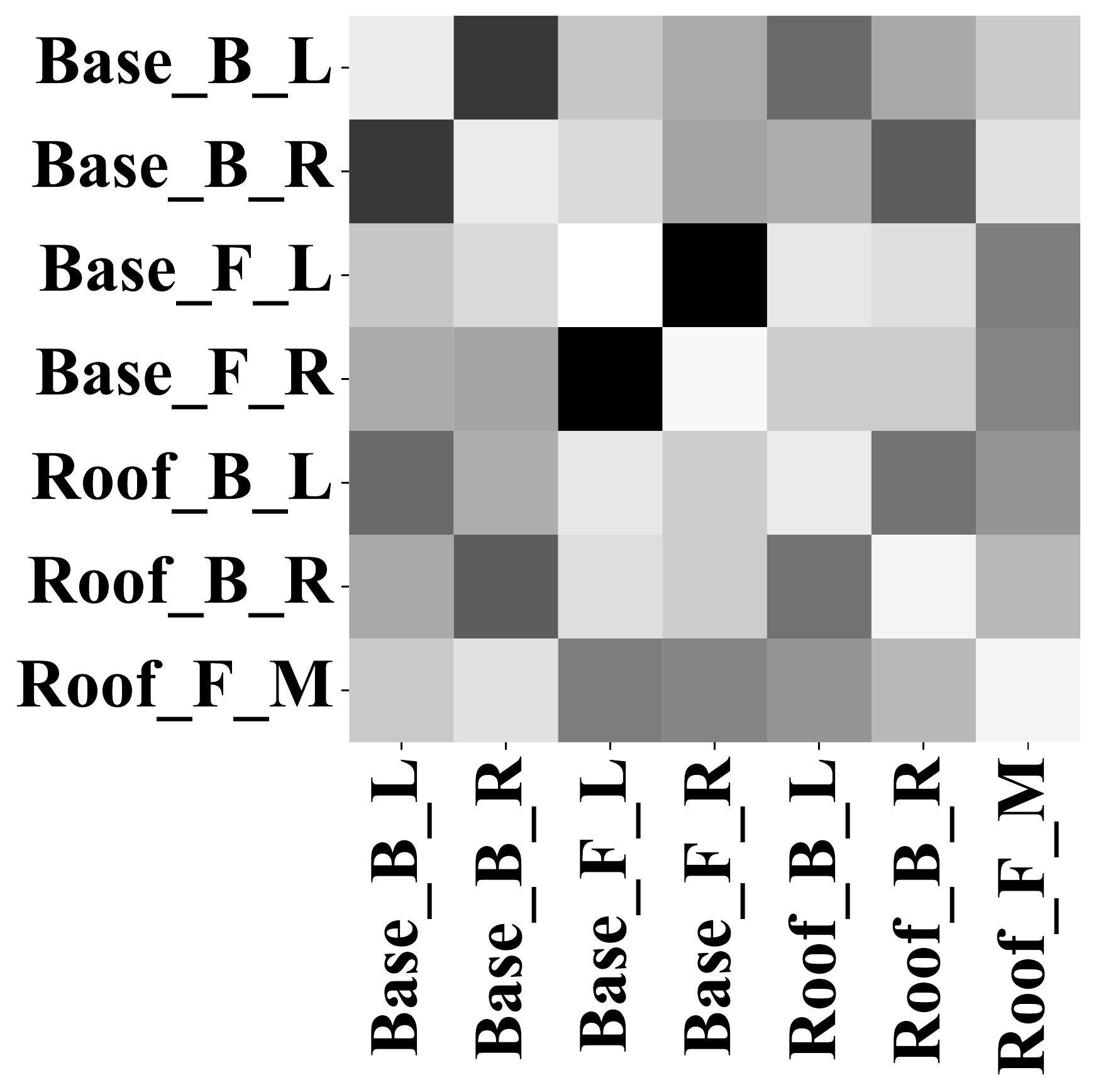}
}
\quad
\subfigure[tvmonitor]{
\includegraphics[width=4cm]{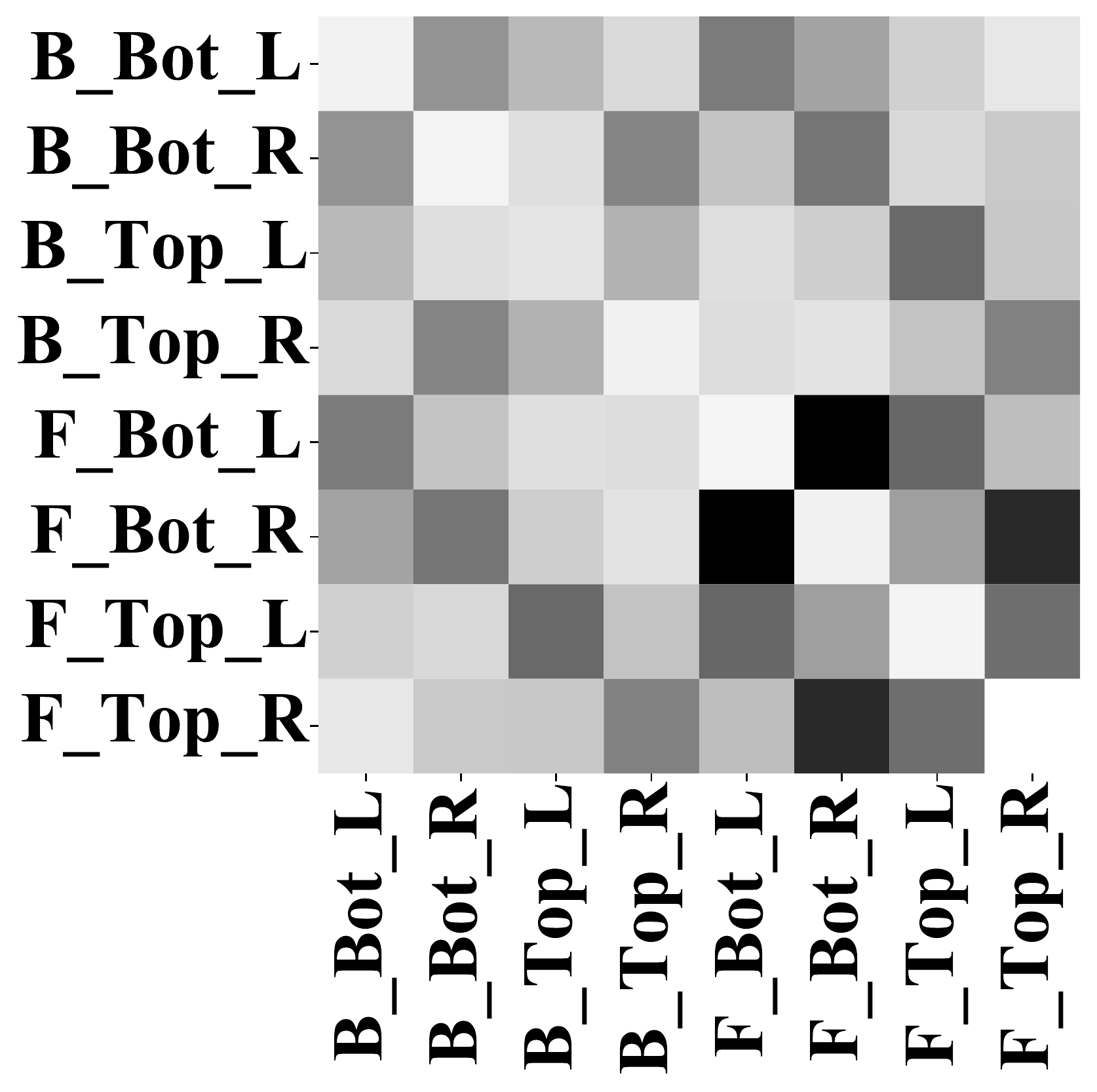}
}
\caption{The learnt adjacency matrix of the categories with less annotations. The darker the element, the larger the value, the lighter the element, the smaller the value.}
\label{fig:hmSimple}
\end{figure}

We split the categories in Pascal VOC into a difficult set and a simple one according to the number of semantic keypoints. For illustration, we visualize the heat maps of learnt weighted adjacency for them in Fig.~\ref{fig:hmDiff} and Fig.~\ref{fig:hmSimple} respectively, where darker elements indicate stronger relationships between keypoints.

It is obviously that the learnt graph patterns by our framework contains rich and discriminative semantic relationships between keypoints in each category. For example in the~\emph{Person} category in Fig.~\ref{fig:hmDiff}, \emph{L\_Eye} has much stronger connections to keypoints~\emph{L\_Ear}, \emph{Nose} and~\emph{R\_Eye} than other keypoints, which is consistence with our semantic adjacency rule. There are various aircraft type in~\emph{aeroplane} category that show great differences in structure and appearance, and its heat map is thus more vague than those of other categories. However, the keypoints in local area still have stronger relationships, for instance, \emph{Nose\_Bottom} has higher linking weights with~\emph{NoseTip} and~\emph{Nose\_Top} than other keypoints. 
Furthermore, it is interesting that all categories about animals except~\emph{bird} have vary similar distributions in heat maps. Specifically, for categories~\emph{cow}, \emph{dog}, \emph{cat}, \emph{horse} and~\emph{sheep}, the~\emph{L\_Eye} consistently has stronger relationships with~\emph{L\_EarBase}, \emph{Nose} and \emph{R\_Eye} than other keypoints.
Besides the categories mentioned above, the heat maps of the left categories including~\emph{bird}, \emph{car} and~\emph{sofa} also shows the clear semantic relationships between pairs of keypoints.
In Fig.~\ref{fig:hmSimple} that illustrates the distributions of learnt patterns of simple categories, the disparity between stronger relationships and weak relationships are more obvious for most heat maps. Especially, for categories~\emph{bottle}, \emph{diningtable} and~\emph{pottedplant} that have regular shapes in real world, stronger connections are more likely to exist between keypoints that have left-to-right or up-and-down relationships.

\subsection{Manually labeled graph patterns}
Benefiting from the fact that keypoints in Pascal VOC~\cite{VOC} have been annotated with semantic names, we are able to manually label a semantic graph pattern for each category, and compare them with the graph patterns learnt by our method for validation.

Since different instances in the same category vary greatly in scale, pose and viewpoint, we label the graph pattern of each category mainly according to the semantic relationships between the keypoints, instead of their spatial locations.
For example, among the keypoints in~\emph{person}, the~\emph{Nose} is more related to the organs on head such as~\emph{Eyes} and~\emph{Ears}, and less related to other organs on body including~\emph{Hands} and~\emph{Elbows}. Thus, for all instance of the \emph{person} category, keypoint~\emph{Nose} is connected to~\emph{Eyes} and~\emph{Ears}, but not to other keypoints no matter how close they are spatially.

\begin{figure}[t]
\begin{center}
\includegraphics[width=8.8cm,height=9cm, trim=0 0 0 0,clip]{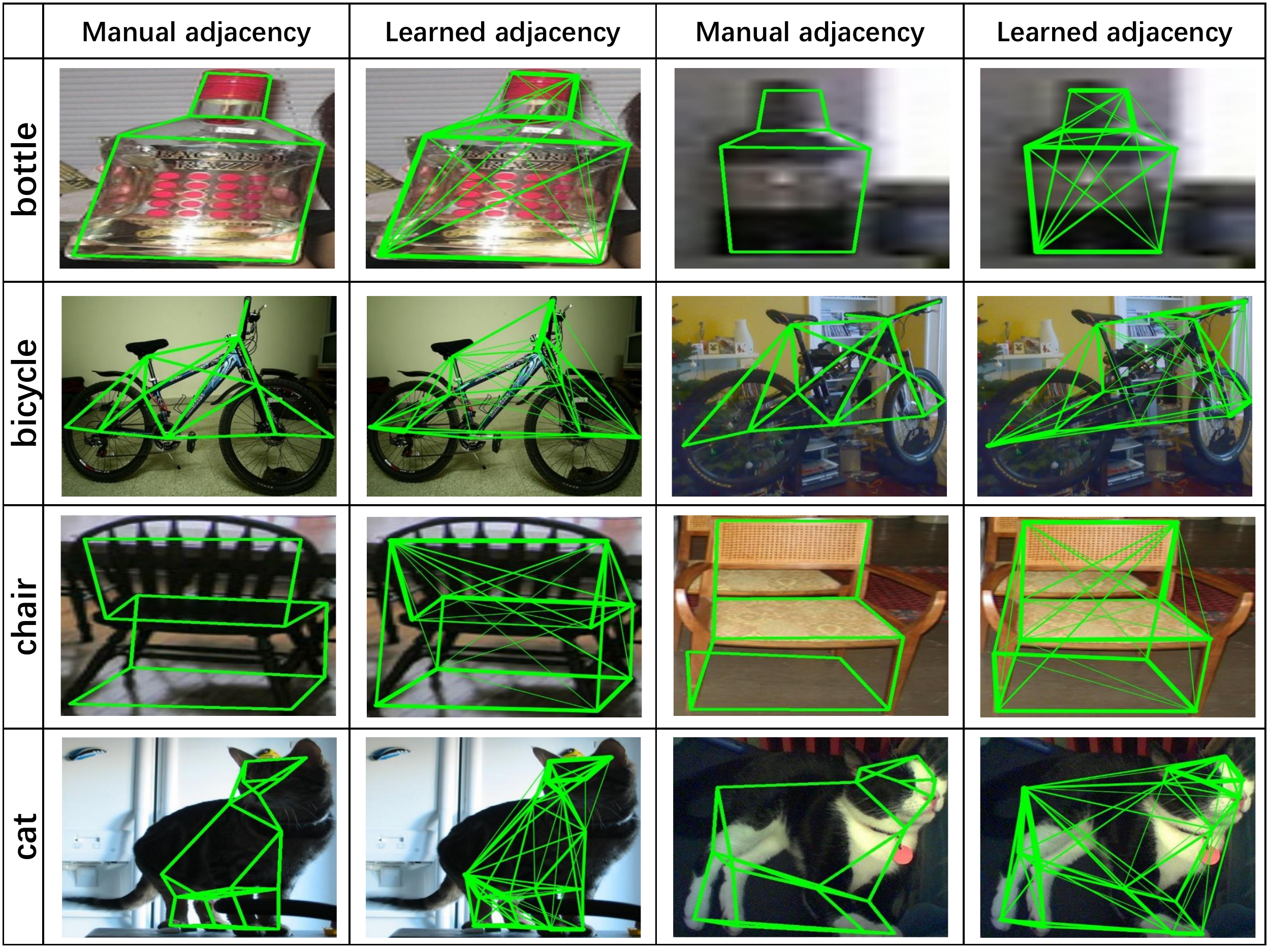}
\end{center}
\caption{Some examples of the manual semantic adjacency and the learnt weight adjacency. For the learnt adjacency, the thicker the line, the greater the weight denoting the linking between two keypoints.}
\label{fig:adj}
\end{figure}

Fig.~\ref{fig:adj} illustrates several representative examples of the manual semantic adjacency and the learnt weighted adjacency, in which the thickness of the line represents the strength of the relationship, and too weak links are filtered out for better illustration. It is observed that the learnt patterns are very similar to the manually labeled ones where the link weights between pairs of keypoints with close semantic relationship are prone to be greater, and the ones between weakly related keypoint pairs in semantic labels is relative small. For example, in the last row expressing the adjacencies of \emph{cat}, the \emph{left ear} has relatively strong relationships with \emph{right ear} and \emph{left eye}, while it has weak relationships with any of the \emph{legs}, which is consistent with the semantic adjacency annotated manually.

\subsection{Comparison with handcrafted graph structures}

For further validation of the effectiveness of the learnt graph patterns, we compare them, as well as manually labeled patterns, with the handcrafted graph patterns in several state-of-the-art graph matching algorithms, including GMN~\cite{DGM}, PCA~\cite{PCA}, IPCA~\cite{PCA-PAMI}, qc-DGM~\cite{qcDGM}, CIE~\cite{CIE}, BBGM~\cite{BBGM}, NGM~\cite{NGM} and NGMv2~\cite{NGM}.
We replace their graph construction strategy with the learnt graph patterns of our model (respectively the manually labeled patterns), and then use the pretrained parameters\footnote{The parameters and results of the pretrained models in Table~\ref{table:MA_LA} are provided at: https://github.com/Thinklab-SJTU/ThinkMatch.} of these models to perform the graph matching task.
In practice, in order to filter possible noises in the learnt adjacent matrix, we keep the 70\% adjacent edges with the highest weights and remove the rest edges.

Table~\ref{table:MA_LA} reports the comparison of matching performance between different types of graph patterns, where MODEL+MA and MODEL+LA denote using the manual semantic adjacency and learnt adjacency for graph construction, respectively.
Obviously, even using the network parameters that are pretrained under the original graph construction strategies without refining, imposing either
the manual semantic patterns or the learnt graph patterns is able to significantly improve the matching accuracy of most baseline algorithm. It indicates that building high-quality of relational biases is crucial for the graph matching framework. Furthermore, using learnt graph patterns achieves comparable performance to that using the manual semantic patterns, which also illustrates the effectiveness and representativeness of the learnt graph patterns.


\section{Conclusion}
In this paper, we propose a joint graph learning and matching network for the feature correspondence problem, by integrating a graph construction procedure into an end-to-end learnable framework. Specifically, we design two types of attention layers, self-attention layer and cross-attention layer, with the multi-head attention strategy. The self-attention layer is devoted to updating the features using learnt graph structures. By contrast, the cross-attention layer updates one feature set according to the information from the other, and it finally outputs a soft matching solution for graph matching.  Extensive experimental results on three public visual matching benchmarks have demonstrated the superiority of the proposed model. In addition, we show that the learnt structures are consistent with manually determined semantic patterns in general, and demonstrate the effectiveness of the learnt graph patterns in improving previous graph matching methods by replacing their graph construction strategies.



\ifCLASSOPTIONcompsoc
  \section*{Acknowledgments}
\else
  \section*{Acknowledgment}
\fi

This work is supported by the National Nature Science Foundation of China (Nos. 62076021 and 61872032) and the Beijing Municipal Natural Science Foundation (Nos.4202060 and 4212041).

\ifCLASSOPTIONcaptionsoff
  \newpage
\fi



%
\bibliographystyle{IEEEtran}
\bibliography{IEEEabrv,reference}

\begin{thebibliography}{10}
\providecommand{\url}[1]{#1}
\csname url@samestyle\endcsname
\providecommand{\newblock}{\relax}
\providecommand{\bibinfo}[2]{#2}
\providecommand{\BIBentrySTDinterwordspacing}{\spaceskip=0pt\relax}
\providecommand{\BIBentryALTinterwordstretchfactor}{4}
\providecommand{\BIBentryALTinterwordspacing}{\spaceskip=\fontdimen2\font plus
\BIBentryALTinterwordstretchfactor\fontdimen3\font minus
  \fontdimen4\font\relax}
\providecommand{\BIBforeignlanguage}[2]{{%
\expandafter\ifx\csname l@#1\endcsname\relax
\typeout{** WARNING: IEEEtran.bst: No hyphenation pattern has been}%
\typeout{** loaded for the language `#1'. Using the pattern for}%
\typeout{** the default language instead.}%
\else
\language=\csname l@#1\endcsname
\fi
#2}}
\providecommand{\BIBdecl}{\relax}
\BIBdecl

\bibitem{BrownL07}
M.~Brown and D.~G. Lowe, ``Automatic panoramic image stitching using invariant
  features,'' \emph{Int. J. Comput. Vis.}, vol.~74, no.~1, pp. 59--73, 2007.

\bibitem{gracker}
T.~Wang and H.~Ling, ``Gracker: {A} graph-based planar object tracker,''
  \emph{{IEEE} Trans. Pattern Anal. Mach. Intell.}, vol.~40, no.~6, pp.
  1494--1501, 2018.

\bibitem{Conte2004}
D.~Conte, P.~Foggia, C.~Sansone, and M.~Vento, ``Thirty years of graph matching
  in pattern recognition,'' \emph{Int. J. Pattern Recognit. Artif. Intell},
  vol.~18, no.~3, pp. 265--298, 2004.

\bibitem{Foggia2014}
P.~Foggia, G.~Percannella, and M.~Vento, ``Graph matching and learning in
  pattern recognition in the last 10 years,'' \emph{Int. J. Pattern Recognit.
  Artif. Intell}, vol.~28, no.~1, pp. 1--40, 2014.

\bibitem{survey_yan}
J.~Yan, X.~Yin, W.~Lin, C.~Deng, H.~Zha, and X.~Yang, ``A short survey of
  recent advances in graph matching,'' in \emph{Int. Conf. Multimed. Retr.},
  2016, pp. 167--174.

\bibitem{NowakVBB18}
A.~Nowak, S.~Villar, A.~S. Bandeira, and J.~Bruna, ``Revised note on learning
  quadratic assignment with graph neural networks,'' in \emph{{IEEE} Data Sci.
  Workshop}, 2018, pp. 229--233.

\bibitem{PCA}
R.~Wang, J.~Yan, and X.~Yang, ``Learning combinatorial embedding networks for
  deep graph matching,'' in \emph{{IEEE} Int. Conf. Comput. Vis.}, 2019, pp.
  3056--3065.

\bibitem{DGM_consensus}
M.~Fey, J.~E. Lenssen, C.~Morris, J.~Masci, and N.~M. Kriege, ``Deep graph
  matching consensus,'' in \emph{Int. Conf. Learn. Represent.}, 2020.

\bibitem{CIE}
T.~Yu, R.~Wang, J.~Yan, and B.~Li, ``Learning deep graph matching with
  channel-independent embedding and hungarian attention,'' in \emph{Int. Conf.
  Learn. Represent.}, 2020.

\bibitem{SuperGlue}
P.~Sarlin, D.~DeTone, T.~Malisiewicz, and A.~Rabinovich, ``Superglue: Learning
  feature matching with graph neural networks,'' in \emph{{IEEE} Conf. Comput.
  Vis. Pattern Recog.}, 2020, pp. 4937--4946.

\bibitem{VOC}
M.~Everingham, L.~V. Gool, C.~K.~I. Williams, J.~M. Winn, and A.~Zisserman,
  ``The pascal visual object classes {(VOC)} challenge,'' \emph{Int. J. Comput.
  Vis.}, vol.~88, no.~2, pp. 303--338, 2010.

\bibitem{transformer}
A.~Vaswani, N.~Shazeer, N.~Parmar, J.~Uszkoreit, L.~Jones, A.~N. Gomez,
  L.~Kaiser, and I.~Polosukhin, ``Attention is all you need,'' in \emph{Conf.
  Neural Inform. Process. Syst.}, 2017, pp. 5998--6008.

\bibitem{LG2M}
M.~Cho, K.~Alahari, and J.~Ponce, ``Learning graphs to match,'' in \emph{{IEEE}
  Int. Conf. Comput. Vis.}, 2013, pp. 25--32.

\bibitem{SPair}
\BIBentryALTinterwordspacing
J.~Min, J.~Lee, J.~Ponce, and M.~Cho, ``Spair-71k: {A} large-scale benchmark
  for semantic correspondence,'' \emph{CoRR}, vol. abs/1908.10543, 2019.
  [Online]. Available: \url{http://arxiv.org/abs/1908.10543}
\BIBentrySTDinterwordspacing

\bibitem{LapMatching}
W.~Lian and L.~Zhang, ``Robust point matching revisited: {A} concave
  optimization approach,'' in \emph{Eur. Conf. Comput. Vis.}, vol. 7573, 2012,
  pp. 259--272.

\bibitem{LapMatching2}
{Wei Lian and Lei Zhang}, ``Point matching in the presence of outliers in both
  point sets: {A} concave optimization approach,'' in \emph{{IEEE} Conf.
  Comput. Vis. Pattern Recog.}, 2014, pp. 352--359.

\bibitem{code}
W.~Lin, F.~Wang, M.~Cheng, S.~Yeung, P.~H.~S. Torr, M.~N. Do, and J.~Lu,
  ``{CODE:} coherence based decision boundaries for feature correspondence,''
  \emph{{IEEE} Trans. Pattern Anal. Mach. Intell.}, vol.~40, no.~1, pp. 34--47,
  2018.

\bibitem{Co-Seg}
H.~Chen, Y.~Lin, and B.~Chen, ``Co-segmentation guided hough transform for
  robust feature matching,'' \emph{{IEEE} Trans. Pattern Anal. Mach. Intell.},
  vol.~37, no.~12, pp. 2388--2401, 2015.

\bibitem{fast-hough}
G.~Tolias and Y.~Avrithis, ``Speeded-up, relaxed spatial matching,'' in
  \emph{{IEEE} Int. Conf. Comput. Vis.}, 2011, pp. 1653--1660.

\bibitem{HVIV}
H.~Chen, Y.~Lin, and B.~Chen, ``Robust feature matching with alternate hough
  and inverted hough transforms,'' in \emph{{IEEE} Conf. Comput. Vis. Pattern
  Recog.}, 2013, pp. 2762--2769.

\bibitem{shapematch}
S.~J. Belongie, J.~Malik, and J.~Puzicha, ``Shape matching and object
  recognition using shape contexts,'' \emph{{IEEE} Trans. Pattern Anal. Mach.
  Intell.}, vol.~24, no.~4, pp. 509--522, 2002.

\bibitem{NM-Net}
C.~Zhao, Z.~Cao, C.~Li, X.~Li, and J.~Yang, ``Nm-net: Mining reliable neighbors
  for robust feature correspondences,'' in \emph{{IEEE} Conf. Comput. Vis.
  Pattern Recog.}, 2019, pp. 215--224.

\bibitem{ZhouT13}
F.~Zhou and F.~D. la~Torre, ``Deformable graph matching,'' in \emph{{IEEE}
  Conf. Comput. Vis. Pattern Recog.}, 2013, pp. 2922--2929.

\bibitem{Cho0DP14}
M.~Cho, J.~Sun, O.~Duchenne, and J.~Ponce, ``Finding matches in a haystack: {A}
  max-pooling strategy for graph matching in the presence of outliers,'' in
  \emph{{IEEE} Conf. Comput. Vis. Pattern Recog.}, 2014, pp. 2091--2098.

\bibitem{WangLLFH19}
T.~Wang, H.~Ling, C.~Lang, S.~Feng, and X.~Hou, ``Deformable surface tracking
  by graph matching,'' in \emph{{IEEE} Int. Conf. Comput. Vis.}, 2019, pp.
  901--910.

\bibitem{LeeLS20}
S.~Lee, J.~Lim, and I.~H. Suh, ``Progressive feature matching: Incremental
  graph construction and optimization,'' \emph{{IEEE} Trans. Image Process.},
  vol.~29, pp. 6992--7005, 2020.

\bibitem{ABPF}
T.~Wang, H.~Ling, C.~Lang, and S.~Feng, ``Graph matching with adaptive and
  branching path following,'' \emph{{IEEE} Trans. Pattern Anal. Mach. Intell.},
  vol.~40, no.~12, pp. 2853--2867, 2018.

\bibitem{tr_2005ICCV_1}
M.~Leordeanu and M.~Hebert, ``A spectral technique for correspondence problems
  using pairwise constraints,'' in \emph{{IEEE} Int. Conf. Comput. Vis.}, 2005,
  pp. 1482--1489.

\bibitem{Fact_GM}
F.~Zhou and F.~D. la~Torre, ``Factorized graph matching,'' \emph{{IEEE} Trans.
  Pattern Anal. Mach. Intell.}, vol.~38, no.~9, pp. 1774--1789, 2016.

\bibitem{PFGM}
M.~Zaslavskiy, F.~R. Bach, and J.~Vert, ``A path following algorithm for the
  graph matching problem,'' \emph{{IEEE} Trans. Pattern Anal. Mach. Intell.},
  vol.~31, no.~12, pp. 2227--2242, 2009.

\bibitem{GNCCP}
Z.~Liu and H.~Qiao, ``{GNCCP} - graduated nonconvexityand concavity
  procedure,'' \emph{{IEEE} Trans. Pattern Anal. Mach. Intell.}, vol.~36,
  no.~6, pp. 1258--1267, 2014.

\bibitem{IPFP}
M.~Leordeanu, M.~Hebert, and R.~Sukthankar, ``An integer projected fixed point
  method for graph matching and {MAP} inference,'' in \emph{Conf. Neural
  Inform. Process. Syst.}, 2009, pp. 1114--1122.

\bibitem{Spec_Tech}
M.~Leordeanu and M.~Hebert, ``A spectral technique for correspondence problems
  using pairwise constraints,'' in \emph{{IEEE} Int. Conf. Comput. Vis.}, 2005,
  pp. 1482--1489.

\bibitem{Graduated_Assignment}
S.~Gold and A.~Rangarajan, ``A graduated assignment algorithm for graph
  matching,'' \emph{{IEEE} Trans. Pattern Anal. Mach. Intell.}, vol.~18, no.~4,
  pp. 377--388, 1996.

\bibitem{BGM}
T.~Cour, P.~Srinivasan, and J.~Shi, ``Balanced graph matching,'' in \emph{Conf.
  Neural Inform. Process. Syst.}, 2006, pp. 313--320.

\bibitem{RWGM}
M.~Cho, J.~Lee, and K.~M. Lee, ``Reweighted random walks for graph matching,''
  in \emph{Eur. Conf. Comput. Vis.}, vol. 6315, 2010, pp. 492--505.

\bibitem{PBGM}
A.~Egozi, Y.~Keller, and H.~Guterman, ``A probabilistic approach to spectral
  graph matching,'' \emph{{IEEE} Trans. Pattern Anal. Mach. Intell.}, vol.~35,
  no.~1, pp. 18--27, 2013.

\bibitem{PCA-PAMI}
R.~Wang, J.~Yan, and X.~Yang, ``Combinatorial learning of robust deep graph
  matching: an embedding based approach,'' \emph{{IEEE} Trans. Pattern Anal.
  Mach. Intell.}, 2020.

\bibitem{KipfW17}
T.~N. Kipf and M.~Welling, ``Semi-supervised classification with graph
  convolutional networks,'' in \emph{Int. Conf. Learn. Represent.}, 2017.

\bibitem{Sinkhorn_network}
G.~Mena, D.~Belanger, G.~Munoz, and J.~Snoek, ``Sinkhorn networks: Using
  optimal transport techniques to learn permutations,'' in \emph{NIPS Workshop
  Optim. Transp. Mach. Learn.}, 2017, pp. 1--10.

\bibitem{DGM}
A.~Zanfir and C.~Sminchisescu, ``Deep learning of graph matching,'' in
  \emph{{IEEE} Conf. Comput. Vis. Pattern Recog.}, 2018, pp. 2684--2693.

\bibitem{BBGM}
M.~Rol{\'{\i}}nek, P.~Swoboda, D.~Zietlow, A.~Paulus, V.~Musil, and G.~Martius,
  ``Deep graph matching via blackbox differentiation of combinatorial
  solvers,'' in \emph{Eur. Conf. Comput. Vis.}, vol. 12373, 2020, pp. 407--424.

\bibitem{SwobodaRAKS17}
P.~Swoboda, C.~Rother, H.~A. Alhaija, D.~Kainm{\"{u}}ller, and B.~Savchynskyy,
  ``A study of lagrangean decompositions and dual ascent solvers for graph
  matching,'' in \emph{{IEEE} Conf. Comput. Vis. Pattern Recog.}, 2017, pp.
  7062--7071.

\bibitem{LoiolaANHQ07}
E.~M. Loiola, N.~M.~M. de~Abreu, P.~O.~B. Netto, P.~Hahn, and T.~M. Querido,
  ``A survey for the quadratic assignment problem,'' \emph{Eur. J. Oper. Res.},
  vol. 176, no.~2, pp. 657--690, 2007.

\bibitem{qcDGM}
Q.~Gao, F.~Wang, N.~Xue, J.-G. Yu, and G.-S. Xia, ``Deep graph matching under
  quadratic constraint,'' in \emph{{IEEE} Conf. Comput. Vis. Pattern Recog.},
  2021, pp. 5069--5078.

\bibitem{Frank&Wolfe56}
M.~Frank and P.~Wolfe, ``An algorithm for quadratic programming,'' \emph{Nav.
  Res. Logist. Q.}, vol.~3, pp. 95--100, 1956.

\bibitem{LGM}
T.~Wang, H.~Liu, Y.~Li, Y.~Jin, X.~Hou, and H.~Ling, ``Learning combinatorial
  solver for graph matching,'' in \emph{{IEEE} Conf. Comput. Vis. Pattern
  Recog.}, 2020, pp. 7565--7574.

\bibitem{NGM}
{Wang, Runzhong and Yan, Junchi and Yang, Xiaokang}, ``Neural graph matching
  network: Learning lawler' s quadratic assignment problem with extension to
  hypergraph and multiple-graph matching,'' \emph{{IEEE} Trans. Pattern Anal.
  Mach. Intell.}, 2021.

\bibitem{DosovitskiyB0WZ21}
A.~Dosovitskiy, L.~Beyer, A.~Kolesnikov, D.~Weissenborn, X.~Zhai,
  T.~Unterthiner, M.~Dehghani, M.~Minderer, G.~Heigold, S.~Gelly, J.~Uszkoreit,
  and N.~Houlsby, ``An image is worth 16x16 words: Transformers for image
  recognition at scale,'' in \emph{Int. Conf. Learn. Represent.}, 2021.

\bibitem{KhanNHZKS21}
\BIBentryALTinterwordspacing
S.~H. Khan, M.~Naseer, M.~Hayat, S.~W. Zamir, F.~S. Khan, and M.~Shah,
  ``Transformers in vision: {A} survey,'' \emph{CoRR}, vol. abs/2101.01169,
  2021. [Online]. Available: \url{https://arxiv.org/abs/2101.01169}
\BIBentrySTDinterwordspacing

\bibitem{tf_vision_2}
T.~Xu, P.~Zhang, Q.~Huang, H.~Zhang, Z.~Gan, X.~Huang, and X.~He, ``Attngan:
  Fine-grained text to image generation with attentional generative adversarial
  networks,'' in \emph{{IEEE} Conf. Comput. Vis. Pattern Recog.}, 2018, pp.
  1316--1324.

\bibitem{tf_vision_3}
L.~Ye, M.~Rochan, Z.~Liu, and Y.~Wang, ``Cross-modal self-attention network for
  referring image segmentation,'' in \emph{{IEEE} Conf. Comput. Vis. Pattern
  Recog.}, 2019, pp. 10\,502--10\,511.

\bibitem{tf_vision_1}
J.~Fu, J.~Liu, H.~Tian, Y.~Li, Y.~Bao, Z.~Fang, and H.~Lu, ``Dual attention
  network for scene segmentation,'' in \emph{{IEEE} Conf. Comput. Vis. Pattern
  Recog.}, 2019, pp. 3146--3154.

\bibitem{image_tfer}
A.~Dosovitskiy, L.~Beyer, A.~Kolesnikov, D.~Weissenborn, X.~Zhai,
  T.~Unterthiner, M.~Dehghani, M.~Minderer, G.~Heigold, S.~Gelly, J.~Uszkoreit,
  and N.~Houlsby, ``An image is worth 16x16 words: Transformers for image
  recognition at scale,'' in \emph{Int. Conf. Learn. Represent.}, 2021.

\bibitem{ChenPFL21iccv}
M.~Chen, H.~Peng, J.~Fu, and H.~Ling, ``{AutoFormer}: Searching transformers
  for visual recognition,'' in \emph{{IEEE} Int. Conf. Comput. Vis.}, 2021.

\bibitem{GAT}
P.~Velickovic, G.~Cucurull, A.~Casanova, A.~Romero, P.~Li{\`{o}}, and
  Y.~Bengio, ``Graph attention networks,'' in \emph{Int. Conf. Learn.
  Represent.}, 2018.

\bibitem{U2GNN}
\BIBentryALTinterwordspacing
D.~Q. Nguyen, T.~D. Nguyen, and D.~Phung, ``Universal graph transformer
  self-attention networks,'' \emph{CoRR}, vol. abs/1909.11855, 2019. [Online].
  Available: \url{https://arxiv.org/abs/1909.11855}
\BIBentrySTDinterwordspacing

\bibitem{hungarian}
H.~W. Kuhn, ``The hungarian method for the assignment problem,'' in \emph{50
  Years of Integer Programming 2010}, 2010, pp. 29--47.

\bibitem{munkres}
J.~Munkres, ``Algorithms for the assignment and transportation problems,''
  \emph{J. Soc. Industr. Appl. Math.}, vol.~5, no.~1, pp. 32--38, 1957.

\bibitem{VGG16}
K.~Simonyan and A.~Zisserman, ``Very deep convolutional networks for
  large-scale image recognition,'' in \emph{Int. Conf. Learn. Represent.},
  Y.~Bengio and Y.~LeCun, Eds., 2015.

\bibitem{resnet}
K.~He, X.~Zhang, S.~Ren, and J.~Sun, ``Deep residual learning for image
  recognition,'' in \emph{{IEEE} Conf. Comput. Vis. Pattern Recog.}, 2016, pp.
  770--778.

\bibitem{splineCNN}
M.~Fey, J.~E. Lenssen, F.~Weichert, and H.~M{\"{u}}ller, ``Splinecnn: Fast
  geometric deep learning with continuous b-spline kernels,'' in \emph{{IEEE}
  Conf. Comput. Vis. Pattern Recog.}, 2018, pp. 869--877.

\bibitem{berkeley}
L.~D. Bourdev and J.~Malik, ``Poselets: Body part detectors trained using 3d
  human pose annotations,'' in \emph{{IEEE} Int. Conf. Comput. Vis.}, 2009, pp.
  1365--1372.

\bibitem{3DPascal}
Y.~Xiang, R.~Mottaghi, and S.~Savarese, ``Beyond {PASCAL:} {A} benchmark for 3d
  object detection in the wild,'' in \emph{{IEEE} Winter Conf. Appl. Comput.
  Vis.}, 2014, pp. 75--82.

\bibitem{VOC2012}
M.~Everingham, S.~M.~A. Eslami, L.~V. Gool, C.~K.~I. Williams, J.~M. Winn, and
  A.~Zisserman, ``The pascal visual object classes challenge: {A}
  retrospective,'' \emph{Int. J. Comput. Vis.}, vol. 111, no.~1, pp. 98--136,
  2015.

\end{thebibliography}
%

\begin{IEEEbiography}[{\includegraphics[width=1in,height=1.25in,clip,keepaspectratio]{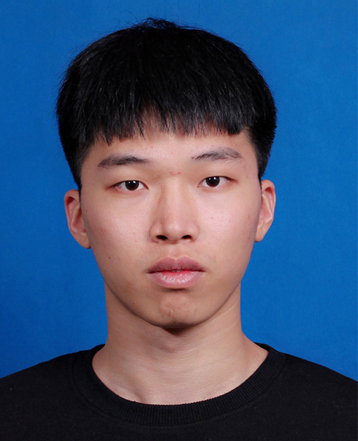}}]{He Liu}
received the B. S. degree in compute science from Hebei University of Technology, Tianjin, China. He is currently pursuing the Ph. D degree in the School of Computer and Information Technology, Beijing Jiaotong University, Beijing, China. His research concentrates on machine learning and graph matching.
\end{IEEEbiography}

\begin{IEEEbiography}[{\includegraphics[width=1in,height=1.25in,clip,keepaspectratio]{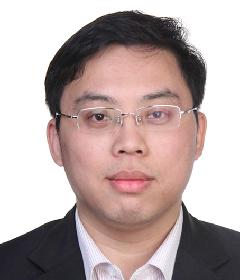}}]{Tao Wang}
received the PhD degree from the School of Computer and Information Technology, Beijing Jiaotong University, Beijing, China, in 2013. He was a visiting professor with the Department of Computer and Information Sciences, Temple University, from 2014 to 2015. He is currently an associate professor with the School of Computer and Information Technology, Beijing Jiaotong University. His research interests include graph matching theory with application to image analysis and retrieval.
\end{IEEEbiography}

\begin{IEEEbiography}[{\includegraphics[width=1in,height=1.25in,clip,keepaspectratio]{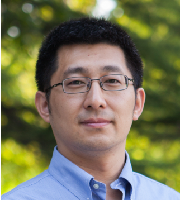}}]{Yidong Li}
was born in 1982, native to Shanxi, associate professor and Ph.D. supervisor. In 2003, he graduated from the Department of information and communication engineering of Beijing Jiaotong University. He received master and Ph. D. degree from the department of computer science of the University of Adelaide in Australia in 2006 and 2011 respectively. Dr. Yi is the vice dean of the school of computer and information technology, Beijing Jiaotong University, executive director of the SAP University Competence Center (China).
\end{IEEEbiography}

\begin{IEEEbiography}[{\includegraphics[width=1in,height=1.25in,clip,keepaspectratio]{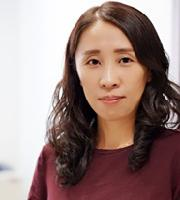}}]{Congyan Lang}
received her Ph.D. degree from the School of Computer and Information Technology, Beijing Jiaotong University, Beijing, China, in 2006. She was a Visiting Professor with the Department of Electrical and Computer Engineering, National
University of Singapore, Singapore, from 2010 to 2011. From 2014 to 2015, she visited the Department of Computer Science, University of Rochester, Rochester, NY, USA, as a Visiting Professor. She is currently a Professor with the School of Computer
and Information Technology, Beijing Jiaotong University. Prof. Lang has published more than 80 research papers in various journals and refereed conferences. Her research areas include computer vision, machine learning, object recognition and segmentation.
\end{IEEEbiography}

\begin{IEEEbiography}[{\includegraphics[width=1in,height=1.25in,clip,keepaspectratio]{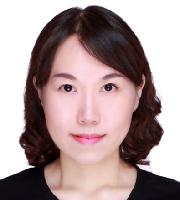}}]{Yi Jin}
received the Ph.D. degree in signal and information processing from the Institute of Information Science, Beijing Jiaotong University, Beijing, China, in 2010, where she is currently an Associate Professor with the School of Computer Science and Information Technology. She was a Visiting Scholar with the School of Electrical and Electronic Engineering, Nanyang Technological University, Singapore, from 2013 to 2014. Her research interests include computer vision, pattern recognition, image processing, and machine learning.
\end{IEEEbiography}

\begin{IEEEbiography}[{\includegraphics[width=1in,height=1.25in,clip,keepaspectratio]{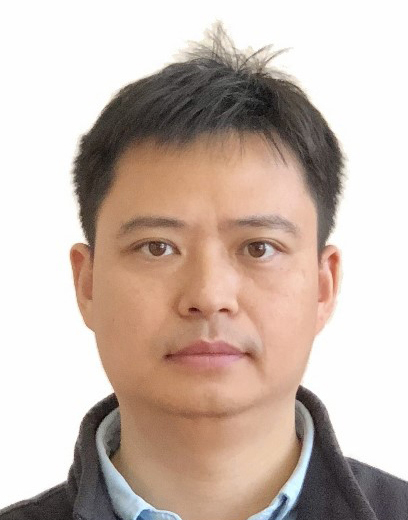}}] {Haibin Ling} received B.S. and M.S. from Peking University in 1997 and 2000, respectively, and Ph.D. from University of Maryland in 2006. From 2000 to 2001, he was an assistant researcher at Microsoft Research Asia; from 2006 to 2007, he worked as a postdoctoral scientist at UCLA; from 2007-2008, he worked for Siemens Corporate Research as a research scientist; and from 2008 to 2019, he was a faculty member of the Department of Computer Sciences for Temple University. In fall 2019, he joined the Department of Computer Science of Stony Brook University, where he is now a SUNY Empire Innovation Professor. His research interests include computer vision, augmented reality, medical image analysis, visual privacy protection, and human computer interaction.
\end{IEEEbiography}
%
%

\end{document}